\documentclass[10pt]{extarticle}
\usepackage[margin=1in]{geometry}
\usepackage{microtype}
\usepackage{graphicx}
\usepackage{subcaption}
\usepackage{booktabs}
\usepackage{natbib}

\usepackage[font=small]{caption}
\usepackage{authblk}

\usepackage{amsmath}
\usepackage{amssymb}
\usepackage{mathtools}
\usepackage{amsthm}

\usepackage{derivative}

\usepackage{titlesec}
\titleformat{\section}{\bfseries\fontsize{11}{13}\selectfont}{\thesection}{1em}{}
\titleformat{\subsection}{\bfseries\fontsize{10}{12}\selectfont}{\thesubsection}{1em}{}
\titleformat{\subsubsection}{\scshape\fontsize{10}{12}\selectfont}{\thesubsubsection}{1em}{}

\theoremstyle{plain}
\newtheorem{theorem}{Theorem}[section]
\newtheorem{informaltheorem}{Informal Theorem}[section]

\newtheorem{lemma}[theorem]{Lemma}
\newtheorem{corollary}[theorem]{Corollary}
\theoremstyle{definition}
\newtheorem{definition}[theorem]{Definition}
\newtheorem{assumption}[theorem]{Assumption}
\newtheorem{example}[theorem]{Example}
\newtheorem{gdcons}[theorem]{GD LR Decay Conditions}
\newtheorem{setting}[theorem]{Setting}

\theoremstyle{remark}

\newcommand{\eqdef}[0]{\coloneqq}
\newcommand{\risk}[0]{\mathcal{R}}
\newcommand{\dataset}[0]{\mathcal{S}}
\newcommand{\riskE}[0]{\risk_\dataset}

\newcommand{\riskHuber}[0]{\hat{\risk}_\dataset}
\newcommand{\riskSample}[0]{\risk_{\{(\x_n, y_n)\}}}

\newcommand{\lr}[1]{{\alpha}_t^{#1^{[m]}}}
\newcommand{\lrcon}[1]{\tilde{\alpha}_t^{#1^{[m]}}}

\newcommand{\MYth}[0]{^{\text{th}}}

\renewcommand{\a}[0]{\mathbf{a}}
\renewcommand{\vec}[1]{\mathbf{#1}}
\newcommand{\x}{\mathbf{x}}
\newcommand{\X}{\mathbf{X}}
\newcommand{\Xc}{\mathcal{X}}
\newcommand{\y}{\mathbf{y}}

\newcommand{\Yc}{\mathcal{Y}}
\newcommand{\z}{\mathbf{z}}

\newcommand{\R}{\mathbb{R}}
\newcommand{\N}{\mathbb{N}}

\newcommand{\Q}{\mathbb{Q}}

\newcommand{\Fc}{\mathcal{F}}

\renewcommand{\Pr}{\mathbb{P}}
\renewcommand{\Q}{\mathbb{Q}}

\newcommand{\E}[0]{\mathbb{E}}

\newcommand{\dd}[0]{\,\mathrm{d}}

\newcommand{\Lip}{\operatorname{Lip}}

\DeclareMathOperator{\supp}{supp}
\DeclareMathOperator{\diam}{diam}

\newcommand{\Cc}{\mathcal{C}}
\newcommand{\Oc}{\mathcal{O}}

\title{Step by Step: Adaptive Gradient Descent for Training $L$-Lipschitz Neural Networks}

\date{}

\usepackage[twoside]{fancyhdr}
\fancyheadoffset{0pt}

\usepackage[dvipsnames]{xcolor}
\definecolor{warmblack}{rgb}{0.0, 0.26, 0.26}
\definecolor{richblack}{rgb}{0.0, 0.25, 0.25}
\definecolor{darkcerulean}{rgb}{0.03, 0.27, 0.49}
\definecolor{smokyblack}{rgb}{0.06, 0.05, 0.03}
\definecolor{warmblack}{rgb}{0.0, 0.26, 0.26}
\definecolor{cobalt}{rgb}{0.0, 0.28, 0.67}
\definecolor{darkcobalt}{rgb}{0.1, 0.38, 0.77}
\usepackage[colorlinks=true,
            linkcolor=cobalt,
            urlcolor=darkcerulean,
            citecolor=cobalt]{hyperref}

\usepackage{style}

\hypersetup{
    pdftitle={Step by Step: Adaptive GD for Training Lipschitz Neural Networks},
    pdfauthor={Sung, Khalil, Kratsios, Samu, Forman}
}

\author[1]{\textbf{Kyle Sung}\thanks{Correspondence to: Kyle Sung \textless sungk5@mcmaster.ca\textgreater}}
\author[1]{\textbf{Kholood Khalil}}
\author[1]{\textbf{Noah Forman}}
\author[1]{\textbf{Steven Samu}}
\author[1,2,3]{\textbf{Anastasis Kratsios}}

\affil[1]{Department of Mathematics and Statistics, McMaster University, Hamilton ON, Canada}
\affil[2]{Vector Institute, Toronto ON, Canada}
\affil[3]{The Ennio De Giorgi Mathematical Research Centre, Scuola Normale Superiore di Pisa, Italy}

\begin{document}

\maketitle

\begin{abstract}
    We show that applying an \emph{eventual decay} to the gradient descent (GD) learning rate (LR) in MSE-based empirical risk (ER) minimization (ERM) ensures the resulting multilayer perceptron (MLP) exhibits a high degree of Lipschitz regularity. Moreover, this LR decay does not hinder the convergence rate of the ERM, now measured with the Huber loss, toward a stationary point of the non-convex ER. From these findings, we derive generalization bounds for MLPs trained with GD and a decaying LR chosen adaptively on the number of parameters.~We validate our theoretical results with numerical experiments, where surprisingly, we observe that MLPs trained with a constant LR exhibit similar learning and regularity properties to those trained with a decaying LR, suggesting MLPs trained with standard GD may already be highly regular learners.
\end{abstract}


\section{Introduction}
\label{s:Introduction}

There has recently been a great deal of research into the Lipschitz regularity of neural networks~\citep{virmaux2018lipschitz,herrera2023locallipschitzboundsdeep,jordan2020exactly,limmer2024reality,khromov2024some}. 
Lipschitz constraints on neural networks promote reliable generalization~\citep{bartlett2017spectrally,hou2023instance} and guarantee robustness to noise or errors in the network's inputs~\citep{tsuzuku2018lipschitz,singla2021improved,zhang2022rethinking,fazlyab2024certified}.  
Moreover, these constraints are not too onerous: 
the class of 
$L$-Lipschitz neural networks is universal in the class of $L$-Lipschitz functions on $[0,1]^d$~\citep{hong2024bridging}. 
Lipschitz neural networks are closely connected to generative modeling via neural optimal transport~\citep{korotinwasserstein,korotin2023neural,benezet2024learning,kolesov2024energyguided,persiianov2024inverse} due to the Kantorovich-Rubinstein duality popularized in deep learning by Wasserstein GANs~\citep{arjovsky2017wasserstein}. 

In practice, neural networks are typically guided toward a desired level of Lipschitz regularity either by: (a) incorporating a differentiable regularizing penalty into the objective function optimized during training~\citep{fazlyab2024certified}, or (b) fundamentally modifying the neural network layers themselves~\citep{trockman2021orthogonalizing,meunier2022dynamical,wang2023direct,araujo2023a}. Approach (a) places these training objectives beside standard empirical risk minimization (ERM), while (b) places these modified architectures beside the scope of student deep learning models. It is, therefore, natural to ask if Lipschitz regularity can be simply achieved in multilayer perceptrons (MLPs) with standard activation functions trained with conventional \emph{gradient descent} (GD) with sub-Gaussian initialization of parameters. Thus, this paper studies the following question:
\begin{align}
    \label{eq:MainQ}
    \tag{Q}
    \text{\emph{Can a neural network achieve a target Lipschitz}}~\text{\emph{constant by adaptively adjusting the GD learning rate?}}
\end{align}
To the best of our knowledge, the answer to this question is only understood at the exact moment of (random) GD parameter \emph{initialization}, for which tight bounds are known~\citep{geuchen2024upperlowerboundslipschitz}, while the general question, concerning 
MLPs trained with randomly initialized GD, remains open.
\noindent
\begin{figure}[h]
    \centering
    \begin{minipage}{0.6\textwidth}
        \centering
            \includegraphics[width=\linewidth]{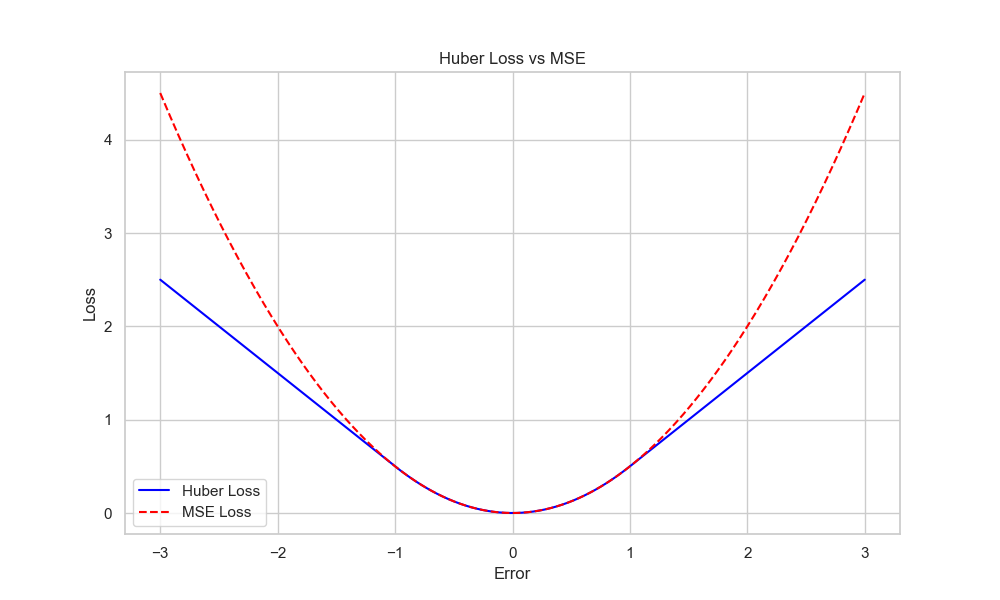}
    \end{minipage}
    ~
    \begin{minipage}{0.4\textwidth}
    \[
    \begin{aligned}
          \ell \colon \R^2 & \rightarrow [0,\infty)
            \\
                \ell(\hat{y},y)
            & \eqdef 
                \begin{cases}
                \frac{1}{2}{|\hat{y}-y|^2}                   & \text{for } |\hat{y}-y| \le \delta, \\
                |\hat{y}-y| - \frac{1}{2}
                & \text{otherwise.}
                \end{cases}
            \end{aligned}
    \]
    \end{minipage}
    \caption{The Huber loss $(\ell)$: a $1$-Lipschitz surrogate of the mean squared error loss.}
    \label{fig:huber_loss}
\end{figure}
Here, we study Question~\eqref{eq:MainQ} by training on the empirical risk (ER) under the mean squared error (MSE) loss function, illustrated by the {\color{red}{dashed red curve}} in Figure~\ref{fig:huber_loss}, and we validate and test on the risk associated to the Huber loss function, illustrated by the {\color{blue}{blue curve}} in Figure~\ref{fig:huber_loss}.  These two loss functions have the same minima, with the MSE allowing for more convenient training dynamics, while the Huber loss is $1$-Lipschitz, which is more convenient for statistical and optimization guarantees.  

\begin{figure}[h]
    \centering
    \includegraphics[width=0.75\linewidth]{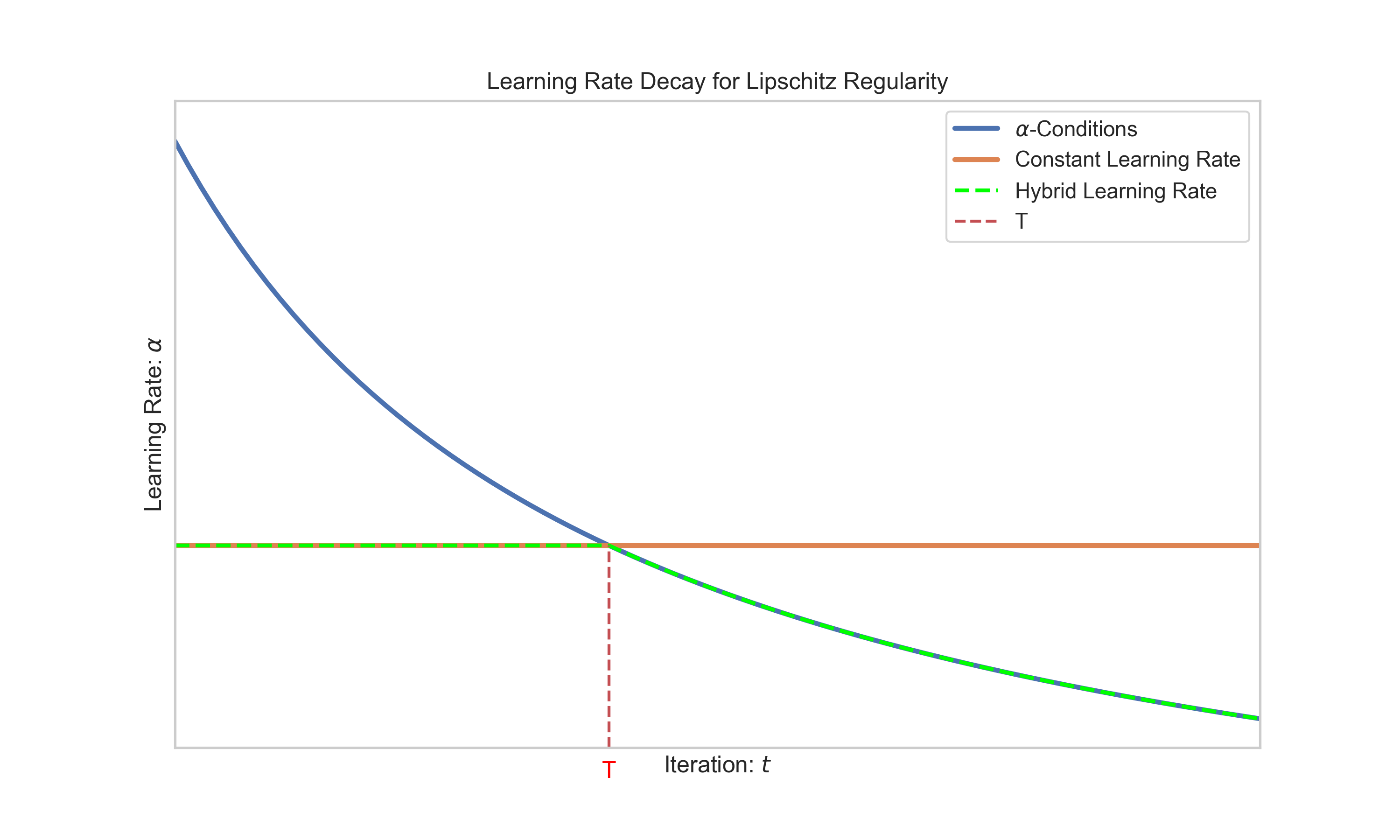}
    \caption{\textbf{Learning Rate Decay:} A decaying LR ({\color{blue}{blue}} curve) guarantees a desired Lipschitz regularity of the MLP (Theorem~\ref{thm:lip_control_via_lr_decay}), while the standard constant LR ({\color{orange}{orange}} line) guarantees convergence to a stationary point of the ER. Following the constant {\color{orange}{orange}} line for a long enough time horizon $T>0$ and then tracing the {\color{blue}{blue}} curve's decay rate yields both quadratic convergence to a stationary point of the ER and the desired degree of Lipschitz regularity of the trained function (Theorem~\ref{thm:convergence_optimal_gd_rate}). In particular, the class of two-layer neural networks trained in this manner with GD tracing this two-part {\color{green}{green}} ``hybrid'' curve generalize at a rate independent of their width (Corollary~\ref{cor:generalization_bounds_gd_trained_networks}).}
    \label{fig:idea}
\end{figure}

This paper provides a \textbf{positive} answer to this question. As shown in Figure~\ref{fig:idea}, the key is that the learning rate (LR) must \emph{eventually decay} ({\color{blue}{blue}} curve) at a specific rate, in contrast to standard GD, which maintains a constant LR ({\color{orange}{orange}} line).  
This decay does not need to begin immediately, and can instead begin after an arbitrarily long but prescribed time horizon ($T>0$).  
Decaying the LR in this way implies the following special case of our main result:
\begin{informaltheorem}
    \label{IntroTheorem}
    Let $f_{\theta_t}$ denote a neural network with a single hidden layer with $p$ neurons and with a $1$-Lipschitz and differentiable activation function. Let $\theta_t$ denote the 
    parameter vector at the $t\MYth$ GD epoch and suppose $\theta_0\sim \mathcal{N}(0, I_P)$.
    If the learning rate decays at $\mathcal{O}(e^{-rt})$ 
    then
    \begin{equation*}
        \underset{t=0,\dots,T}{\max}
        \,
                \operatorname{Lip}(f_{\theta_t}) 
            \in 
                \mathcal{O}
                \biggl(
                    \big ( 
                    \,
                        \sqrt{p}
                        + \frac{
                            \frac{1}{r} (2 - e^{-rT})
                        }{N}
                    \big )^2
                \biggr)
    \end{equation*}
    with probability at least $1 - 4 e^{-p}$.
\end{informaltheorem}

Consequently, we find that running standard GD with a fixed LR up to time $T$, followed by decay at the prescribed rate, ensures both optimal convergence to a stationary point of the ERM functional and that the trained two-layer MLP achieves the desired Lipschitz regularity.

\paragraph{Our Contribution}
Our first main result (Theorem~\ref{thm:lip_control_via_lr_decay}) shows that for any sufficiently large Lipschitz constant $L$, an MLP with Lipschitz activation functions trained with GD becomes $L$-Lipschitz after $T$ iterations, provided our data-dependent 
GD LR Decay Conditions~\ref{gdcons:grad} are satisfied.

As a consequence, we derive a generalization bound for MLPs trained with GD for $T$ iterations on $N$ i.i.d.\ noisy training instances (Corollary~\ref{cor:generalization_bounds_gd_trained_networks}) in $\R^d \times \R$. This generalization bound converges at the non-parametric rate of $\mathcal{O}(1/\sqrt[d]{N})$ with high probability, and the constant in the bound can be explicitly controlled by adjusting $T$ and $L$.

Importantly, we show that our step size schedule is compatible with traditional GD convergence requirements. Specifically, under additional mild assumptions on the activation function's derivatives, an optimal convergence rate to a stationary point of the ERM loss can be achieved by starting with a constant step size and letting it decay at the prescribed rate (Theorem~\ref{thm:convergence_optimal_gd_rate}) of $\mathcal{O}(1/\sqrt{T})$.

\subsection{Related Work}
\label{s:Introduction__ss:RelatedWork}

\paragraph{The Inductive Bias Encoded into Neural Networks by Gradient Descent}

Substantial research has focused on the implicit biases induced by GD during the ERM in neural network architectures. Progress has been made in explaining these biases by showing that sufficiently wide two-layer MLPs tend to converge toward cubic interpolating splines~\citep{jin2023implicit} and exhibit benign overfitting~\citep{pmlr-v178-shamir22a}. These properties of GD-trained networks are also reflected in the kernelized infinite-width limit for neural networks with randomized hidden layers~\citep{heiss2023implicit,mei2022generalization}, which themselves are well-understood surrogates for standard two-layer MLPs trained with GD~\citep{gonon2020risk,gonon2023approximation}. Indeed, it is known that there is a gap between the statistical behaviour of two-layer networks when only the final layer is trained when compared to two-layer neural networks trained using only a single GD step; the latter learns superior latent feature representations~\citep{ba2022high}.

In the case of deep neural networks, it has been shown that GD leads MLPs with more than two layers and identity activations to solve a principal component analysis (PCA) problem~\citep{arora2018a}. 
Additionally, there is a large body of literature on the training dynamics of neural networks in the mean-field or infinite-width regime~\citep{pmlr-v195-abbe23a,pmlr-v119-golikov20a}. Other works examine the negative impact of overparameterization on GD convergence rates in the student-teacher setting~\citep{xu2023over}. 

\paragraph{Variations of Gradient Descent}

GD and its variants have been extensively studied in the non-convex optimization literature. Research has focused on their convergence rates to stationary points of the ER functional~\citep{pmlr-v247-patel24a,pmlr-v247-zeng24a}, stationary points of specific non-convex function classes~\citep{faw2023beyond,anonymous2025methods}, and under spectral conditions~\citep{velikanov2024tight}. The impact of hyperparameters like the LR on GD convergence~\citep{pmlr-v247-wu24b}, as well as its stability properties~\citep{mulayoff2024exact,zhu2024uniform,zhu2024stability}, has also been well explored. Additionally, GD has proven effective in solving specialized problems like matrix completion~\citep{baes2021low,pmlr-v247-ma24a}.
Recent work has also explored the design of GD-type algorithms to meet meta-optimality criteria~\citep{casgrain2019latent,pmlr-v134-casgrain21a}, studying non-Euclidean settings~\citep{alimisis2021momentum,hsieh2024riemannian}, and extending GD to proximal splitting methods for objectives that combine differentiable and lower semi-continuous convex parts~\citep{patrascu2021stochastic,minh2022strong}.

\subsection{Paper Outline}
\label{s:Introduction__ss:Outline}
Section~\ref{s:prelim} provides an overview of the notation, background, definitions, and training dynamics of neural network optimization with GD.
Section~\ref{s:MainResults} presents our GD LR Decay Conditions and main results: the Lipschitz regularity findings in subsection~\ref{s:MainResults__ss:LipGuide}, the implied generalization bounds in subsection~\ref{s:MainResults__ss:Implications} with dependence on the number of trainable neural network parameters, and the alignment of our results with well-known optimization guarantees for GD with a fixed step size in subsection~\ref{s:MainResults__ss:Optim}. 
All proof details are relegated to Appendix~\ref{appendix:proofs}.

Our theoretical results are then validated on experiments in Section~\ref{s:Experiments}, where we confirm that neural networks trained with our LR decay exhibit superior Lipschitz regularity compared to those trained with constant LR, while retaining predictive power.

\section{Preliminaries} 
\label{s:prelim}

\subsection{Notation}
 \label{s:prelim__ss:notation}

Let $\N_+$ denote the positive integers and $\N\coloneq \N_+\cup\{0\}$.
Let $f,g\colon \N\to \R$; we write $f\lesssim g$ if there is a constant $c>0$ such that $f(t)\leq c\,g(t)$ for every sufficiently large $t\in \N$; equivalently, we write $f\in \mathcal{O}(g)$.
We use subscripts to denote the time evolution of a parameter or the index of a training sample. We use 
square-bracketed superscripts to denote layers of a neural network.
All random variables are defined on the probability space $(\Omega,\mathcal{F},\mathbb{P})$.

\subsection{Background}
\label{s:prelim__ss:Defs}
The definitions required to formulate our main results are aggregated here.

\paragraph{Norms} Let $\|\cdot \|$ denote the Euclidean norm on $\R^d$ and denote the spectral norm on the set of matrices.

\paragraph{Lipschitz Continuity}
We study neural networks of a given Lipschitz regularity. We say a map $f \colon \R^d\to \R$ is Lipschitz continuous if there exists $L>0$ such that 
\begin{equation}
    \label{eq:Lipschitz_definition}
    | f(\mathbf{x}_1) - f(\mathbf{x}_2) |
    \leq 
    L \, \| \mathbf{x}_1 - \mathbf{x}_2 \|
\end{equation}
for every $\mathbf{x}_1,\mathbf{x}_2\in \R^d$. 
If $L$ is the smallest number satisfying Equation~\eqref{eq:Lipschitz_definition}, we say that $f$ is $L$-\emph{Lipschitz} and write $\operatorname{Lip}(f) = L$.
By Rademacher's Theorem~\citep{Federer_GeometricMeasureTheory_1978}, if a function is $L$-Lipschitz then it is Lebesgue differentiable almost everywhere and the norm of its gradient is no larger than $L$ (wherever it is defined).

\paragraph{Isotropic Sub-Gaussian Distributions}
A random vector $\X$ taking values in $\R^p$ is said to be \emph{isotropic} if $\mathbb{E}[\X\X^{\top}]=I_p$ where $I_p$ is the $p\times p$ identity matrix. A random vector is said to be $c$-\emph{sub-Gaussian} with tail parameter $c>0$ if for all ``tail levels'' $t\geq 0$, we have $\Pr(\|\X\| \geq t)\leq 2e^{-t^2/c^2}$.

\paragraph{Multilayer Perceptrons}

Fix a depth $M\in \N_+$, an input size $d_0 \in \N_+$, widths $d_1,\ldots, d_{M - 1} \in \N_+$, an output size $d_M = 1$, and a maximum admissible weight $w \in \R_+ \cup \{\infty\}$. Let $A^{[m]} \colon \R^{d_{m-1}}\to \R^{d_m}$ denote the affine map 
\[
    A^{[m]}(\vec x) = W^{[m]} \vec x + b^{[m]}
\] 
for $1\leq m \leq M$. By a slight abuse of notation, let $\theta \coloneq (W^{[1]},b^{[1]},\ldots,W^{[M]},b^{[M]})$ 
be the flattened parameter vector and
let $P$ denote the total number of parameters.

Following~\citet{MR4243432}, we distinguish the function realized by a neural network from its parameter space. We denote the \emph{parameter space} of the class of neural networks by $\Theta_w \coloneq \{ \theta \in \R^P : \| \theta \|_\infty \leq w \}$.

\begin{definition}[Realization of an MLP]
    \label{def:MLP}
    Fix an 
    activation function $\sigma \colon \R\to \R$. Then, any parameter vector $\theta \in \Theta_w$ realizes an $M$-layer multilayer perceptron (MLP) $f_\theta \colon \R^{d_0} \to \R$ mapping any $\vec x \in \R^{d_0}$ to \begin{equation*}
        f_\theta (\vec x)
        \coloneq
        (A^{[M]} \circ \sigma \bullet \cdots \circ \sigma \bullet A^{[1]} ) (\vec x),
    \end{equation*}
    where $\bullet$ denotes componentwise composition, that is, 
    $
        \sigma \bullet \mathbf{v} 
        \coloneq 
        \big (
            \sigma(v^i)
        \big )_{i=1}^p
        $.
\end{definition}

\subsection{Training Objectives and Dynamics}
We now describe our training and evaluation objectives, as well as the dynamics of our GD policy.

\subsubsection{Training and Evaluation Objectives} 
Let $(\mathbf{x}_1, y_1),\ldots, (\mathbf{x}_N, y_N)\in \dataset$ be i.i.d.\ random variables on $\R^{d_0} \times \R$ drawn according to a data-generating distribution (probability measure) $\mathbb{Q}$ on $\R^{d_0} \times \R$. 
Our objective is to minimize the \emph{empirical risk} (ER) associated with the MSE:
\begin{equation}
    \begin{aligned}
    \label{eq:objective_raw}
    \underbrace{\riskE (\theta) 
    \eqdef 
        \frac{1}{N} \sum \limits_{n=1}^N \lVert f_\theta ( \mathbf{x}_n ) - y_n \rVert^2}_{\text{MSE-based ER}}
    \quad\text{and}\quad
    \underbrace{\riskHuber (\theta) 
    \eqdef 
        \frac{1}{N} \sum \limits_{n=1}^N 
            \ell\big (
                    f_\theta ( \mathbf{x}_n ) 
                , 
                    y_n 
            \big )}_{\text{Huber-based ER}}
    \end{aligned}
\end{equation}
where $f_{\theta}$ is an MLP as in Definition~\ref{def:MLP} and $\ell$ is the Huber loss function (Figure~\ref{fig:huber_loss}).
The out-of-sample performance of our models is quantified by their \emph{true risk}, defined
by
\begin{equation*}
\label{eq:true_Huber_risk}
    \risk (\theta)
    \coloneq 
    \mathbb{E}_{(\mathbf{x},y)\sim \mathbb{Q}}\big[
        \ell(f_{\theta}(\mathbf{x}),y)
    \big]
.
\end{equation*}

\noindent Having defined the MSE-based ER, we may use it as a training objective for GD-based ERM.

\subsubsection{Gradient Descent with Variable Step Size}

Fix an initial parameter $\theta_0$, which is a random vector in $\Theta_w$ defined on $(\Omega,\mathcal{F},\mathbb{P})$. We consider \emph{variable learning rates} $\boldsymbol{\alpha} \coloneq (\alpha_t)_{t=0}^{\infty}$ where the vector of LRs at iteration $t$ for each neural network parameter belongs to $\alpha_t\coloneq (\alpha_t^{W^{[1]}},\alpha_t^{b^{[1]}},\ldots, \alpha_t^{W^{[M]}},\alpha_t^{b^{[M]}})\in [0,\infty)^{2M}$.

The \emph{iterates} of the GD algorithm used to minimize the MSE ER functional $\riskE$, defined in~Equation~\eqref{eq:objective_raw}, over the (random) training data $\dataset$, and given a variable LR $\boldsymbol{\alpha}$ can be written explicitly\footnote{See Lemma~\ref{lem:explicit_gd_formulations} for an explicit expression for the gradient of the ER with respect to the parameters in each layer.}, and define a stochastic process $\boldsymbol{\theta} \coloneq (\theta_t)_{t=0}^{\infty}$ taking values in $\Theta_w$ and defined on $(\Omega,\mathcal{F},\mathbb{P})$. 
For each $t\in \N_+$, the $t\MYth$ iterate is given \emph{recursively} by
\begin{equation}
    \begin{aligned}
    \label{eqn:gd_mlp_defn}
    W_t^{[m]} &\coloneq W_{t-1}^{[m]} - \alpha_t^{W^{[m]}} \nabla_{W^{[m]}} \riskE (\theta_{t-1}) \\ b_t^{[m]} &\coloneq b_{t-1}^{[m]} - \alpha_t^{b^{[m]}} \nabla_{b^{[m]}} \riskE (\theta_{t-1}).
    \end{aligned}
\end{equation}
Note that $\boldsymbol{\theta}$ is measurable and adapted to the filtration $\boldsymbol{\mathcal{F}}\coloneq (\Fc_t)_{t=0}^{\infty}$ where, for each $t$, 
$\Fc_t$ is the $\sigma$-algebra generated by $\{\dataset\}\cup\{\theta_s\}_{s=0,\ldots,t}$. In particular, $\boldsymbol{\theta}$ depends on the training data $\dataset$ and on the initial parameter distribution $\theta_0$. 
One can couple the LRs by setting $\alpha_t^{W^{[1]}}\coloneq \alpha_t^{b^{[1]}} \coloneq \cdots \coloneq \alpha_t^{W^{[M]}}\coloneq \alpha_t^{b^{[M]}}$. Though our main result allows this common simplification, we do not require it.

\subsubsection{Rate Functions} 
We will enforce convergence to a given Lipschitz constant at a pre-specified rate: for example,\ super-exponential, exponential, polynomial, logarithmic, or even constant.
\begin{definition}[Rate Function]
    \label{def:RateFunction}
        A \emph{rate function} is a bounded, non-decreasing, and absolutely continuous map $G\colon [1,\infty)\to [0,\infty)$ whose (Lebesgue) a.e.\ derivative $g\coloneq G'$ is 
        continuous.
    \smallskip
    \hfill\\
    We denote the asymptotically maximal value of a rate function $G$ by $G^{\star}\coloneq \sup_{t\geq 1}\, G(t)$, and we extend $G$ to $[1,\infty]$ by setting $G(\infty) \coloneq G^{\star}$.
\end{definition}

\noindent The simplest examples of rate functions yield exponential, polynomial, and constant rates.

\begin{example}[Exponential and Polynomial Rate Functions]
    \label{ex:polynomial_convergence}
    Fix $\lambda>0$ and $r>0$. Then, $G_{\exp:r}(t) \coloneq \lambda(1-e^{-rt})$ and $G_{\operatorname{poly}:r}(t) \coloneq \lambda\big ( 1 + \frac{t^{1-r}}{1-r}\big )$ are rate functions satisfying $G(0)=\lambda$ and whose first derivative decays at exponential and polynomial rates, respectively.
\end{example}
Focal examples are rate functions whose derivative $g$ is illustrated by the {\color{green}{green}} piecewise smooth curve in Figure~\ref{fig:idea}, which begin constant and then rapidly decay toward zero.
\begin{example}[Hybrid-Exponential Rate Function]
    \label{ex:truncated_exponential}
    Fix any constant $\lambda > 0$ and a time-horizon $\tau \in \N_+$. 
    The hybrid-exponential rate function is given implicitly by its derivative, defined for each $t\geq 0$ by
    \[
    g(t) \coloneq \lambda\,
    \begin{cases}
        1 & \mbox{ if } t\leq \tau
        \\
        e^{-r(t-\tau)} & \mbox{ if } t>\tau
    .
    \end{cases}
    \]
    Specifically, $G(t) = \lambda + \int_0^t\, g(s)\,\mathrm{d}s$. Moreover, $G(1) = \lambda$.
\end{example}

\section{Main Results} \label{s:MainResults}

We will work under the following setting:

\begin{setting}
    \label{setting:main}
    Fix an input size $d_0 \in \N_+$, a sample size $N\in \N_+$, and let $\dataset \coloneq \{(\x_n, y_n)\}_{n=1}^N$ be identically distributed (but not necessarily independent) random variables in $\R^{d_0} \times \R$ defined on the probability space $(\Omega, \Fc, \Pr)$ with law $\pi$. Fix a Lipschitz continuous and differentiable activation function $\sigma \colon \R\to \R$ and a rate function $G$ (Definition~\ref{def:RateFunction}). Let $f_\theta\colon \R^{d_0} \to \R$ be an MLP (Definition~\ref{def:MLP}).
\end{setting}

\begin{assumption}[Standard Parameter Initialization]
    \label{assumption:standard_param_initialization}
    Let $\theta_0$ be independent of $\dataset$ and suppose the initial weight matrices $\sqrt{d_1} W_0^{[1]}, \ldots, \sqrt{d_M} W_0^{[M]}$ have independent, centered, isotropic, $c$-sub-Gaussian rows.
\end{assumption}

The rescaling of the weight matrices is typical in MLP initialization and is well established, as in Xavier and He initialization \citep{pmlr-v9-glorot10a,KaimingHe2015}.

\begin{gdcons} 
    \label{gdcons:grad}
    Let $G$ be a rate function and let $
        C_{W^{[1]}},
        C_{b^{[1]}},
        \ldots,
        C_{W^{[M]}},
        C_{b^{[M]}}
    $ be free positive parameters. We require that the learning rates $
        \alpha_t^{W^{[1]}},
        \alpha_t^{b^{[1]}},
        \ldots,
        \alpha_t^{W^{[M]}},
        \alpha_t^{b^{[M]}}
    $ in Equation~\eqref{eqn:gd_mlp_defn} satisfy
    \begin{equation*}
        \begin{alignedat}{3}
            \alpha_t^{W^{[m]}} & \leq  
                \dfrac{C_{W^{[m]}} g(t)}{\sum\limits_{n=1}^N \left\| (\nabla_{W^{[m]}} f_{\theta_{t-1}}(\x_n))^\top  \Big [f_{\theta_{t-1}}(\x_n) - y_n \Big ] \right\|_2}
            \\
            \alpha_t^{b^{[m]}} & \leq  
                \dfrac{C_{b^{[m]}} g(t)}{\sum\limits_{n=1}^N \left\| (\nabla_{b^{[m]}} f_{\theta_{t-1}}(\x_n))^\top  \Big [f_{\theta_{t-1}}(\x_n) - y_n \Big ] \right\|_2}
            .
        \end{alignedat}
    \end{equation*}
\end{gdcons}

\noindent We are now in place to present our main results.

\subsection{Guidance Toward Lipschitz Regularity via Learning Rate Decay} \label{s:MainResults__ss:LipGuide}

Our first main result shows that by respecting 
the GD LR Decay Conditions~\ref{gdcons:grad}, we obtain the following high-probability bound on the Lipschitz constant of any MLP trained by GD. The following is a rigorous version of Informal Theorem~\ref{IntroTheorem}.

\begin{theorem}[Lipschitz Control with Learning Rate Decay]
    \label{thm:lip_control_via_lr_decay}
    In Setting~\ref{setting:main}, suppose the parameter initialization satisfies Assumption~\ref{assumption:standard_param_initialization}. If the variable learning rate $\boldsymbol{\alpha}$ satisfies the GD LR Decay Conditions~\ref{gdcons:grad} then there is an absolute constant $\kappa > 0$ such that for every $\eta > 0$ and every number of GD epochs $T \in \N_+$, 
    \begin{align}
        \Lip(f_{\theta_t})
        &\leq
        \Lip(\sigma)^{M-1} \!\! \prod_{m=1}^M \!
        \Bigg[ \sqrt{\frac{d_m}{d_{m-1}}} 
        + \kappa \! \left[1\! +\! \dfrac{\eta}{\sqrt{d_{m-1}}}\right] 
         + \frac{2C_{W^{[m]}}}{N} (G(t) + g(1)) \Bigg]
        \label{eqn:lip_control_via_lr_decay__MAIN}
    \end{align}
    with probability at least $1 - 2Me^{-\eta^2}$, where $\kappa$ depends only on the distributions of $W_0^{[1]},\ldots, W_0^{[M]}$.
\end{theorem}

We focus our examples on two-layer MLPs, as much of the literature focuses on the Lipschitz continuity of two-layer networks \citep{pmlr-v134-bubeck21a,NEURIPS2021_f197002b} (subsection~\ref{s:Introduction__ss:RelatedWork}). Our first example follows the {\color{blue}blue} curve in Figure~\ref{fig:idea} with polynomial decay. Since it was recently shown by~\citet{hong2024bridging} and \citet{riegler2024generatingrectifiablemeasuresneural} that certain interpolating MLPs can match the Lipschitz regularity of the target function, we freely allow ourselves to specify the target Lipschitz constant.  

\begin{example}
    [Lipschitz Bounds for Polynomial LR Decay]
    \label{ex:lip_bounds_poly_lr_decay}
    In the setting of Theorem~\ref{thm:lip_control_via_lr_decay}, let $M=2$, $d_1 \geq d_0$, $C_{W^{[1]}}=C_{W^{[2]}}=1$, $\sigma \coloneq \operatorname{tanh}$, and let $G \coloneq G_{\operatorname{poly}:r}$ be the polynomial decay rate of Example~\ref{ex:polynomial_convergence} with $r>0$. Then, there is an absolute constant\footnote{Set $\tilde k\coloneq 2 \max \{ 1 , \kappa \}$ and $\eta \coloneq \sqrt{d_1}$.} $\tilde\kappa >0$ such that for every number of GD epochs $t\in \N_+$, 
    \[
        \Lip(f_{\theta_t})
        \leq
        \left[
            \tilde \kappa
            \sqrt{\dfrac{d_1}{d_0}} 
            - \frac{
                2 [
                    2\lambda(1-r) - t^{1-r}
                ]
            }{
                N (1-r)
            }
        \right]^2
    \]
    holds with probability at least $1 - 4e^{ - d_1}$.
\end{example}

Returning to our focal example illustrated by the {\color{green}green} curve in Figure~\ref{fig:idea}, we may apply Theorem~\ref{thm:lip_control_via_lr_decay} to obtain Lipschitz guarantees for the hybrid-exponential rate function of Example~\ref{ex:truncated_exponential}.

\begin{example}
    [Lipschitz Bounds for Hybrid-Exponential LR Decay]
    \label{ex:lip_bounds_hybrid_exponential}
    In the setting of Theorem~\ref{thm:lip_control_via_lr_decay}, let $M=2$, $d_1 \geq d_0$, $C_{W^{[1]}}=C_{W^{[2]}}=1$, $\sigma \coloneq \operatorname{tanh}$, and let $G$ be the hybrid-exponential rate function of Example~\ref{ex:truncated_exponential}
    with $r >0$ and $\tau \in \N_+$. 
    Then, there is an absolute constant\footnote{Set $\bar \kappa \coloneq 2 \max \{ 1 , \kappa \}$ and $\eta \coloneq \sqrt{d_1}$.} $\bar \kappa >0$ such that for every number of GD epochs $t\in \N_+$, 
    \[ 
        \Lip( f_{\theta_t} )
        \leq
        \left[ \bar \kappa \sqrt{\dfrac{d_1}{d_0}} + \frac{2}{N} \Big(\lambda ( \tau + 2 ) + \frac{1 - e^{-r (t - \tau)}}{r} \Big) \right]^2
    \]
    holds with probability at least $1 - 4e^{-d_1}$.
\end{example}

Interestingly, as the sample size $N$ grows arbitrarily large, the Lipschitz constant of the
GD iterations remain uniformly bounded in time. This mirrors the convergence of GD with our step size
iterations to a ``well-behaved'' subclass of neural networks given independently of the sample size.

Our main result has several consequences in learning theory and in the optimization theory of neural
networks, which hold under additional mild structural assumptions. 

\subsection{LR Decay Implies Generalization Bounds with Linear Parametric Dependence} \label{s:MainResults__ss:Implications}

We first examine the learning-theoretic implications of our main result. The guarantee shows that by
specifying the number of GD steps $ t \in \N_+ $ and the rate function $G$, we may modulate the generalization of our GD-trained neural network, trained on the (random) training set $\dataset$.
The bound implied by our analysis matches the rate by~\citet{hou2023instance} but with user-specifiable constant implied by adaptive learning rate decay.

\begin{corollary}
    [Generalization Bounds for GD-Trained Networks with Linear Width Dependence]
    \label{cor:generalization_bounds_gd_trained_networks}
    In the setting of Theorem~\ref{thm:lip_control_via_lr_decay}, suppose $\dataset = \{(\x_n, y_n)\}_{n=1}^N$ are i.i.d. with compactly supported law $\Q$ and suppose that $d_0 + d_M > 2$. For every number of GD epochs $T \in \N_+$ and every \emph{confidence level} $\delta \in (0, 1]$, the following holds with probability at least $1 - \delta$:
    \begin{equation*}
        | 
            \risk (f_{\theta_T}) - \riskE (f_{\theta_T})
        |
        \leq
        \Lambda
        C_{\Q}
        \biggl(
            \frac{C_{d_0 + d_M}}{
            N^{1/d_0}
            } + \frac{\ln(4M/\delta)}{\sqrt{2N}}
        \biggr)
    \end{equation*}
    where $\Lambda \coloneq \Lambda(T, N)$ denotes the right-hand side of Equation~\eqref{eqn:lip_control_via_lr_decay__MAIN} with $\eta \coloneq \frac{\diam(\supp(\Q))}{\sqrt{2N}}$, $C_\Q \coloneq \linebreak \diam(\supp(\Q))$, and $C_{d_0 + d_M}>0$ is a dimensional constant of order $\Oc (\sqrt{d_0 + d_M})$. Critically, $C_{ d_0 + d_M }$ and $\eta$ do not depend on $d_1,\ldots, d_{M-1}$ and $\Lambda (T, N) \in  \Oc \left( 
            \left(
            \Lip(\sigma) \cdot 
            \left[
            \sqrt{\max \{d_0,\ldots,d_{M-1}\}} 
                \:+\: 
            \frac{G(T)}{N}
            \right]
            \right)^M
        \right)$.
\end{corollary}

We elucidate Corollary~\ref{cor:generalization_bounds_gd_trained_networks} in the case of random inputs on the $d_0 - 1$ dimensional sphere $\mathbb{S}^{d_0 - 1} \coloneq \{\x \in \R^{d_0}: \| \x \| < 1/2\}$ of radius $1/2$.

\begin{example}[Random Inputs on Spheres with Polynomial LR Decay]
    \label{ex:random_inputs_spheres_poly_lr_decay}
    In the setting of Corollary~\ref{cor:generalization_bounds_gd_trained_networks} and Example~\ref{ex:lip_bounds_poly_lr_decay}, let $\sigma \coloneq \operatorname{tanh}$, and let $G \coloneq G_{\operatorname{poly}:r}$ be the polynomial decay rate of Example~\ref{ex:polynomial_convergence} with $r > 0$. Suppose $\Q \coloneq \mu \otimes \nu$, where $\mu$ is any probability distribution on $\mathbb{S}^{d_0 - 1}$ and $\nu$ is any probability distribution on $[0,1]$. Then, for every number of GD epochs $T \in \N_+$ and every \emph{confidence level} $\delta \in (0,1]$,
    \begin{equation*}
    \begin{split}
        \big| 
            \risk (f_{\theta_T})\! -\! \riskE (f_{\theta_T})
        \big|
        \!\leq\! 
        \left[
            \tilde \kappa
            \sqrt{\dfrac{d_1}{d_0}} 
            - \frac{
                2 [
                    2\lambda(1\!-\!r) - T^{1-r}
                ]
            }{
                N (1-r)
            }
        \right]^{\!2} 
        \left(
            \frac{C_{d_0 + 1}}{\sqrt[d_0]{N}}
            +
            \frac{\ln(8 / \delta)}{\sqrt{2N}}
        \right)
        \in
        \Oc
        \left(
            \frac{d_1 \sqrt{\ln(8/\delta)}}{\sqrt[d_0]{N}}
        \right)
    \end{split}
    \end{equation*}
    holds with probability at least $1 - \delta$.
\end{example}

We now show that our main result is compatible with existing optimization guarantees for GD,
allowing both convergence and the regularity and learning guarantees discussed.

\subsection{Coalescence of \texorpdfstring{Theorem~\ref{thm:lip_control_via_lr_decay}}{Our Main Result} with Optimization Guarantees for GD} \label{s:MainResults__ss:Optim}

The asymptotically decaying step size required by Theorem~\ref{thm:lip_control_via_lr_decay} is an unorthodox choice in the optimization
literature.~This raises a natural question:~does this eventual modification to the LR impact the convergence of GD to a \emph{stationary point} of the ER? Surprisingly, we find that our
eventual step size modification comes at no cost to the GD convergence rate and only yields the additional
benefit of gaining control of the Lipschitz constant of the trained MLP.

We require the following additional mild regularity requirements on the activation functions used, as
well as some regularity of the target function and the input data.

\begin{assumption}[Additional Structure for Convergence]
    \label{assumption:additional_structure_for_convergence}
    Suppose the feature vectors $(\x_n)_{n=1}^N$ have finite second moment. 
    Assume also that the activation function $\sigma$ is bounded with bounded first and second derivatives.
\end{assumption}

Under the structural conditions in Assumption~\ref{assumption:additional_structure_for_convergence}, our LR modification maintains the convergence guarantee of GD. Importantly, it is not clear apriori that the ER functional $\riskE$ is locally Lipschitz, even in view of the results by \citet{JMLR:v24:22-1381} and \citet{herrera2023locallipschitzboundsdeep}. This raises questions about defining the optimal constant GD LR for rapid convergence, which typically depends on the reciprocal Lipschitz constant of the loss \citep[Section 1.2.3]{Nesterov}. However,
our next and final result shows that, under the GD LR Decay Conditions~\ref{gdcons:grad}, the ER is locally Lipschitz, and the hybrid LR decay in Figure~\ref{fig:idea} ensures both convergence and Lipschitz regularity of the
learned network.

\begin{theorem}[Convergence at the Optimal GD-Rate]
    \label{thm:convergence_optimal_gd_rate}
    In Setting~\ref{setting:main} and under Assumptions~\ref{assumption:standard_param_initialization}~and~\ref{assumption:additional_structure_for_convergence}, let $f_{\theta_t}$ be a multilayer perceptron. Then, $\riskE$ has a finite Lipschitz constant on $\Theta_w$, that is, $\Lip_\theta (\riskE) < \infty$.\\
    \noindent Consequently, the following strengthened constraints are well-defined and ensure that the Lipschitz bounds in~Equation~\eqref{eqn:lip_control_via_lr_decay__MAIN} hold with probability at least $1 - 2 M e^{-\eta^2}$:
    \begin{equation}
    \begin{split}
        \label{eqn:combined_lr_requirements}
        \alpha_t^{W^{[m]}} 
        \coloneq 
            \min \!
            \left\{
                1 / \Lip(\riskE)
                , \;
                \lrcon{W}
            \right\}
        \\
        \alpha_t^{b^{[m]}} 
        \coloneq 
            \min \!
            \left\{
                1 / \Lip(\riskE)
                , \;
                \lrcon{b}
            \right\}
        ,
    \end{split}
    \end{equation}
    where $\lrcon{W}$ and $\lrcon{b}$ denote the right-hand sides of the equation in the GD LR Decay Conditions~\ref{gdcons:grad}. 
    Moreover, choosing the free parameters $C_{W^{[m]}}, C_{b^{[m]}}$ large enough, GD 
    converges to a stationary point of $\riskHuber$ at a rate of
    \begin{equation*}
        \min_{t = 0,\ldots, T}
        \| 
            \nabla_\theta \riskHuber (\theta_t)
        \| 
        \in 
        \Oc 
            \biggl(
                \frac{1}{\sqrt{T}}
            \biggr)
        .
    \end{equation*}
\end{theorem}

This theorem implies a trade-off between GD convergence and regularity of the learned network. Choosing $T$ sufficiently large allows one to achieve a parameter set yielding an ER within $\varepsilon$-tolerance of a stationary point, then training with our GD LR Decay Conditions ensures Lipschitz regularity throughout the rest of the training schedule.

\section{Experimental Results} \label{s:Experiments}

\begin{figure}[h!]
    \centering
    \begin{subfigure}{0.49\textwidth}
        \centering
        \includegraphics[width=\linewidth]{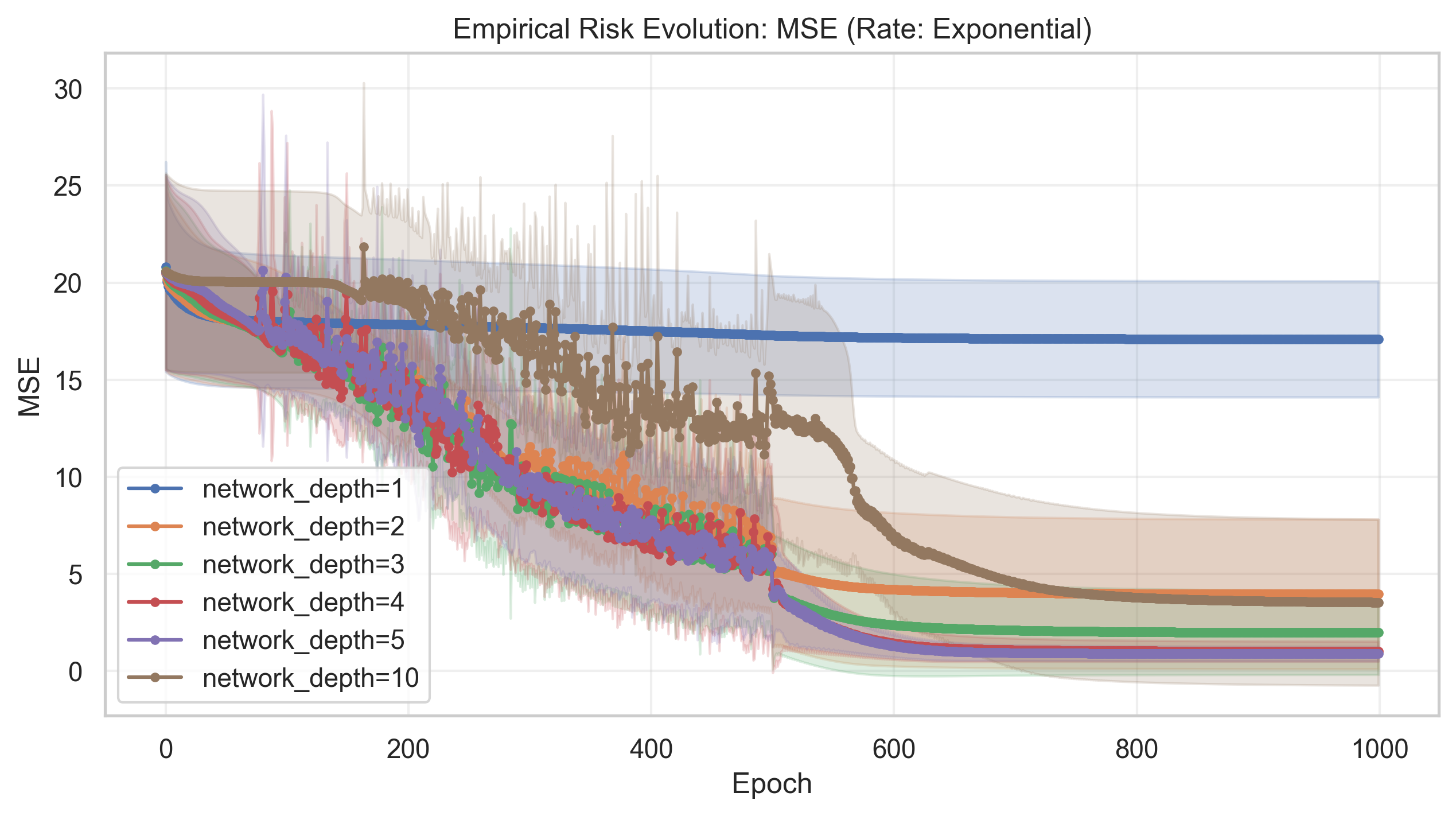}
    \end{subfigure}
    \hfill
    \begin{subfigure}{0.49\textwidth}
        \centering
        \includegraphics[width=\linewidth]{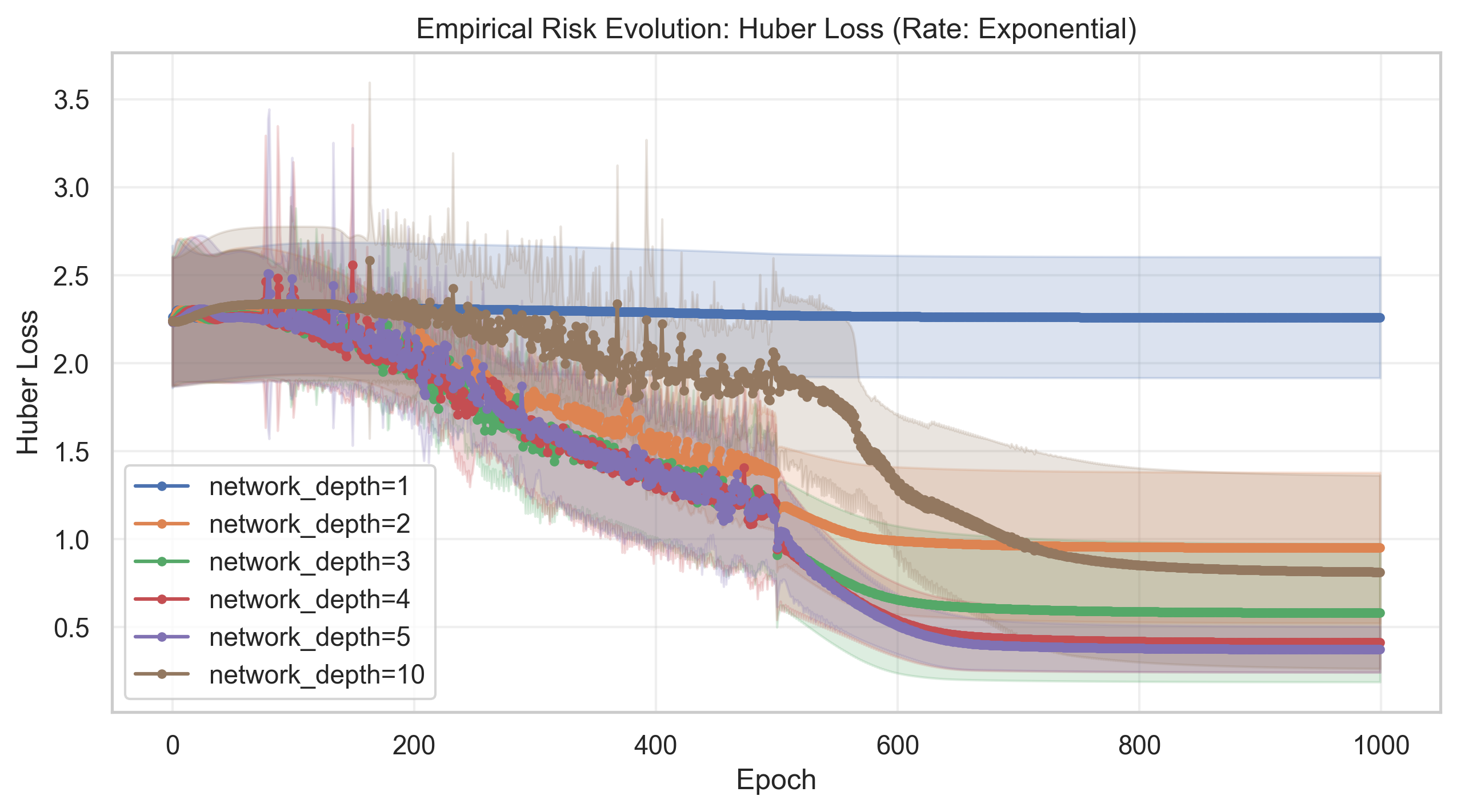}
    \end{subfigure}
    \begin{subfigure}{0.49\textwidth}
        \centering
        \includegraphics[width=\linewidth]{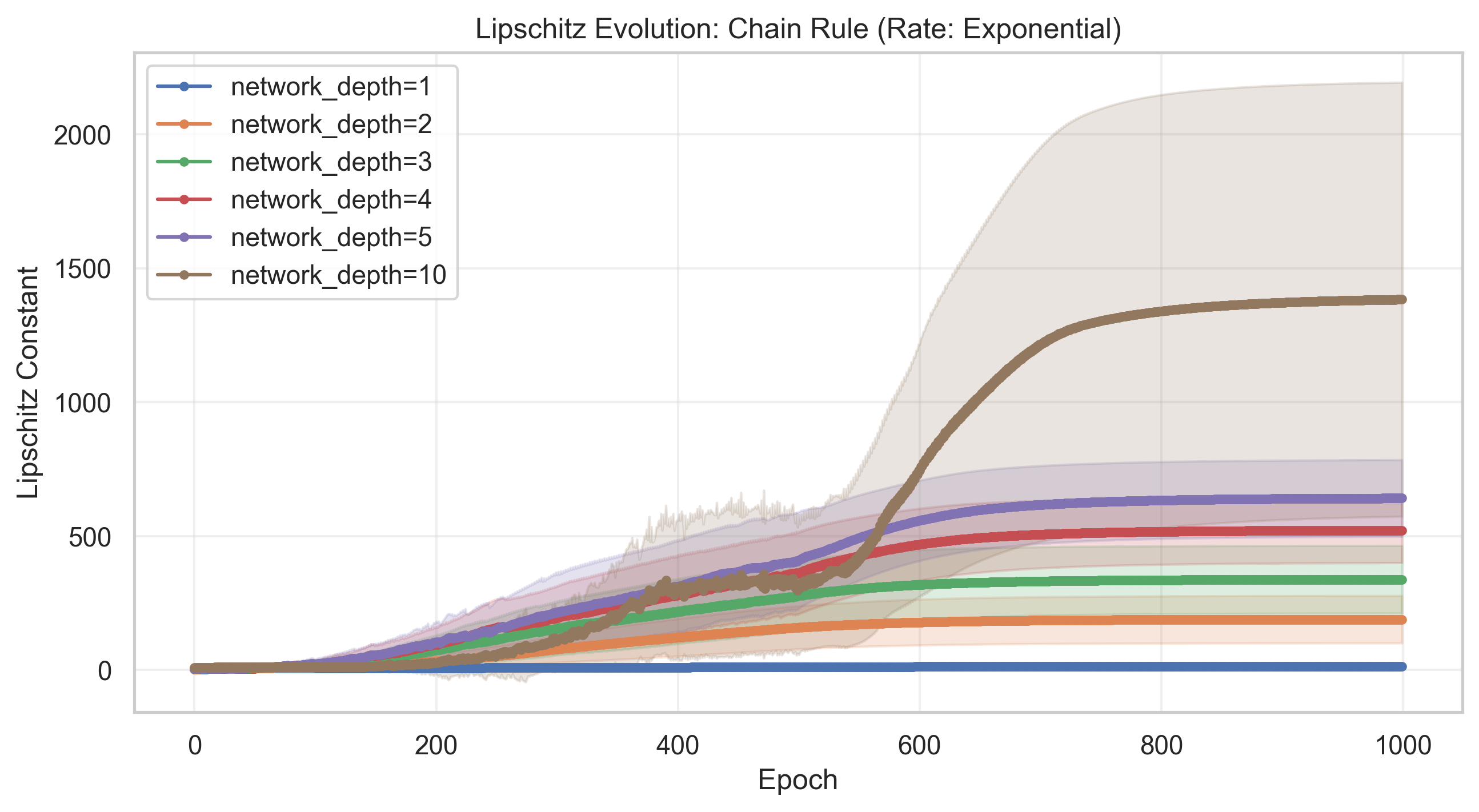}
    \end{subfigure}
    \hfill
    \begin{subfigure}{0.49\textwidth}
        \centering
        \includegraphics[width=\linewidth]{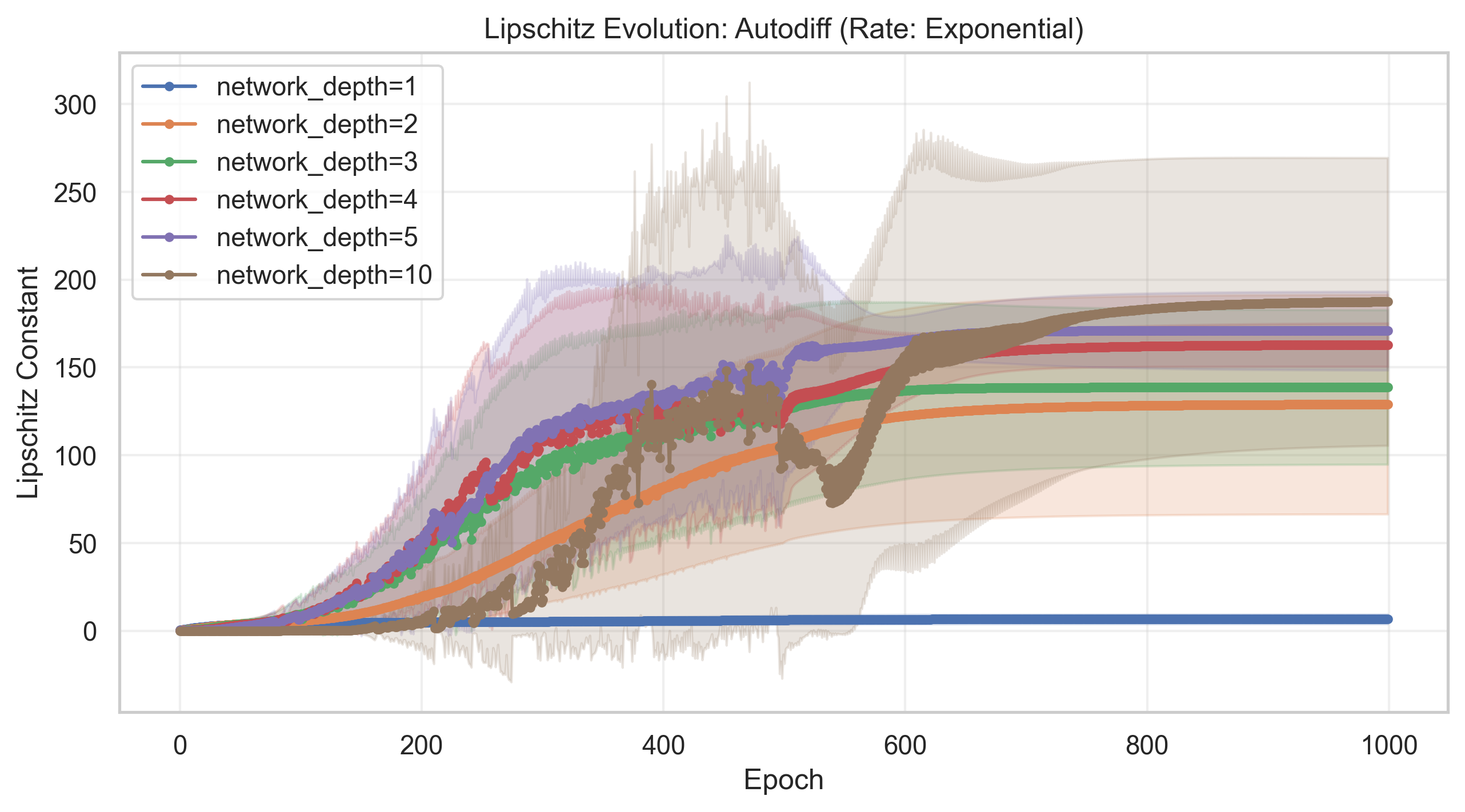}
    \end{subfigure}
    \caption{\textbf{Depth Ablation}: We train MLPs with width 100 and $\tanh$ activation for 1000 epochs on 100 i.i.d. samples of the Forrester function with standard Gaussian noise with noise level $\beta = 0.5$ and initial LR $\alpha = 0.01$, starting our GD LR Decay Conditions at epoch $T=500$ using the exponential rate function (Example~\ref{ex:polynomial_convergence}) with $r=0.03$. We lower bound the Lipschitz constant with 6000 gradient samples. We perform this process 10 times, plotting the mean and one standard deviation.
    }
    \label{fig:experiment:depth_ablation}
\end{figure}
\begin{figure}[h!]
    \centering
    
    \begin{subfigure}{0.49\textwidth}
        \centering
        \includegraphics[width=\linewidth]{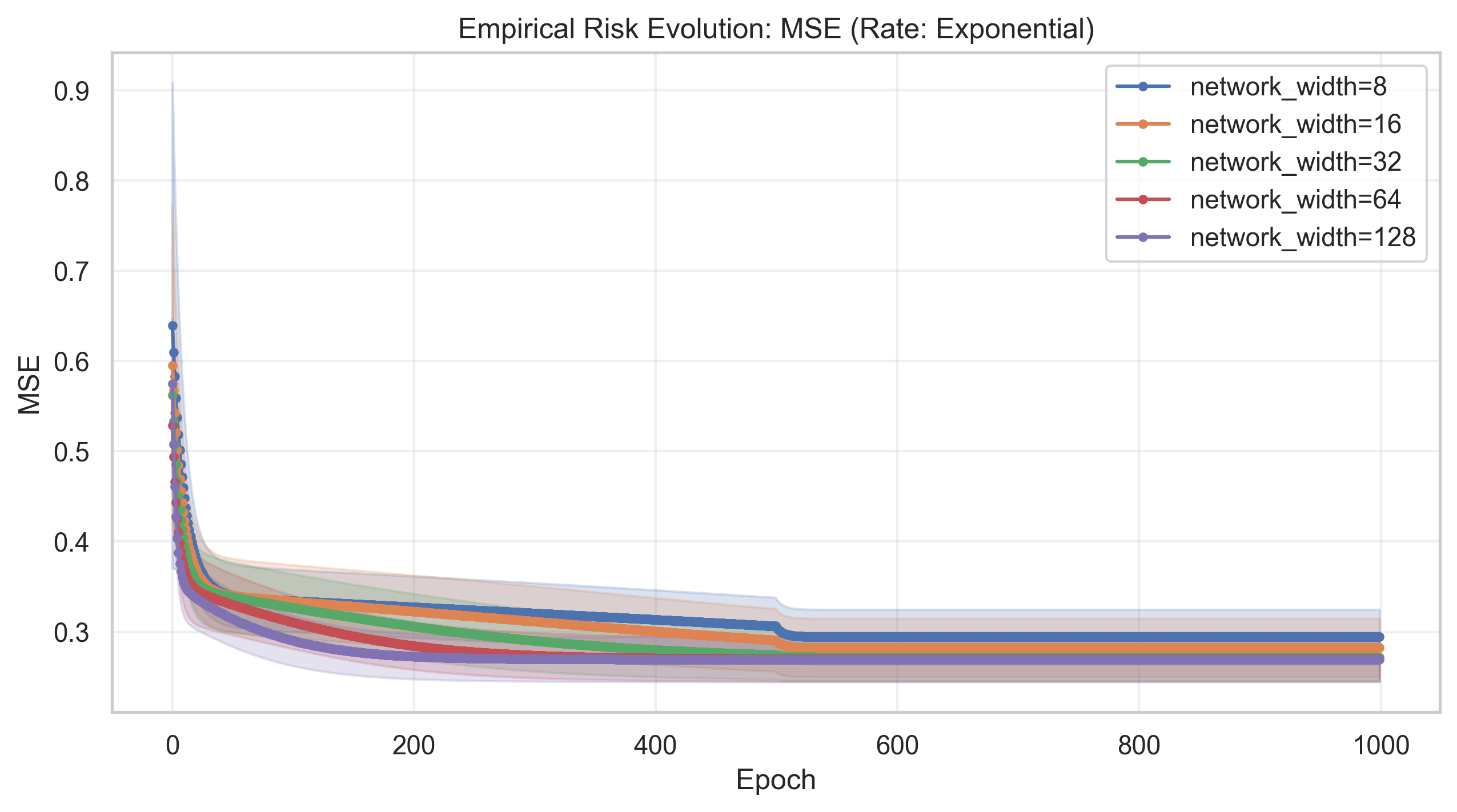}
    \end{subfigure}
    \hfill
    \begin{subfigure}{0.49\textwidth}
        \centering
        \includegraphics[width=\linewidth]{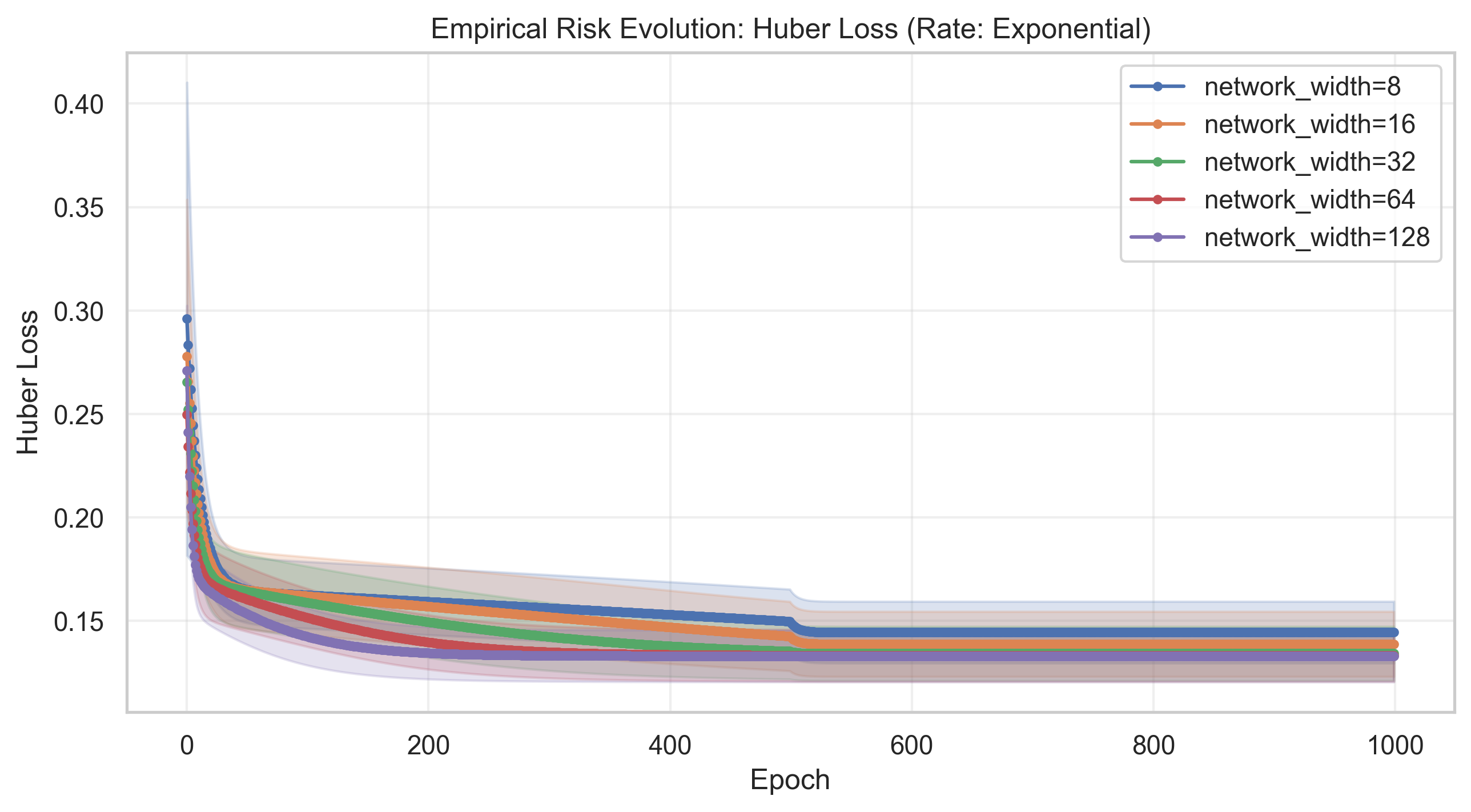}
    \end{subfigure}

    \begin{subfigure}{0.49\textwidth}
        \centering
        \includegraphics[width=\linewidth]{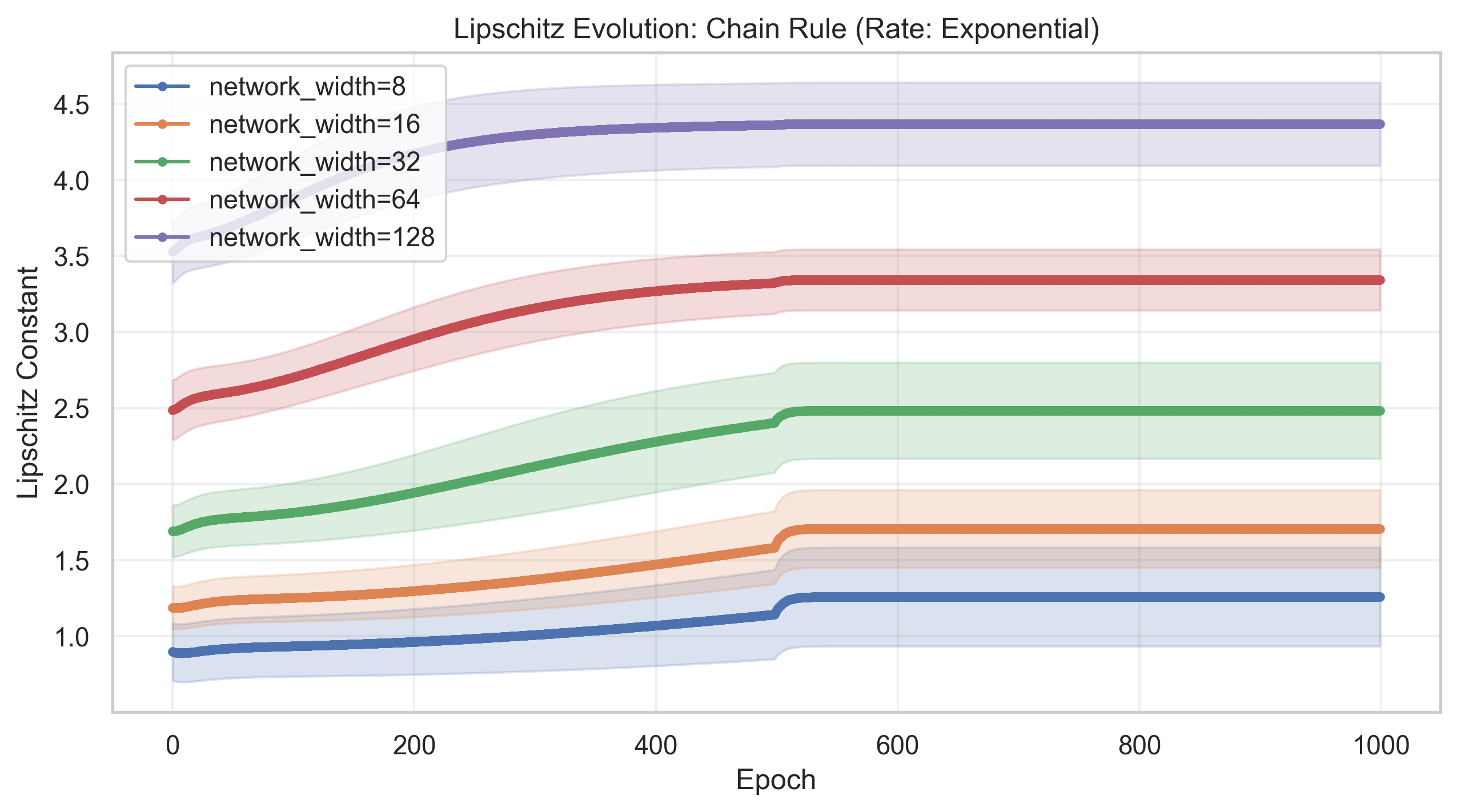}
    \end{subfigure}
    \hfill
    \begin{subfigure}{0.49\textwidth}
        \centering
        \includegraphics[width=\linewidth]{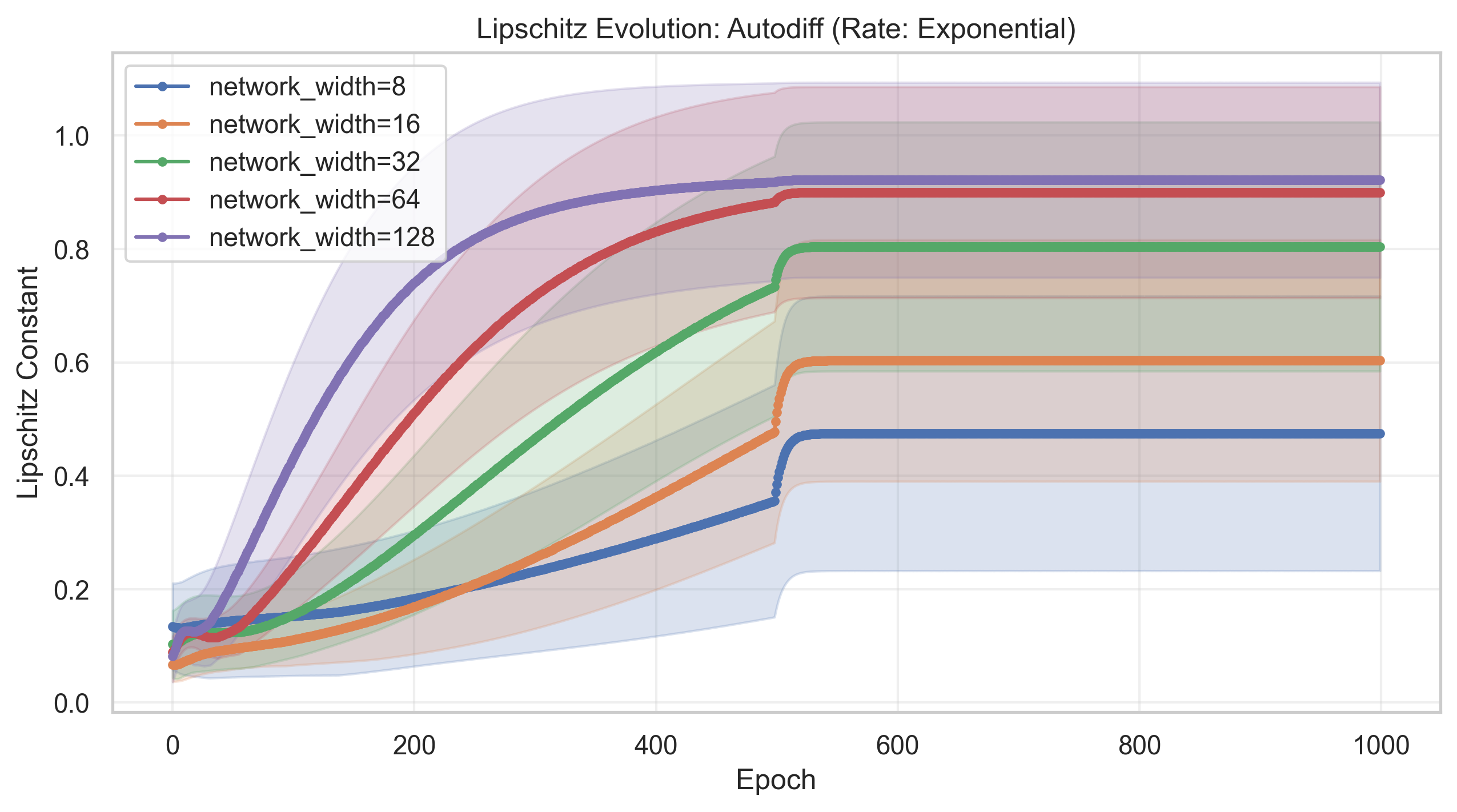}
    \end{subfigure}
    \caption{\textbf{Width Ablation}: We train MLPs with depth 3 and $\tanh$ activation for 1000 epochs on 100 i.i.d. uniform samples of the Franke function with standard Gaussian noise with noise level $\beta = 0.5$ and initial LR $\alpha = 0.01$, starting our GD LR Decay Conditions at epoch $T=500$ using the exponential rate function (Example~\ref{ex:polynomial_convergence}) with $r=0.03$. We lower bound the Lipschitz constant with 6000 gradient samples. We perform this process 10 times, plotting the mean and one standard deviation.}    
    \label{fig:experiment:width_ablation}
\end{figure}

We now investigate the impact of GD LR decay on the performance of MLPs and compare it to standard GD with constant LR.
Each experiment is modeled as a non-parametric regression: fix a target function $h\colon D\subset \R^{d_0}\to \R$ and let $
    Y_n
    =
    h(\X_n) + \beta \varepsilon_n
$
where $\{(\X_n, \varepsilon_n)\}_{n=1}^N$ are i.i.d. random variables with $\X_n \sim \operatorname{Uniform} (\operatorname{domain}(h))$ and each $\varepsilon_n$ is an independent standard Gaussian random variable and given a noise level $\beta \geq 0$. We use several standard benchmark regression functions, outlined in Appendix~\ref{appendix:sec:experiment_details:subsec:target_functions}.

Since evaluating the true Lipschitz constant of an unknown function is NP-hard
\citep{NEURIPS2018_d54e99a6}, we evaluate the Lipschitz constants of our learned models in two ways: (A) we upper bound the Lipschitz constant using the product of the operator norms of the weight matrices, and (B) lower bound the Lipschitz constant by sampling a large number of points and returning the maximum gradient norm at each of the points.

Each experiment is independent and only validates our theory. Namely, that a decaying LR still yields expressive networks (Theorem~\ref{thm:convergence_optimal_gd_rate}) which learn only polynomially dependent on their width, and thus of their number of parameters (Corollary~\ref{cor:generalization_bounds_gd_trained_networks}), and that the GD LR Decay Conditions~\ref{gdcons:grad} indeed yield networks with \emph{relatively} small Lipschitz constants (Theorem~\ref{thm:lip_control_via_lr_decay}).

\begin{figure}[h!]
    \centering
    
    \begin{subfigure}{0.49\textwidth}
        \centering
        \includegraphics[width=\linewidth]{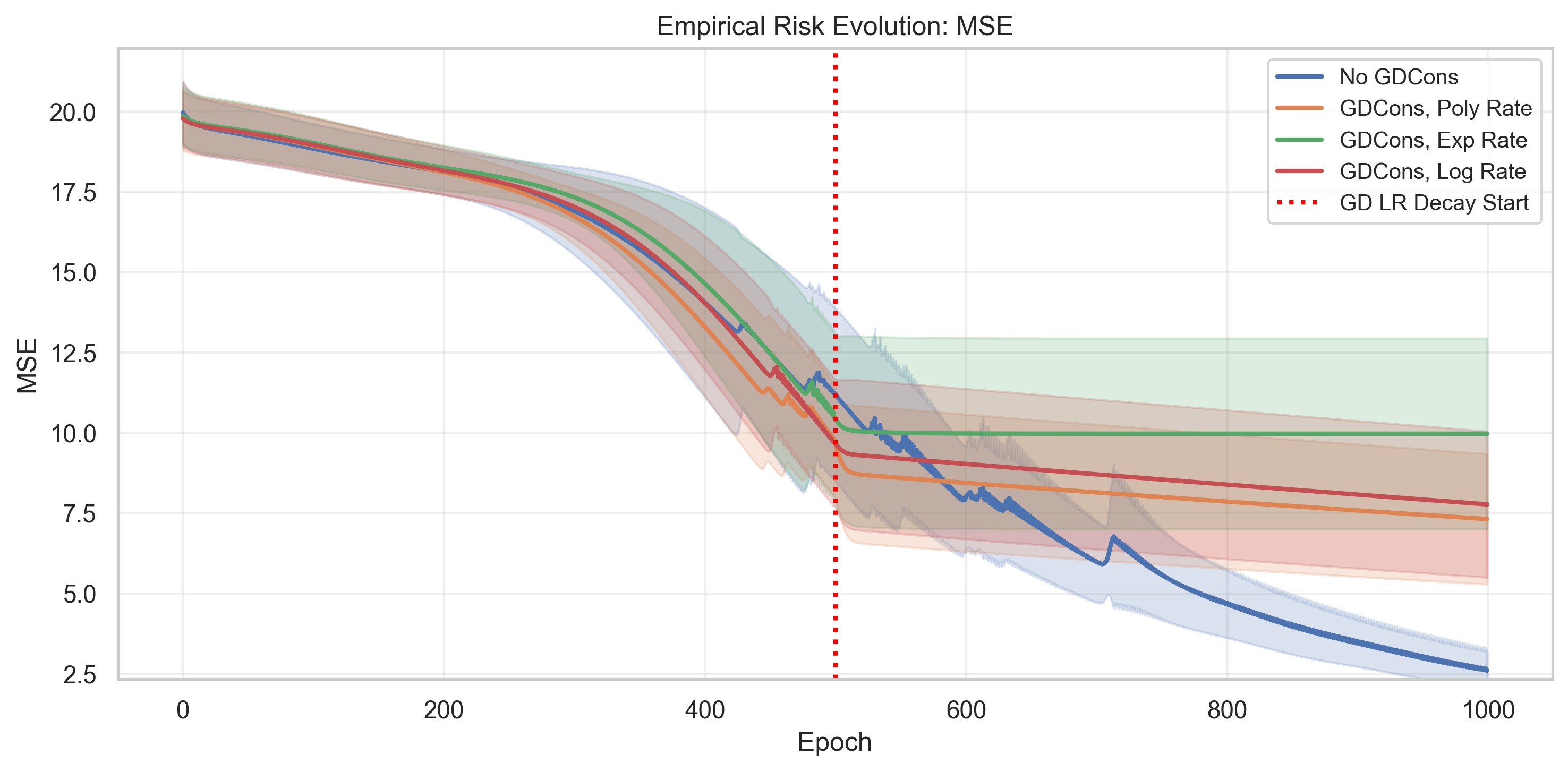}
    \end{subfigure}
    \hfill
    \begin{subfigure}{0.49\textwidth}
        \centering
        \includegraphics[width=\linewidth]{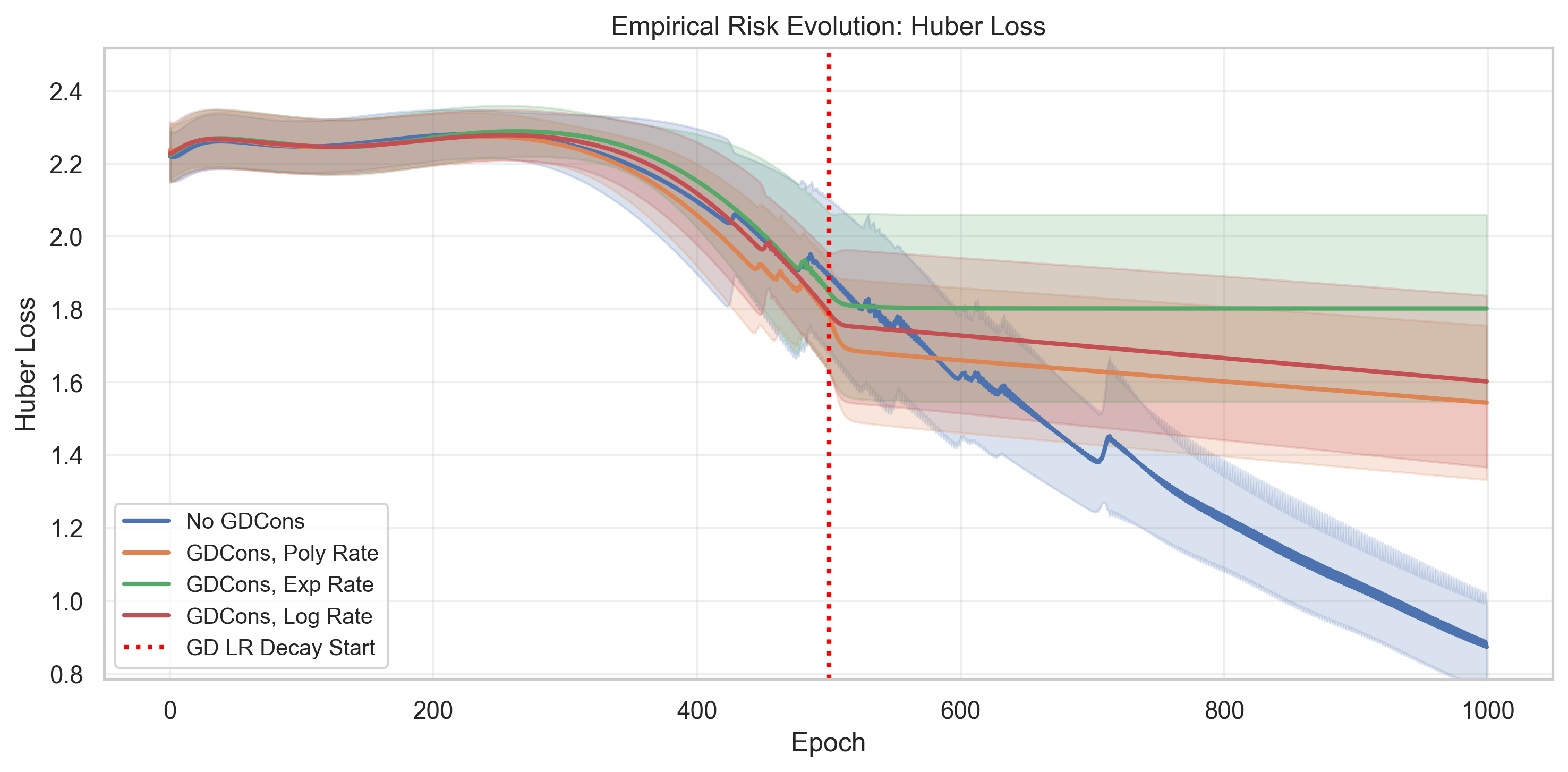}
    \end{subfigure}
    \begin{subfigure}{0.49\textwidth}
        \centering
        \includegraphics[width=\linewidth]{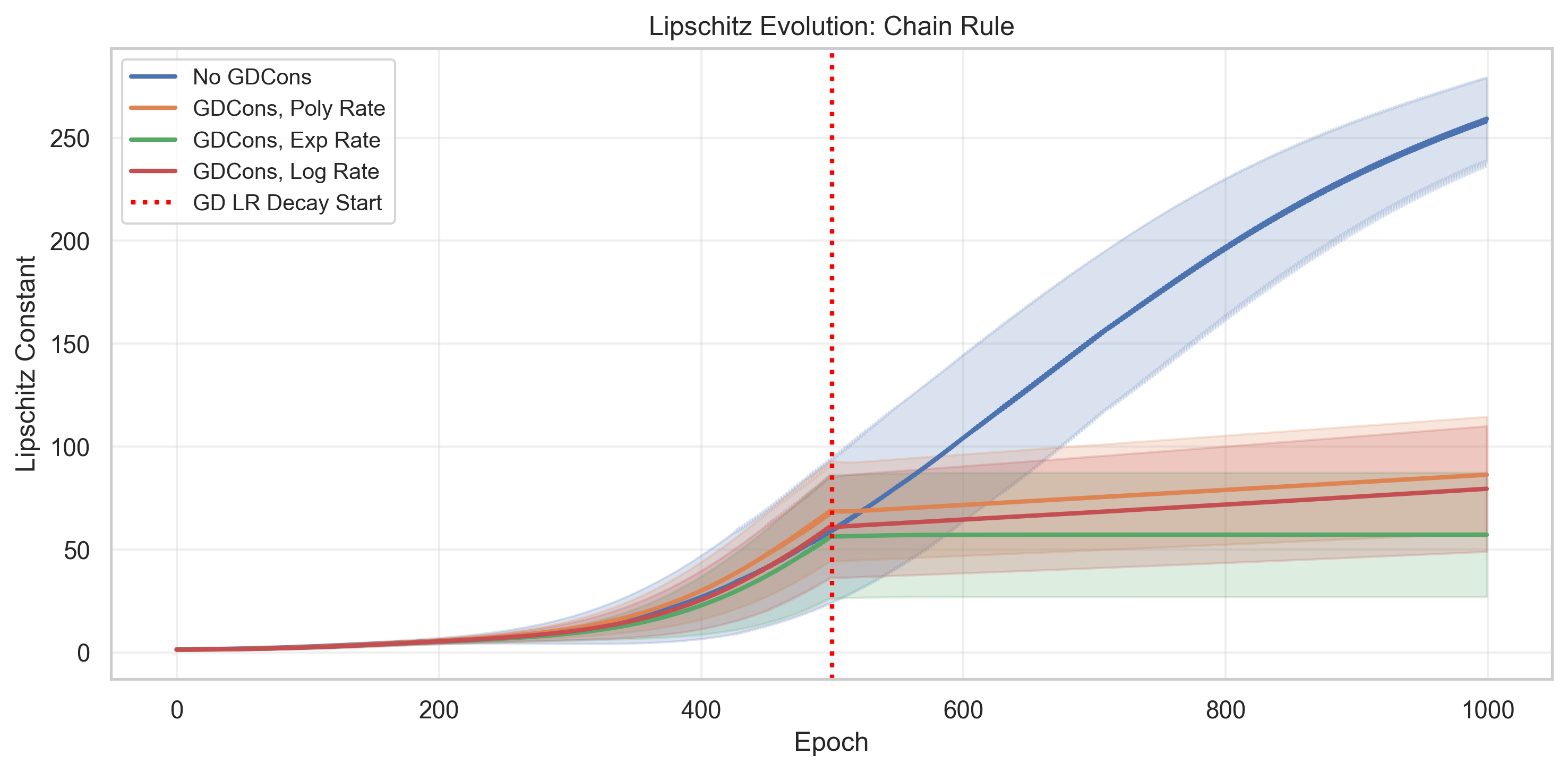}
    \end{subfigure}
    \hfill
    \begin{subfigure}{0.49\textwidth}
        \centering
        \includegraphics[width=\linewidth]{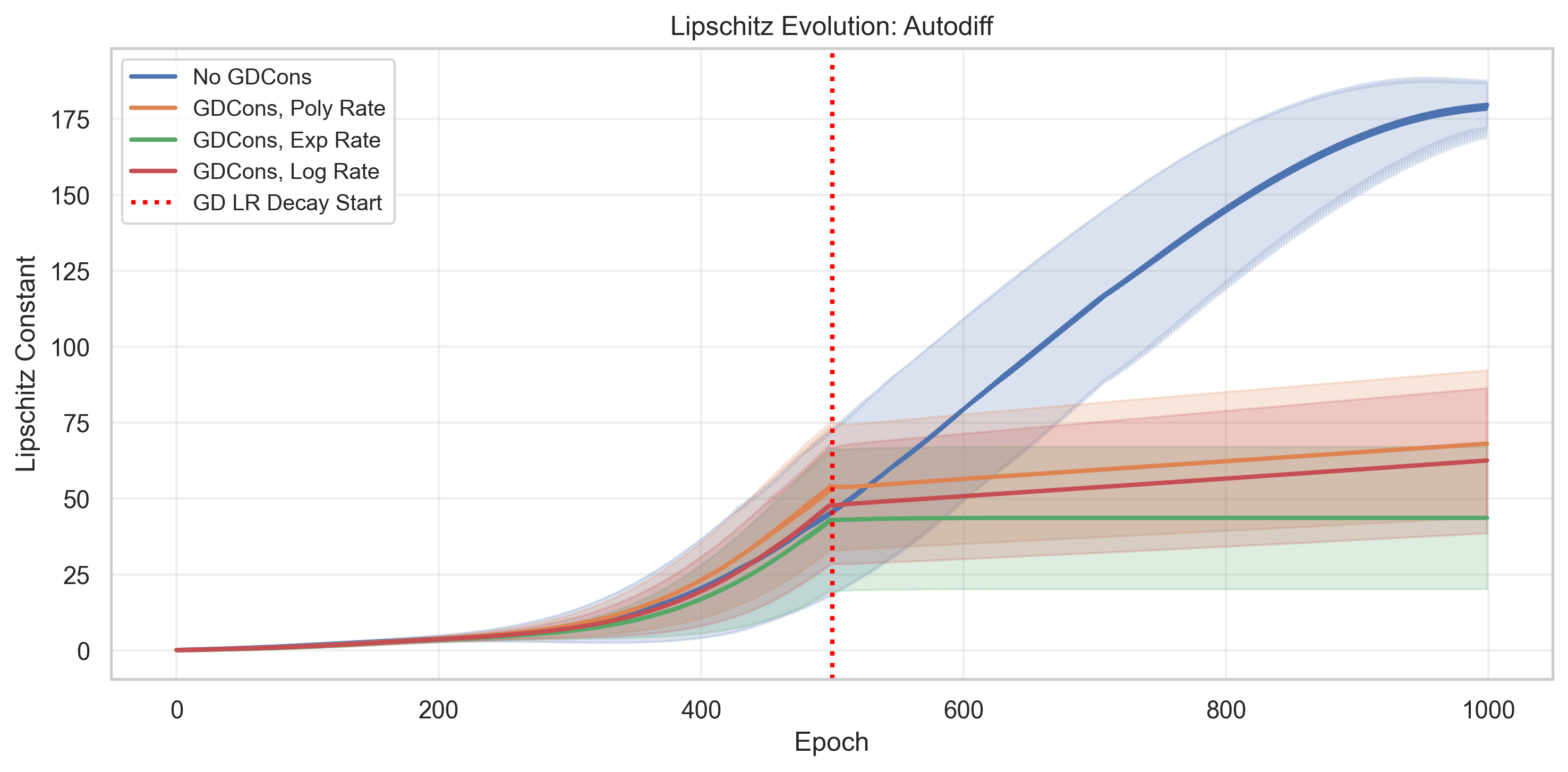}
    \end{subfigure}
    \caption{\textbf{Rate Function Training Dynamics}: We train MLPs with width $16$, depth $3$ and $\operatorname{tanh}$ activation for $1000$ epochs on $2000$ i.i.d. uniform samples of the Forrester function with standard Gaussian noise with noise level $\beta = 0.3$, with initial LR $\alpha = 0.01$, starting our GD LR Decay Conditions at epoch $T=500$ using exponential, polynomial, and logarithmic rate functions with $r = 0.03$. We lower bound the Lipschitz constant with $6000$ gradient samples. We perform this process $10$ times, plotting the mean and standard deviation.}
    \label{fig:experiment:rate_function_training_dynamics}
\end{figure}

\paragraph{Effect of Number of Parameters} We examine the Lipschitz constant evolution during training, varying width (Figure~\ref{fig:experiment:width_ablation})  and depth (Figure~\ref{fig:experiment:depth_ablation}). As predicted by Theorem~\ref{thm:lip_control_via_lr_decay}, the Lipschitz constant 
grows at most polynomially with network width and exponentially with network depth.

\paragraph{Effect of Sample Size} We examine the Lipschitz constant evolution during training, varying sample size (Figure~\ref{fig:experiment:sample_size_ablation}), which appears in the right-hand side of Theorem~\ref{thm:lip_control_via_lr_decay} and Corollary~\ref{cor:generalization_bounds_gd_trained_networks}. MLPs trained on varying sample sizes appear to exhibit similar learning and regularity properties.

\paragraph{Effect of Rate Function} Varying rate functions affects the convergence of the ERM. Critically, this allows one to customize the trade off between regularity of the learned function and the convergence of the ERM.

\section{Conclusion and Future Work}

We showed that training neural networks with an eventual decay in the learning rate enforces strong Lipschitz regularity (Theorem~\ref{thm:lip_control_via_lr_decay}), without affecting the convergence rate of the empirical risk (measured with Huber loss) to a stationary point (Theorem~\ref{thm:convergence_optimal_gd_rate}). Additionally, we derived generalization bounds that are dependent \emph{polynomially} on network width and \emph{exponentially} on network depth (Corollary~\ref{cor:generalization_bounds_gd_trained_networks}), suggesting that overparameterization does not degrade the statistical performance of these networks. Our experiments also indicate that networks trained with a decaying learning rate exhibit similar regularity properties to those trained with a constant learning rate, suggesting that gradient descent may naturally promote regularity in neural networks.

In future work, we aim to explore GD-based algorithms with more complex dynamics, such as Conjugate Gradients or Heavy-Ball/momentum methods, or any GD-type method considered by~\citet{velikanov2024tight}. Stochastic variants and proximal splitting-type extensions could also be investigated. One could also expand these results to other neural network architectures, or generalize to a regime assuming only local uniform continuity of the realization map. However, we expect the core message to remain unchanged: a small learning rate decay can ensure that GD-type algorithms minimize empirical risk while yielding neural networks with high Lipschitz regularity, with high probability.

\section*{Acknowledgements}

K.\ Sung was supported by the Natural Sciences and Engineering Research Council of Canada (NSERC) through two Undergraduate Student Research Awards (USRAs) in 2024 and 2025.

A.\ Kratsios acknowledges the financial support from an NSERC Discovery Grant No.\ RGPIN-2023-04482 and No.\ DGECR-2023-00230.  A.\ Kratsios was also supported by the project Bando PRIN 2022 named ``Qnt4Green - Quantitative Approaches for Green Bond Market: Risk Assessment, Agency Problems and Policy Incentives'', codice 2022JRY7EF, CUP E53D23006330006, funded by European Union – NextGenerationEU, M4c2.
A.\ Kratsios also acknowledges that resources used in preparing this research were provided, in part, by the Province of Ontario, the Government of Canada through CIFAR, and companies sponsoring the Vector Institute\footnote{\href{https://vectorinstitute.ai/partnerships/current-partners/}{https://vectorinstitute.ai/partnerships/current-partners/}}. N.\ Forman was supported by NSERC Discovery Grant No.\ RGPIN-2020-06907.

\section*{Author Contributions}

K. Sung~developed the main technical lemmata and completed the generalization of the main theoretical results from two-layer to $M$-layer networks. K. Sung~also contributed to writing and editing the manuscript, and performed the majority of numerical experiments including MLP implementation and training.

A. Kratsios~posed the problem, outlined the solution, and produced the main theoretical results for two-layer networks. They linked Lipschitz regularity to statistical guarantees for the trained networks, and proved the generalization bounds.
A. Kratsios~also contributed substantially to writing and editing the manuscript, the literature review, and supervision.

K. Khalil~and S. Samu~supported the implementation of the ML pipelines.

N. Forman~provided extensive support and guidance in supervision of the project.
N. Forman~helped edit the manuscript, conceptualized the Lipschitz estimation algorithms, and informed research directions.

\bibliographystyle{unsrtnat}
\bibliography{references}

\newpage
\appendix
\onecolumn

\section{Proofs} \label{appendix:proofs}

In Setting~\ref{setting:main}, we introduce further notation to refer to the hidden layers of a multilayer perceptron. Let $\mathbf{a}$ and $\mathbf{z}$ denote respectively the propogated affine maps and activated affine maps, that is, \begin{equation*}
    \begin{split}
        \vec{a}_{n}^{[m]} &\coloneq W^{[m]} \vec{z}_n^{[m-1]} + b^{[m]}
        \\
        \vec{z}_n^{[m]} &\coloneq \sigma \bullet ( \vec{a}_n^{[m]} )
        \\
        \vec{z}_n^{[0]} &\coloneq \x_n
        \\
        \vec{a}_n^{[M]} &\coloneq f_\theta (\x_n)
        .
    \end{split}
\end{equation*}
Then, $f_\theta \colon \R^{d_0}\to \R$ is a multilayer perceptron given by \[
    f_\theta \colon (
        \x = \z^{[0]} 
        \mapsto
        \a^{[1]}
        \mapsto
        \z^{[1]}
        \mapsto
        \cdots
        \mapsto
        \z^{[M-1]}
        \mapsto
        \a^{[M]}
    )
    .
\]

Following \citet[Lemma 1]{naumov2017feedforwardrecurrentneuralnetworks}, we produce explicit expressions for the gradient of the loss function with respect to each parameters, modified to suit a regression problem by removing the activation function from the output layer.

\begin{lemma}[Explicit GD Formulations]
    \label{lem:explicit_gd_formulations}
    In Setting~\ref{setting:main}, let $f_\theta \colon \R^{d_0} \to \R$ be a multilayer perceptron (Definition~\ref{def:MLP}) trained by gradient descent in~Equation~\eqref{eqn:gd_mlp_defn}. 
    Then, the update rule is
    \begin{equation}
        \notag
        \begin{alignedat}{3}
            W_t^{[m]} &= 
                W_{t-1}^{[m]} 
                && -  \alpha_t^{W^{[m]}} 
                \frac{2}{N} \sum \limits_{n=1}^N \vec{v}_n^{[m]} (\vec{z}_n^{[m-1]})^\top
            \\
            b_t^{[m]} &= 
                b_{t-1}^{[m]} 
                && - \alpha_t^{b^{[m]}} 
                \frac{2}{N} \sum \limits_{n=1}^N \vec{v}_n^{[m]},
        \end{alignedat}
    \end{equation}
    where 
    \begin{equation}
        \notag
        \begin{split}
            \vec{v}_n^{[M]} &\coloneq  2 ( f_\theta(\x_n) - y_n )
            \\
            \vec{v}_n^{[m]} &\coloneq ( (W_{t-1}^{[m + 1]} )^\top \vec{v}_n^{[m + 1]}) \odot \sigma' \bullet ( \vec{a}_n^{[m]} ), \quad 1 \leq m < M
        \end{split}
    \end{equation}
    and where $ \odot $ denotes the Hadamard (elementwise) product and $\bullet$ denotes componentwise composition.
\end{lemma}

\begin{proof}
    In the setting of this proof, we let subscripts with parentheses denote the components of a vector or matrix. Superscripts with brackets will continue to denote the layers of a multilayer perceptron.

    We may rewrite the samplewise empirical risk with respect to the single sample $(\vec{x}_n, y_n) \in \dataset$ as \[
        \riskSample (\theta)
        =
        \lVert f_\theta(\x_n) - y_n \|^2
        .
    \]
    Taking the partial derivative of the empirical risk with respect to a single weight matrix parameter, we have by the chain rule:
    \begin{align*}
        \pdv{\riskSample}{w_{(i,j)}^{[m]}} 
        &= \left( \pdv{\riskSample}{z_{(i)}^{[m]}} \right) \left( \pdv{z_{(i)}^{[m]}}{a_{(i)}^{[m]}} \right) \left( \pdv{a_{(i)}^{[m]}}{w_{(i,j)}^{[m]}} \right)
        \\&= \left( \pdv{ \riskSample }{ z_{(i)}^{[m]}} \right) \sigma ' \bullet (a_{(i)}^{[m]}) z_{(j)}^{[m-1]}
        \\&= v_{(i)}^{[m]} z_{(j)}^{[m-1]},
    \end{align*} 
    where $\displaystyle v_{(i)}^{[m]} \coloneq \left( \pdv{ \riskSample }{ z_{(i)}^{[m]}} \right) \sigma ' \bullet (a_{(i)}^{[m]})$.
    
    Observe that in the case of the output layer $m=M$, we have 
    \begin{align*}
        \pdv{ \riskSample }{ z_{i}^{[M]}} 
        = 
        2 ( f_\theta(\x_n) - y_n ) , 
    \end{align*}
    while for the hidden layers, using chain rule, we have
    \begin{align}
    \pdv{ \riskSample }{ z_{(i)}^{[m]}} &= 
        \sum_{j=1}^{n} \left( \pdv{ \riskSample }{ z_{(j)}^{[m + 1]}} \right) \sigma ' \bullet (a_{(j)}^{[m + 1]}) w_{(j,i)}^{[m + 1]} \label{gradient_z_hidden_layer} \notag
    \\
    &= 
        \sum_{j=1}^{n} v_{(j)}^{[m + 1]} w_{(j,i)}^{[m + 1]} \notag
    .
    \end{align}
    Finally, assembling the indices $i$ and $j$ into a vector and matrix form we obtain the matrix expression for the gradient $\displaystyle \pdv{\riskSample}{W^{[m]}}$ which appears on the right-hand side of the gradient descent expression. The derivation for the bias vector is similar.
\end{proof}

\begin{lemma}[Relating $L_p$ Norm Bounds]
    \label{lem:norm_bounds_tight_lp}
    For $1 \leq j \leq k \leq \infty$ and $\vec{v} \in \R^d$, we have \[
        \| \vec{v} \|_k
        \leq
        \| \vec{v} \|_j 
        \leq 
        d^{\frac{1}{j} - \frac{1}{k}} \|\vec{v}\|_k.
    \]
\end{lemma}

\begin{lemma}[Control on the Growth Rate of the Norm of GD-Trained Weight Matrices]
    \label{lem:control_growth_norm_gd_trained_weight_matrices:implicit}
    In Setting~\ref{setting:main}, if the gradient descent process $\boldsymbol{\theta}$ with variable learning rate $\boldsymbol{\alpha}$ satisfies the GD LR Decay Conditions~\ref{gdcons:grad}, then
    \[
        \| W_{t}^{[1]}\| ,
        \| b_{t}^{[1]}\| ,
        \ldots, \| W_{t}^{[M]}\| ,
        \| b_{t}^{[M]}\|
        \lesssim
        G(t)
        .
    \]
\end{lemma}
\begin{proof}
    Setting
    \[
        \begin{aligned}
            \gamma_t^{W^{[m]}} &\coloneq \alpha_t^{W^{[m]}} \sum\limits_{n=1}^{N}  \left\| (\nabla_{W^{[m]}} f_{\theta_{t-1}}(\x_n))^\top  \Big [f_{\theta_{t-1}}(\x_n) - y_n \Big ] \right\|_2
            \\
            \gamma_t^{b^{[m]}} &\coloneq \alpha_t^{b^{[m]}} \sum\limits_{n=1}^{N}  \left\| (\nabla_{b^{[m]}} f_{\theta_{t-1}}(\x_n))^\top  \Big [f_{\theta_{t-1}}(\x_n) - y_n \Big ] \right\|_2
        \end{aligned}
    \]
    allows us to restate our GD LR Decay Conditions~\ref{gdcons:grad} as
    \begin{equation}
        \label{eqn:lem:control_norm_growth_gd_weights:implicit:gamma}
        \begin{aligned}
            \gamma_t^{W^{[m]}} &\leq C_{W^{[m]}} g(t)
            \\
            \gamma_t^{b^{[m]}} &\leq C_{b^{[m]}} g(t).
        \end{aligned}
    \end{equation}
    Taking the norm on both sides of the gradient descent process in Equation~\eqref{eqn:gd_mlp_defn}, we have
    \begin{equation}
        \label{eqn:lem:control_norm_growth_gd_weights:implicit:gd}
        \begin{aligned}
            \|W_t^{[m]}\| &\leq \|W_{t-1}^{[m]}\| + \frac{2\alpha_t^{W^{[m]}}}{N} \sum\limits_{n=1}^{N} \left\| (\nabla_{W^{[m]}} f_{\theta_{t-1}}(\x_n))^\top  \Big [f_{\theta_{t-1}}(\x_n) - y_n \Big ] \right\|_2
            \\
            \|b_t^{[m]}\| &\leq \|b_{t-1}^{[m]}\| + \frac{2\alpha_t^{b^{[m]}}}{N} \sum\limits_{n=1}^{N} \left\| (\nabla_{b^{[m]}} f_{\theta_{t-1}}(\x_n))^\top  \Big [f_{\theta_{t-1}}(\x_n) - y_n \Big ] \right\|_2.
        \end{aligned}
    \end{equation}
    Combining Equation~\eqref{eqn:lem:control_norm_growth_gd_weights:implicit:gamma} and Equation~\eqref{eqn:lem:control_norm_growth_gd_weights:implicit:gd}, we have
    \begin{equation*}
        \begin{aligned}
            \|W_t^{[m]}\| &\leq \|W_{t-1}^{[m]}\| + \frac{2\gamma_t^{W^{[m]}}}{N}
            \\
            \|b_t^{[m]}\| &\leq \|b_{t-1}^{[m]}\| + \frac{2\gamma_t^{b^{[m]}}}{N}
            .
        \end{aligned}
    \end{equation*}
    By repeatedly applying the triangle inequality, we have
    \begin{equation*}
        \begin{aligned}
            \|W_t^{[m]}\| &\leq \|W_{0}^{[m]}\| + \frac{2}{N} \sum\limits_{s=1}^t \gamma_s^{W^{[m]}}
            \\
            \|b_t^{[m]}\| &\leq \|b_{0}^{[m]}\| + \frac{2}{N} \sum\limits_{s=1}^t \gamma_s^{b^{[m]}}
            .
        \end{aligned}
    \end{equation*}
    Applying the assumptions on $\gamma_1^{W^{[m]}},\gamma_1^{b^{[m]}},\ldots,\gamma_t^{W^{[m]}},\gamma_t^{b^{[m]}}$ and the rate function $G$, we have
    \begin{equation*}
        \begin{alignedat}{3}
            \|W_t^{[m]}\| &\leq \|W_0^{[m]}\| + \frac{2}{N} \sum\limits_{s=1}^t C_{W^{[m]}} g(s) &&\leq \|W_0^{[m]}\| + \frac{2}{N} C_{W^{[m]}} (G(t) + g(1))
            \\
            \|b_t^{[m]}\| &\leq \|b_0^{[m]}\| + \frac{2}{N} \sum\limits_{s=1}^t C_{b^{[m]}} g(s) &&\leq \|b_0^{[m]}\| + \frac{2}{N} C_{b^{[m]}} (G(t) + g(1))
            .
        \end{alignedat}
    \end{equation*}
    Thus, we have $\| W_{t}^{[1]}\| ,
        \| b_{t}^{[1]}\| ,
        \ldots, \| W_{t}^{[M]}\| ,
        \| b_{t}^{[M]}\|
        \lesssim
        G(t).$
\end{proof}

Having bounded the norms of the weight matrices, we can derive a bound on the maximum norms on the parameter space the gradient descent process, and thus ensure that each of the parameters remain in the $w$-cube in $\mathbb{R}^P$.

\begin{lemma}[Deterministic Bounds on the Parameter Space]
    \label{lem:deterministic_bounds_on_param_space}
    In Setting~\ref{setting:main}, there exists $w > 0$ such that $\theta_t \in \Theta_w$ for every $t\in \N_+$.
\end{lemma}
\begin{proof}
    By Lemma~\ref{lem:control_growth_norm_gd_trained_weight_matrices:implicit}, we have that:
    \begin{equation}
        \notag
        \begin{aligned}
            \| W_t^{[m]} \| &\leq \| W_0^{[m]} \| + \frac{2}{N} C_{W^{[m]}} (G(t) + g(1))
            \\
            \| b_t^{[m]} \| &\leq \| b_0^{[m]} \| + \frac{2}{N} C_{b^{[m]}} (G(t) + g(1))
            .
        \end{aligned}
    \end{equation}
    By Lemma~\ref{lem:norm_bounds_tight_lp}, for an arbitrary matrix $A \in \R^{d_{m} \times d_{m-1}}$, \[
        \| A \|_2 \leq \| A \|_F = \| \operatorname{flatten}(A) \|_2\leq \sqrt{d_m d_{m-1}}\| \operatorname{flatten}(A) \|_\infty ,
    \]
    and thus we can modify our bounds
        \begin{equation}
        \label{eqn:deterministic_bounds_on_param_space_FINAL}
        \begin{aligned}
            \| W_t^{[m]} \| &\leq \sqrt{d_md_{m-1}}\| \operatorname{flatten}(W_0^{[m]}) \|_\infty + \frac{2}{N} C_{W^{[m]}} (G(t) + g(1))
            \\
            \| b_t^{[m]} \| &\leq \sqrt{d_m}\| b_0^{[m]} \|_\infty + \frac{2}{N} C_{b^{[m]}} (G(t) + g(1))
            .
        \end{aligned}
    \end{equation}
    Setting $w$ equal to the maximum value of the expressions on the right-hand side of Equation~\eqref{eqn:deterministic_bounds_on_param_space_FINAL} for each $1 \leq m \leq M$ concludes the proof.
\end{proof}

For completeness, we record the following simple lemma. It essentially states that the composition of Lipschitz functions is again Lipschitz.

\begin{lemma}[MLP Lipschitz Bounds with Chain Rule]
    \label{lem:mlp_lip_bounds_via_chain_rule}
    Let $f_\theta$ be a multilayer perceptron as in Definition~\ref{def:MLP}. Then, 
    \begin{equation}
        \notag
        \Lip(f_\theta) 
        \leq 
        \lVert W^{[1]} \rVert_2 \cdots \lVert W^{[M]} \rVert_2 \Lip(\sigma)^{M-1}
        .
    \end{equation}
\end{lemma}
\begin{proof}
    This follows directly from the chain rule.
\end{proof}

\begin{lemma}
    \label{lem:deterministic_mlp_lip_bound_weight_matrices}
    In Setting~\ref{setting:main}, if the gradient descent process $\boldsymbol\theta$ with variable learning rate $\boldsymbol{\alpha}$ satisfies the GD LR Decay Conditions~\ref{gdcons:grad}, then
    \begin{equation}
        \notag
        \Lip(f_{\theta_t}) \leq \Lip(\sigma)^{M-1} \prod \limits_{m=1}^M \left[\| W_0^{[m]} \| + \frac{2C_{W^{[m]}}}{N} (G(t) + g(1)) \right]
        .
    \end{equation}
\end{lemma}
\begin{proof}
    By Lemma~\ref{lem:control_growth_norm_gd_trained_weight_matrices:implicit}, we have that $\| W_t^{[1]}\| , \ldots , \| W_t^{[M]}\| \lesssim G(t)$. Using the Lipschitz constant from
    Lemma~\ref{lem:mlp_lip_bounds_via_chain_rule} completes the proof.\footnote{One could improve this bound by using the Lipschitz bounds from \citet{JMLR:v24:22-1381} or \citet{herrera2023locallipschitzboundsdeep}. Our paper illustrates that learning rate constraints during training can prescribe the Lipschitz regularity of neural networks.} 
\end{proof}

Having gained control over the growth rate of the norms of the GD-trained weight matrices defining our
neural network, we may directly deduce an upper-bound for the Lipschitz constant of the function realized
by those trained neural network parameters. Together, Lemma~\ref{lem:deterministic_mlp_lip_bound_weight_matrices} and some results from random matrix
theory yield our main theorem.

\subsection{\texorpdfstring{Proof of Theorem~\ref{thm:lip_control_via_lr_decay}}{Proof of Theorem 1}}

\begin{proof}
    Recall that the operator norm of a matrix coincides with its maximal singular value.

    By our assumptions that the rescaled initialized weight matrices have isotropic sub-Gaussian rows in  Assumption~\ref{assumption:standard_param_initialization}, for each $m = 1, \ldots, M$, we let $\tilde W_0^{[m]} \coloneq \sqrt{d_{m-1}} W_0^{[m]}$ be an isotropic matrix.
    
    Then, we apply the version of Gordon's Majorization Theorem~\citep[Theorem 4.6.1]{VershyyninBook} and a union bound to deduce that there exists a constant $\kappa > 0$ depending only on the law of the first row of $\tilde{W}_0^{[1]},\ldots, \tilde{W}_0^{[M]}$ such that for every $\eta > 0$,
    \begin{align}
        &\quad\:\,
        \Pr 
        \left(
            \bigcap_{m=1}^M 
            \left\{ 
                \| \tilde{W}_0^{[m]}\| 
                \leq
                \sqrt{d_m} + \kappa 
                \left[
                    \sqrt{d_{m-1}} + \eta
                \right]
            \right\} 
        \right) \label{eqn:RM_Concentration_multilayer___BEGIN}
        \\
        &\geq 1 - \sum_{m=1}^M \Pr
            \left(
                \| \tilde{W}_0^{[m]}\| 
                \leq
                \sqrt{d_m} + \kappa 
                \left[
                    \sqrt{d_{m-1}} + \eta
                \right]
            \right) \notag
        \\
        &\geq 1 - \sum_{m=1}^M 2e^{-\eta^2} \notag
        \\
        &\geq 1 - 2M e^{-\eta^2} \label{eqn:RM_Concentration_multilayer___END}
        .
    \end{align}
    
    Since our assumptions place us squarely in the scope of Lemma~\ref{lem:deterministic_mlp_lip_bound_weight_matrices}, then combining the estimate in Equation~\eqref{eqn:RM_Concentration_multilayer___BEGIN}--\eqref{eqn:RM_Concentration_multilayer___END} with the estimate in Lemma~\ref{lem:deterministic_mlp_lip_bound_weight_matrices} we deduce that for every $\eta > 0$, the following holds with probability at least $1 - 2Me^{-\eta^2}$:
    \begin{equation*}
        \begin{aligned}
            \Lip( f_{\theta_t} )
            &\leq 
                \Lip(\sigma)^{M-1} \prod \limits_{m=1}^M \left[\| W_0^{[m]} \| + \frac{2C_{W^{[m]}}}{N} (G(t) + g(1)) \right]
            \\&\leq 
                \Lip(\sigma)^{M-1} \prod \limits_{m=1}^M \left[ \frac{\sqrt{d_m}}{\sqrt{d_{m-1}}} + 
                \kappa 
                \left[
                    1 + \dfrac{\eta}{\sqrt{
                        d_{m-1} 
                    }}
                \right]
                + \frac{2C_{W^{[m]}}}{N} (G(t) + g(1)) \right]
            ,
        \end{aligned}
    \end{equation*}
    as desired.
\end{proof}

\subsection{\texorpdfstring{Proof of Corollary~\ref{cor:generalization_bounds_gd_trained_networks}}{Proof of Corollary 1}}

We denote the space of $L$-Lipschitz maps from $\Xc$ to $\Yc$ by $\Lip(\Xc, \Yc; L)$ and the space of continuous maps from $\Xc$ to $\Yc$ by $\Cc(\Xc, \Yc)$. We define the uniform norm $\lVert \,\cdot\,\rVert_\infty$ of a function $f$ from a function space by \[
    \lVert f \rVert_\infty \coloneq \sup_{\mathbf{x}\in \mathcal{X}} \, \lVert f(\mathbf{x}) \rVert_2.
\]

In what follows, we use $\mathcal{W}_1$ to denote the $1$-Wasserstein distance on $\R^{d_0+d_M}$. Recall that for every pair of Borel probability measures $\mu, \nu$ on $\R^{d_0+d_M}$ with finite mean, we define \begin{equation*}
    \mathcal{W}_1 (\mu, \nu) \coloneq \inf_{\pi \in \Pi(\mu, \nu)} \left(
        \int_{\R^{d_0+d_M} \times \R^{d_0+d_M}} \rho (x, y) \dd \pi(x,y)
    \right)
\end{equation*} 
where $\Pi(\mu, \nu)$ consists of all joint probability measures on $\R^{2d}$ with marginals $\mu$ and $\nu$.

\begin{proof}
    Let $\Lambda \colon (\N_+ \cup \{+\infty\})\times \R \to [0,\infty)$ denote the right-hand side of Equation~\eqref{eqn:lip_control_via_lr_decay__MAIN}. Let $\eta > 0$ and $\Q^N \coloneq \frac{1}{N} \sum\limits_{n=1}^N \delta_{ ( \x_n, y_n ) }$ denote the empirical measure associated to the (random) dataset $\dataset$. By Theorem~\ref{thm:lip_control_via_lr_decay}, $f_{\theta_t} \in \Lip(\R^{d_0}, \R^{d_M}; \Lambda(t, \eta))$ with probability at least $1 - 2M e^{-\eta^2}$. Therefore, the following holds with probability at least $1 - 2M e^{-\eta^2}$:
    \begin{align}
        | 
            \risk (f_{\theta_T}) - \riskE (f_{\theta_T})
        |
        &= \notag
            \biggl|
                \E_{(\x,y) \in \Q} [
                    \| 
                        f_{\theta_t}(\x) - y
                    \|
                ]
                -
                \E_{(\x, y)\in \Q^N} [
                    \| 
                        f_{\theta_t}(\x) - y
                    \|
                ]
            \biggr|
        \\&\leq \notag
            \sup \limits_{f \in \Lip(\R^{d_0}, \R^{d_M} ; \Lambda(t, \eta))} 
            \Bigl|
                \E_{(\x,y) \in \Q} [
                    \| 
                        f_{\theta_t}(\x) - y
                    \|
                ]
                -
                \E_{(\x, y)\in \Q^N} [
                    \| 
                        f_{\theta_t}(\x) - y
                    \|
                ]
            \Bigr|
        \\&\leq \label{eqn:generalization_bound}
            \Lambda (t,\eta) \mathcal{W}_1 (\Q, \Q^N)
            ,
    \end{align}
    where Equation~\eqref{eqn:generalization_bound} holds by the Kantorovich-Rubinstein duality \citep[Theorem 11.8.2]{DudleyRealProb_2002Book}. Since we assumed $\dataset $ consisted of $N$ i.i.d. samples drawn from $\Q$, then upon applying the concentration inequality for $\mathcal{W}_1 (\Q, \Q^N)$ \citep{kloeckner2020empirical} in the version formulated by \citet[Lemma B.5]{hou2023instance}, we deduce that for every $\varepsilon > 0$, we have that 
    \begin{equation}
        \label{eqn:concentration}
        \mathcal{W}_1 ( \Q , \Q^N ) 
        \leq
        \frac{\diam(\supp(\Q )) C_{d_0 + d_M}}{\sqrt[d]{N}} + \varepsilon
    \end{equation}
    holds with probability at least \[
        1 - 2 e^{-2N \varepsilon^2 / \diam ( \supp ( \Q ) )^2}
        ,
    \]
    where $\supp(\Q)$ denotes the compact support of $\Q$ and the dimensional constant $C_{d_0 + d_M}$ is given by \[
        C_{d_0 + d_M} 
        \coloneq 
        2 
        \biggl( 
            \frac{\frac{d_0 + d_M}{2} - 1}{2( 1 - 2^{1 - (d_0 + d_M)/2})}
        \biggr)^{2/d+D}
        \biggl(
            1 + \frac{1}{2(\frac{d_0+d_M}{2} - 1)}
        \biggr)
        (
            d_0+d_M
        )^{1/2}
        \in
        \Oc ((d_0+d_M)^{1/2}).
    \] 
    Taking a union bound allows us to combine the concentration inequality in Equation~\eqref{eqn:concentration} with the bound in Equation~\eqref{eqn:generalization_bound} to deduce that for every $\eta, \varepsilon >0$, with probability at least \begin{equation}
         1 
         - 2M e^{-\eta^2} 
         - 2 e^{-2N \varepsilon^2 / \diam ( \supp ( \Q ) )^2}
         ,
         \label{eqn:probability_generalization_lower_bound}
    \end{equation}
    the following holds:
    \[
        | 
            \risk (f_{\theta_T}) - \riskE (f_{\theta_T})
        |
        \leq 
        \Lambda (t,\eta)
        \biggl(
            \frac{\diam(\supp(\Q )) C_{d_0 + d_M}}{\sqrt[d_0]{N}} + \varepsilon
        \biggr).
    \]
    Since $\eta>0$ was a free parameter, we retroactively set it to be $\eta \coloneq \dfrac{\diam(\supp(\Q))}{\sqrt{2N}}$. Then, the probability in Equation~\eqref{eqn:probability_generalization_lower_bound} can be bounded below by 
    \begin{equation*}
        1 - 4M e^{-2N \varepsilon^2 / \diam(\supp(\Q ))^2}.
    \end{equation*}
    Let $\delta >0$ be given. Similarly, we retroactively set $\displaystyle \varepsilon \coloneq \frac{\diam(\supp(\Q)) \sqrt{\ln (4M / \delta)}}{\sqrt{2N}}$ yielding the claim.
\end{proof}

\subsection{\texorpdfstring{Proof of Theorem~\ref{thm:convergence_optimal_gd_rate}}{Proof of Theorem 2}}

\begin{proof}
    First, observe that the requirement in Equation~\eqref{eqn:combined_lr_requirements} implies that the GD LR Decay Conditions~\ref{gdcons:grad} holds. Since we are in Setting~\ref{setting:main}, then Lemma~\ref{lem:deterministic_bounds_on_param_space} implies that for every $\eta >0$, there exists some $w \coloneq w(\eta) > 0$ such that $\theta_t \in \Theta_w$ with probability at least $1 - 2M e^{-\eta^2}$.

    Observe also that the finite second moment condition implies that \[
        \E[
            \| (\x_1, \ldots, \x_n) \|_F^2
        ]^2
        \; \leq \;
        \sum\limits_{n=1}^N \E[
            \| \x_n \|^2
        ]^2
        \; < \;
        \infty,
    \]
    hence the expected Frobenius norm of our data is finite. 
    Lastly, since we additionally assume that $\sigma, \sigma', \sigma''$ are bounded and that $\ell$ has a uniformly bounded derivative (that is, $g_{\max}' \coloneq \sup_{\x,y \in \R^{d_0}\times \R^{d_M}} \| \nabla \ell (f_\theta(\x),y) \| < \infty$), 
    then all the conditions for \citep[Corollary 3.2 and Theorem 3.3]{herrera2023locallipschitzboundsdeep} are met, thus the ER functional $\riskE$ and its gradient $\nabla \risk_\dataset$ are Lipschitz with the following constants:
    \begin{equation*}
        \begin{alignedat}{3}
            0 &\leq \Lip(\risk_\dataset) &&\leq \displaystyle
                g'_{\max} \sqrt{d_{M-1}d_M w^2 L_{N_M}^2 + d_{M-1} (\sigma_{\max})^2 +1}
            \\[6pt]
            0 &\leq \Lip(\nabla \risk_\dataset) &&\leq 
                \sqrt{a + b}
            \,,
        \end{alignedat}
    \end{equation*}
    where \[
    \begin{split}
        L_{N_M}^2 &\leq \left[ \left(\max_{1\leq m \leq M} \{d_m\} \right)^2 w \right]^{2(M-1)} (\sigma'_{\max})^{2M}(S^2 + 1) \\ &\qquad + \sum\limits_{k=1}^{M-1} \left[ \left(\max_{1\leq m \leq M} \{d_m\} \right)^2 w \right]^{2(k-1)} 
        (\sigma'_{\max})^{2k}
        \left[
            \left(\max_{1\leq m \leq M} \{d_m\} \right) (\sigma_{\max})^2 + 1
        \right]
        \\
        L_{\nabla N_M}^2 &= \Oc \left(
            M(\sigma''_{\max})^2 (\sigma_{\max}^4) 2^{M-1} (\max_{1\leq m \leq M}\{d_m\})^{10M - 9} (\sigma'_{\max} w)^{4(M-1)} (S^4 + 1)
        \right)
        \\
        a &\coloneq \max\Big\{ 
            3 L_{N_M}^2 [(g'_{\max})^2 + (g''_{\max})^2 d_{M-1}^2 d_M w^2 (\sigma_{\max})^2 + 2 (g''_{\max})^2 d_{M-1} d_M w^2 L_{N_M}^2
            ,
            \\&\qquad\qquad
            (g''_{\max})^2 (d_{M-1} \sigma_{\max}^2 + 1) (3d_{M-1} \sigma_{\max}^2 + 2)
        \Big\}
        \\
        b &\coloneq \max\Big\{
            (g'_{\max} d_{M-1} d_M w^2 L_{\nabla N_M} + g''_{\max} d_{M-1} d_M w^2 L_{N_M}^2)^2 
            \\&\qquad\qquad + L_{N_M} (g'_{\max} + d_{M-1}d_M w^2 g''_{\max} (d_{M-1} \sigma^2_{\max} + 1)^{1/2})^2
        \Big\}
        ,
    \end{split}
    \]
    and where $S$ is the expected Euclidean norm of a random data point $\x_n$ drawn from the distribution of training samples.
    
    To apply the results of \citet[Section 1.2.3]{Nesterov}, it remains to verify that our learning rate is constant for the first $T$ iterations with step size $1 / \Lip(\riskE)$. We must pick $C_{W^{[1]}}, \ldots, C_{W^{[M]}}$ in the GD LR Decay Conditions~\ref{gdcons:grad} large enough (which can be done since $T$ is finite) such that for each $t = 1,\ldots, T$, we have \[
        \frac{1}{\Lip( \riskE )} 
        \; \leq \; 
        \min_{m=1,\ldots,M}
            \left\{
                \lrcon{W}
            \right\}
        .
    \]
    Therefore, for $t = 1, \ldots, T$ and $m= 1, \ldots, M$, the learning rate constraints in Equation~\eqref{eqn:combined_lr_requirements} simplify to 
    \begin{equation}
        \notag
        \begin{aligned}
            \lr{W}
            &=
            \min
            \left\{
                1 / \Lip(\riskE)
                , \;
                \lrcon{W}
            \right\}
            \hspace{-1em}&&=
            1 / \Lip(\riskE)
        \\
            \lr{b}
            &=
            \min
            \left\{
                1 / \Lip(\riskE)
                , \;
                \lrcon{b}
            \right\}
            \hspace{-1em}&&=
            1 / \Lip(\riskE)
            .
        \end{aligned}
    \end{equation}
    Thus, for $t = 1, \ldots, T$, our gradient descent process has (finite) constant step size $1 / \Lip(\riskE)$. Consequently, we apply the convergence result for GD with constant step size by \citet[Section 1.2.3]{Nesterov} and we have that for every $\theta_0 \in \Theta_w$ and every $T\in \N_+$, for some constant $\omega > 0$, we have
    \begin{align}
        \min_{t=1\ldots, T} \| \nabla_\theta \riskE (\theta) \|
        &\leq 
            \frac{1}{\sqrt{T + 1}} 
            \left[
                \frac{\Lip(\nabla \riskE)}{\omega}
                \riskE(\theta_0)
                -
                \inf_{\theta \in \Theta_w} \riskE(\theta)
            \right] 
            \label{eqn:align:convergence_loss_stationary_point__BEGIN}
        \\&\leq 
            \frac{1}{\sqrt{T + 1}} 
            \left[
                \frac{\Lip(\nabla \riskE)}{\omega}
                \riskE (\theta_0)
            \right] \notag
        \\&\leq 
            \frac{\Lip(\nabla \riskE)}{\omega \sqrt{T + 1}} 
            \sup_{\theta \in \Theta_w} 
            \frac{1}{N} \sum\limits_{n=1}^N
            \ell (f_\theta(\x_n), y_n)^2
            \label{eqn:align:convergence_loss_stationary_point__END}
        .
    \end{align}
    Now, the continuity of $\ell$ and 
    the maps $\x \mapsto f_\theta(\x)$ and $\theta \mapsto f_\theta(\x)$
    together with the compactness of $\Theta_w$ imply that 
    \[
        C 
        \coloneq 
        \sup_{\theta \in \Theta_w} 
        \frac{1}{N} \sum \limits_{n=1}^N
            \ell (f_\theta(\x_n), y_n)^2
    \]
    is finite. This, combined with Equations~\eqref{eqn:align:convergence_loss_stationary_point__BEGIN}--\eqref{eqn:align:convergence_loss_stationary_point__END} imply that
    \[
        \min_{t = 0, \ldots, T} 
        \| \nabla_\theta \riskE ( \theta_t ) \|
        \lesssim
        \frac{\Lip(\nabla \riskE)}{\sqrt{T + 1}}
        .
    \]
    Thus, gradient descent with constant step size for a finite number of steps, then using our learning rate prescription in Equation~\eqref{eqn:combined_lr_requirements}, guarantees Lipschitz regularity of the learned network and convergence to a stationary point of the empirical risk at rate $\Oc (1 / \sqrt{T})$.
\end{proof}

\subsection{Analysis for GD LR Decay Conditions~\ref{gdcons:chained}}

If sampling the gradient of the multilayer perceptron in the parameter space directly is too expensive for each training iteration, one can use the following strengthened learning rate conditions.

\begin{gdcons}
    \label{gdcons:chained}
    Let $G$ be a rate function and let $
        C_{W^{[1]}},
        C_{b^{[1]}},
        \ldots,
        C_{W^{[M]}},
        C_{b^{[M]}}
    $ be free positive parameters. We require that the learning rates $
        \alpha_t^{W^{[1]}},
        \alpha_t^{b^{[1]}},
        \ldots,
        \alpha_t^{W^{[M]}},
        \alpha_t^{b^{[M]}}
    $ in Equation~\eqref{eqn:gd_mlp_defn} satisfy\footnote{If the gradient of the MLP in the parameter space is tractable with respect to $\dataset$, one can choose learning rates satisfying the learning rates in the GD LR Decay Conditions~\ref{gdcons:grad}.}
    \begin{equation*}
        \begin{split}
            \alpha_t^{W^{[m]}} &\leq \lrcon{W} \coloneq \frac{ C_{W^{[m]}} g(t) }{\sum \limits_{n=1}^N \begin{cases}
                    \sqrt[4]{d_{m}\cdots d_{M - 1}}
                    \| f_\theta(\x_n) - y_n \| \Lip(\sigma)^{M-m}
                    \left [ \prod \limits_{k = m + 1}^{M} 
                    \| W_{t-1}^{[k]} \|_{op; 4\to 4} \right ] \| \vec{z}_n^{[m-1]} \| & \text{ if } 1 \leq m < M
                    \\
                    \| f_\theta(\x_n) - y_n \| \| \vec{z}_n^{[M-1]} \| & \text{ if } m = M
                \end{cases}
            }
            \\
            \alpha_t^{b^{[m]}} &\leq \lrcon{b} \coloneq \frac{ C_{b^{[m]}} g(t) }{\sum \limits_{n=1}^N \begin{cases}
                    \sqrt[4]{d_{m}\cdots d_{M - 1}}
                    \| f_\theta(\x_n) - y_n \| \Lip(\sigma)^{M-m}
                    \left [ \prod \limits_{k = m + 1}^{M} 
                    \| W_{t-1}^{[k]} \|_{op; 4\to 4} \right ] & \text{ if } 1 \leq m < M
                    \\
                    \| f_\theta(\x_n) - y_n \| & \text{ if } m = M
                    .
                \end{cases}
            }
        \end{split}
    \end{equation*}
\end{gdcons}

\begin{lemma}[Cauchy-Schwarz-Type Inequality for Hadamard Products]
    \label{lem:cauchy_schwarz_hadamard}
    Let $\x, \y \in \R^d$. Then, 
    \[
        \| \x \odot \y \|_2 \leq \| \x \|_4 \| \y \|_4
        .
    \]
\end{lemma}
\begin{proof}
    \begin{align*}
        \| \x \odot \y \|_2 
        &= \sqrt{\sum\limits_{j=1}^d x_j^2y_j^2} 
        \\&= \sqrt{(x_1^2,\ldots, x_d^2) \boldsymbol{\cdot} (y_1^2,\ldots, y_d^2)}
        \\&\leq \sqrt{
            \|(x_1^2,\ldots, x_d^2)\|^2
        } \sqrt{
            \|(y_1^2,\ldots, y_d^2)\|^2
        }
        \\&= \| \x\|_4 \| \y \|_4.
    \end{align*}
\end{proof}

\begin{lemma}
    \label{lem:activation_norm_bound}
    Let $\sigma \colon \R \to \R$ be a Lipschitz continuous function with $\Lip(\sigma) = \Lip(\sigma)$.  
    Then, for $\x \in \R^d$ and $p\in \N_+ \cup \{+\infty\}$,
    \[
    \begin{split}
        \| \sigma' \bullet (\x) \|_p &\leq \Lip(\sigma) \sqrt[p]{d}
        \,.
    \end{split}
    \]
\end{lemma}
\begin{proof}
    Let $\mathbf{1}_d \in \R^d$ denote the vector containing ones in each entry. Then,
    \[
        \| \sigma' \bullet (\x) \|_p 
        \leq \| \Lip(\sigma)\mathbf{1} \|_p 
        \leq \Lip(\sigma) \| \mathbf{1}\|_p 
        = \Lip(\sigma) \sqrt[p]{d}
        .
    \]
\end{proof}

\begin{lemma}[Norm Bounds for GD-Trained Weight Matrices]
    \label{lem:norm_bounds_gd_trained_weight_matrices}
    In Setting~\ref{setting:main}, if for every $t\in \N_+$, the gradient descent process $\boldsymbol{\theta}$ with variable learning rate $\boldsymbol{\alpha}$ satisfies for every $m$,
    \begin{equation*}
        \begin{split}
            \| W_t^{[m]} \| &\leq
                \| W_0^{[m]} \| + \sum\limits_{s=1}^t 
                \frac{2 \alpha_t^{W^{[m]}}}{N} \sum \limits_{n=1}^N \begin{cases}
                    \sqrt[4]{d_{m}\cdots d_{M}}
                    \| f(\x_n) - y_n \|_4 \Lip(\sigma)^{M-m+1}
                    \left [ \prod \limits_{k = m + 1}^{M} 
                    \| W_{t-1}^{[k]} \|_{op; 4\to 4} \right ] \| \vec{z}_n^{[m-1]} \| & \text{ if } 1 \leq m < M
                    \\
                    \| f(\x_n) - y_n \|_4 \Lip(\sigma) \sqrt[4]{d_M} \| \vec{z}_n^{[M-1]} \| & \text{ if } m = M
                \end{cases}
            \\
            \| b_t^{[m]} \| &\leq
                \| b_0^{[m]} \| + \sum\limits_{s=1}^t
                \frac{2 \alpha_t^{b^{[M]}}}{N} \sum \limits_{n=1}^N \begin{cases}
                    \sqrt[4]{d_{m}\cdots d_{M}}
                    \| f(\x_n) - y_n \|_4 \Lip(\sigma)^{M-m+1}
                    \left [ \prod \limits_{k = m + 1}^{M} 
                    \| W_{t-1}^{[k]} \|_{op; 4\to 4} \right ] & \text{ if } 1 \leq m < M
                    \\
                    \| f(\x_n) - y_n \|_4 \Lip(\sigma) \sqrt[4]{d_M} & \text{ if } m = M
                \end{cases}
        \end{split}
    \end{equation*}
\end{lemma}
\begin{proof}
    By applying the triangle and Cauchy-Schwarz inequalities in~Equation~\eqref{eqn:gd_mlp_defn}, we have
    \begin{equation}
        \notag
        \begin{split}
            \| W_t^{[m]} - W_{t-1}^{[m]} \| &\leq
                \frac{2 \alpha_t^{W^{[m]}}}{N} \sum \limits_{n=1}^N \| \vec{v}_n^{[m]} \| \| \vec{z}_n^{[m-1]} \|
            \\
            \| b_t^{[m]} - b_{t-1}^{[m]} \| &\leq
                \frac{2 \alpha_t^{b^{[m]}}}{N} \sum \limits_{n=1}^N \| \vec{v}_n^{[m]} \|
            .
        \end{split}
    \end{equation}
    We may similarly apply the triangle inequality, the Cauchy-Schwarz-Type Inequality for Hadamard Products in Lemma~\ref{lem:cauchy_schwarz_hadamard}, and the bound in Lemma~\ref{lem:activation_norm_bound} to produce, for $\vec{v}_n^{[M]}$:
    \begin{equation}
        \notag
        \begin{split}
            \| \vec{v}_n^{[M]} \| 
            &= 
                \| 2 ( f(\x_n) - y_n ) \|
            \\&= 
                2 \| f(\x_n) - y_n  \|
        \end{split}
    \end{equation}
    and for $1 \leq m < M$,
    \begin{align}
        \| \vec{v}_n^{[m]} \|
        &= \| ( (W_{t-1}^{[m + 1]} )^\top \vec{v}_n^{[m + 1]} \odot \sigma' \bullet ( \vec{a}_n^{[m]} ) \| \notag
        \\&\leq \| ( (W_{t-1}^{[m + 1]} )^\top \vec{v}_n^{[m + 1]} \|_4 \| \sigma' \bullet ( \vec{a}_n^{[m]} ) \|_4 \notag
        \\&\leq \| W_{t-1}^{[m + 1]} \|_{op; 4\to 4} \| \vec{v}_n^{[m + 1]} \|_4 \| \sigma' \bullet ( \vec{a}_n^{[m]} ) \|_4 \notag
        \\&\leq \| W_{t-1}^{[m + 1]} \|_{op; 4\to 4} \; \| \vec{v}_n^{[m + 1]} \|_4 \Lip(\sigma) \sqrt[4]{d_m} \notag
        \\&\leq \| W_{t-1}^{[m + 1]} \|_{op; 4\to 4} \; \| \vec{v}_n^{[m + 1]} \| \Lip(\sigma) \sqrt[4]{d_m}. \label{eqn:v_n_bound_single_step__END}
    \end{align}
    Solving recursively, using Equation~\eqref{eqn:v_n_bound_single_step__END}, we have
    \begin{align}
        \| \vec{v}_n^{[m]} \|
        &\leq 
            \| W_{t-1}^{[m + 1]} \|_{op; 4\to 4} \;
            \| \vec{v}_n^{[m + 1]} \| 
            \Lip(\sigma) \sqrt[4]{d_{m}} \notag
        \\&\leq 
            \| W_{t-1}^{[m + 1]} \|_{op; 4\to 4} \; \| W_{t-1}^{[m + 2]} \|_{op; 4\to 4} \; 
            \| \vec{v}_n^{[m + 2]} \|
            \Lip(\sigma)^2 \sqrt[4]{d_{m+1}}
            \sqrt[4]{d_{m}} \notag
        \\&\;\;\vdots \notag
        \\&\leq 
            \sqrt[4]{d_{m}\cdots d_{M - 1}}
            \| \vec{v}_n^{[M]} \|_4 \Lip(\sigma)^{M-m}
            \prod \limits_{k = m + 1}^{M} 
            \| W_{t-1}^{[k]} \|_{op; 4\to 4} \notag
        \\&\leq 
            2 \sqrt[4]{d_{m}\cdots d_{M - 1}}
            \| f(\x_n) - y_n \| \Lip(\sigma)^{M-m}
            \prod \limits_{k = m + 1}^{M} 
            \| W_{t-1}^{[k]} \|_{op; 4\to 4} \notag
            ,
    \end{align}
    and thus for each $t$ and each $1\leq m < M$, 
    \begin{equation}
        \label{eqn:finite_difference_param_norm_bounds_hidden_layers}
        \begin{split}
            \| W_t^{[m]} - W_{t-1}^{[m]} \| &\leq  
                \frac{2 \alpha_t^{W^{[m]}}}{N} \sum \limits_{n=1}^N \sqrt[4]{d_{m}\cdots d_{M - 1}}
            \| f(\x_n) - y_n \| \Lip(\sigma)^{M-m}
            \left [ \prod \limits_{k = m + 1}^{M} 
            \| W_{t-1}^{[k]} \|_{op; 4\to 4} \right ] \| \vec{z}_n^{[m-1]} \|
            \\
            \| b_t^{[m]} - b_{t-1}^{[m]} \| &\leq 
                \frac{2 \alpha_t^{b^{[m]}}}{N} \sum \limits_{n=1}^N \sqrt[4]{d_{m}\cdots d_{M - 1}}
            \| f(\x_n) - y_n \| \Lip(\sigma)^{M-m}
            \left [ \prod \limits_{k = m + 1}^{M} 
            \| W_{t-1}^{[k]} \|_{op; 4\to 4} \right ]
            ,
        \end{split}
    \end{equation}
    and for $ m = M $, we have
    \begin{equation}
        \label{eqn:finite_difference_param_norm_bounds_final_layer}
        \begin{split}
            \| W_t^{[M]} - W_{t-1}^{[M]} \| &\leq
                \frac{2 \alpha_t^{W^{[M]}}}{N} \sum \limits_{n=1}^N \| f(\x_n) - y_n \|  \| \vec{z}_n^{[M-1]} \|
            \\
            \| b_t^{[M]} - b_{t-1}^{[M]} \| &\leq
                \frac{2 \alpha_t^{b^{[M]}}}{N} \sum \limits_{n=1}^N \| f(\x_n) - y_n \| 
            .
        \end{split}
    \end{equation}
    By applying the triangle inequality, we have 
    \begin{equation}
        \notag
        \begin{aligned}
            \| W_t^{[m]} \| 
            &\leq  
            \| W_0^{[m]} \| + \| W_t^{[m]} - W_0^{[m]} \| 
            && \leq && 
            \|  W_0^{[m]} \| + \sum\limits_{s=1}^t \|  W_{s}^{[m]} - W_{s-1}^{[m]} \|
            \\
            \| b_t^{[m]} \| 
            &\leq  
            \| b_0^{[m]} \| + \| b_t^{[m]} - b_0^{[m]} \| 
            && \leq && 
            \|  b_0^{[m]} \| + \sum\limits_{s=1}^t \|  b_{s}^{[m]} - b_{s-1}^{[m]} \|
            ,
        \end{aligned}
    \end{equation}
    and substituting the bounds obtained in Equation~\eqref{eqn:finite_difference_param_norm_bounds_hidden_layers} and Equation~\eqref{eqn:finite_difference_param_norm_bounds_final_layer}, we have
    \begin{equation}
        \notag
        \begin{split}
            \| W_t^{[m]} \| &\leq
                \| W_0^{[m]} \| + \sum\limits_{s=1}^t 
                \frac{2 \alpha_t^{W^{[m]}}}{N} \sum \limits_{n=1}^N \begin{cases}
                    \sqrt[4]{d_{m}\cdots d_{M - 1}}
                    \| f(\x_n) - y_n \| \Lip(\sigma)^{M-m}
                    \left [ \prod \limits_{k = m + 1}^{M} 
                    \| W_{t-1}^{[k]} \|_{op; 4\to 4} \right ] \| \vec{z}_n^{[m-1]} \| & \text{ if } 1 \leq m < M
                    \\
                    \| f(\x_n) - y_n \| \| \vec{z}_n^{[M-1]} \| & \text{ if } m = M
                \end{cases}
            \\
            \| b_t^{[m]} \| &\leq
                \| b_0^{[m]} \| + \sum\limits_{s=1}^t
                \frac{2 \alpha_t^{b^{[M]}}}{N} \sum \limits_{n=1}^N \begin{cases}
                    \sqrt[4]{d_{m}\cdots d_{M - 1}}
                    \| f(\x_n) - y_n \| \Lip(\sigma)^{M-m}
                    \left [ \prod \limits_{k = m + 1}^{M} 
                    \| W_{t-1}^{[k]} \|_{op; 4\to 4} \right ] & \text{ if } 1 \leq m < M
                    \\
                    \| f(\x_n) - y_n \| & \text{ if } m = M
                \end{cases}
        \end{split}
    \end{equation}
    as desired.
\end{proof}

Having bounded the norms of the weight matrices, we may use the estimates in Lemma~\ref{lem:norm_bounds_gd_trained_weight_matrices} to deduce conditions on the variable learning rates of the gradient descent, which force the neural network's weights to grow at a given target rate. This target growth rate is encoded by a prescribed rate function.

\begin{lemma}[Control on the Growth Rate of the Norm of GD-Trained Weight Matrices]
    \label{lem:control_growth_norm_gd_trained_weight_matrices}
    In Setting~\ref{setting:main}, if the gradient descent process $\boldsymbol\theta$ with variable learning rate $\boldsymbol{\alpha}$ satisfies the GD LR Decay Conditions~\ref{gdcons:chained}, then \[
        \| W_{t}^{[1]}\| ,
        \| b_{t}^{[1]}\| ,
        \ldots, \| W_{t}^{[M]}\| ,
        \| b_{t}^{[M]}\|
        \lesssim
        G(t)
        .
    \]
\end{lemma}
\begin{proof}
    Setting 
    \begin{equation*}
        \begin{split}
            \gamma_t^{W^{[m]}} &\coloneq \alpha_t^{W^{[m]}} \sum \limits_{n=1}^N \begin{cases}
                    \sqrt[4]{d_{m}\cdots d_{M - 1}}
                    \| f(\x_n) - y_n \| \Lip(\sigma)^{M-m}
                    \left [ \prod \limits_{k = m + 1}^{M} 
                    \| W_{t-1}^{[k]} \|_{op; 4\to 4} \right ] \| \vec{z}_n^{[m-1]} \| & \text{ if } 1 \leq m < M
                    \\
                    \| f(\x_n) - y_n \| \| \vec{z}_n^{[M-1]} \| & \text{ if } m = M
                \end{cases}
            \\
            \gamma_t^{b^{[m]}} &\coloneq \alpha_t^{b^{[m]}} \sum \limits_{n=1}^N \begin{cases}
                    \sqrt[4]{d_{m}\cdots d_{M - 1}}
                    \| f(\x_n) - y_n \| \Lip(\sigma)^{M-m}
                    \left [ \prod \limits_{k = m + 1}^{M} 
                    \| W_{t-1}^{[k]} \|_{op; 4\to 4} \right ] & \text{ if } 1 \leq m < M
                    \\
                    \| f(\x_n) - y_n \| & \text{ if } m = M
                \end{cases}
        \end{split}
    \end{equation*}
    allows us to restate our GD LR Decay Conditions~\ref{gdcons:chained} as
    \begin{equation*}
        \begin{split}
            \gamma_t^{W^{[m]}} &\leq C_{W^{[m]}} g(t)
            \\
            \gamma_t^{b^{[m]}} &\leq C_{b^{[m]}} g(t)
            .
        \end{split}
    \end{equation*}
    Then, by Lemma~\ref{lem:norm_bounds_gd_trained_weight_matrices}, we have
    \begin{equation}
        \begin{aligned}
            \| W_t^{[m]} \| &\leq \| W_0^{[m]} \| + \frac{2}{N} \sum\limits_{s=1}^t \gamma_s^{W^{[m]}} \!\!\!\!\!\! &&\leq \| W_0^{[m]} \| + \frac{2}{N} \sum\limits_{s=1}^t C_{W^{[m]}} g(s) \!\!\!\!\!\! &&\leq \| W_0^{[m]} \| + \frac{2}{N} C_{W^{[m]}} (G(t) + g(1))
            \\
            \| b_t^{[m]} \| &\leq \| b_0^{[m]} \| + \frac{2}{N} \sum\limits_{s=1}^t \gamma_s^{b^{[m]}} \!\!\!\!\!\! &&\leq \| b_0^{[m]} \| + \frac{2}{N} \sum\limits_{s=1}^t C_{b^{[m]}} g(s) \!\!\!\!\!\! &&\leq \| b_0^{[m]} \| + \frac{2}{N} C_{b^{[m]}} (G(t) + g(1))
            ,
        \end{aligned}
    \end{equation}
    as desired.
\end{proof}

This theorem again places us within the scope of Lemma~\ref{lem:control_growth_norm_gd_trained_weight_matrices:implicit}, and thus satisfying the GD LR Decay Conditions~\ref{gdcons:chained} also is sufficient for the Lipschitz bound in Theorem~\ref{thm:lip_control_via_lr_decay} and its consequences.

\section{Experiment Details} \label{appendix:sec:experiment_details}

We trained our models on a computer with the following specifications.

\begin{itemize}
    \item \textbf{CPU}: 12th Gen Intel(R) Core(TM) i5-12400F 2.50 GHz. 
    \item \textbf{RAM}: 32.0 GB. 
    \item \textbf{GPU}: NVIDIA GeForce RTX 3060.
\end{itemize}

We use PyTorch with CUDA for all scientific computing and array/Tensor operations \citep{PyTorch}. We produce plots using Matplotlib \citep{MatPlotLib} and Seaborn \citep{seaborn}, using NumPy \citep{numpy} for compatibility.

\subsection{Target Functions} \label{appendix:sec:experiment_details:subsec:target_functions}

\noindent In each experiment, the task is a non-parametric regression, modeled as  where $h$ could be:

\vspace{1em}
\citet{Franke1979} 

$h\colon [0,1]^2 \to \R$, \; $h(\x) = 0.75\exp\left(-\dfrac{(9x_1-2)^2}{4} - \dfrac{(9x_2-2)^2}{4}\right) + 0.75\exp\left( - \dfrac{(9x_1+1)^2}{49} - \dfrac{9x_2 + 1}{10}\right) + 0.5\exp\left(-\dfrac{(9x_1-7)^2}{4} - \dfrac{(9x_2 - 3)^2}{4}\right) - 0.2 \exp\left(-(9x_1 - 4)^2 - (9x_2 - 7)^2\right)$.

\vspace{1em}
\citet{Forrester2008} 

$h\colon [0,1] \to \R$, \; $h(x) = (6x-2)^2\sin(12x - 4)$.

\section{Additional Linear Regression Analysis}

In this section, we produce an explicit expression for the Lipschitz constant of a linear regression model trained with standard gradient descent and a constant learning rate on fixed dataset $\dataset$.

\begin{theorem}[Lipschitz Constant for Linear Regression]
    Let $\dataset =\{ (x_n, y_n )\}_{n=1}^N$ be a dataset where $x_n\in \mathbb{R}^d$ and $y_n \in \mathbb{R}$. Consider the linear model $f_{\mathbf{a},b}(x) = \mathbf{a}^\top x + b$ where $\mathbf{a}\in \mathbb{R}^{d}$ and $b \in \mathbb{R}$. Let $\theta \coloneq (\mathbf{a}, b)$ and $\mathbf{x} \coloneq (x, 1)$, thus we have $f_\theta(\mathbf{x}) = \theta^\top \mathbf{x}$. Fix an initial parameter vector $\theta_0 \in \mathbb{R}^{d+1}$ and a constant step size $\alpha>0$. Performing gradient descent, training on the mean squared error based empirical risk, \[
        \theta_{t+1} \coloneq \theta_t - \alpha \nabla_{\theta} \frac{1}{N}\sum \limits_{n=1}^N \Big|
            \theta_t^\top \mathbf{x}_n - y_n
        \Big|^2,
    \]
    then the learned function after $t$ epochs is Lipschitz continuous with respect to the input $\mathbf{x}$ with Lipschitz constant \[
        \Lip(f_{\theta_t}) = \left\| \left(
            \theta_0 - \left(\sum \limits_{n=1}^N \mathbf{x}_n \Big(
            \theta_0^\top \mathbf{x}_n - y_n
            \Big)\right) \frac{1 - \left[1- \dfrac{2\alpha}{N} \sum \limits_{n=1}^N \lVert\mathbf{x}_n\rVert_2^2\right]^t}{ \sum \limits_{n=1}^N \lVert\mathbf{x}_n\rVert_2^2}
        \right)^{(1:d)}
        \right\|_2 .
    \]
\end{theorem}

\begin{proof}
    First, we simplify the gradient computation in the gradient descent step: \begin{align}
        \theta_{t+1} 
        &= \theta_t - \alpha \nabla_{\theta} \frac{1}{N}\sum \limits_{n=1}^N \Big|
            \theta_t^\top \mathbf{x}_n - y_n
            \Big|^2 \notag
        \\&= \theta_t - \alpha \frac{1}{N}\sum \limits_{n=1}^N \nabla_{\theta} \Big|
            \theta_t^\top \mathbf{x}_n - y_n
            \Big|^2 \notag
        \\&= \theta_t -  \frac{2\alpha}{N}\sum \limits_{n=1}^N \mathbf{x}_n \Big(
            \theta_t^\top \mathbf{x}_n - y_n
            \Big) , \label{eqn:linear_regression_analysis}
    \end{align}
    where Equation~$\eqref{eqn:linear_regression_analysis}$ is obtained by taking the gradient.

    Now, let $\Delta \theta_t \coloneq \theta_t - \theta_{t-1}$ denote the backward difference. Substituting our gradient descent step at each point yields \begin{align}
        \Delta \theta_{t}
        &=  \theta_t - \theta_{t-1}\notag
        \\
        &= \left[\theta_{t-1} -  \frac{2\alpha}{N}\sum \limits_{n=1}^N \mathbf{x}_n \Big( \theta_{t-1}^\top \mathbf{x}_n - y_n \Big)\right] 
        - 
        \left[\theta_{t-2} -  \frac{2\alpha}{N}\sum \limits_{n=1}^N \mathbf{x}_n \Big( \theta_{t-2}^\top \mathbf{x}_n - y_n \Big)\right]\notag
        \\
        &= (\theta_{t-1} - \theta_{t-2}) - \frac{2\alpha}{N} \sum \limits_{n=1}^N \left[
            \mathbf{x}_n(\theta_{t-1}^\top \mathbf{x}_n - y_n) - \mathbf{x}_n(\theta_{t-2}^\top \mathbf{x}_n - y_n)
            \right] \notag
        \\
        &= \Delta \theta_{t-1} - \dfrac{2\alpha}{N} \sum \limits_{n=1}^N \mathbf{x}_n \left[\theta_{t-1}^\top \mathbf{x}_n - \theta_{t-2}^\top\mathbf{x}_n \right]\notag
        \\
        &= \Delta \theta_{t-1} - \dfrac{2\alpha}{N} \sum \limits_{n=1}^N \mathbf{x}_n \left[\Delta\theta_{t-1}^\top\mathbf{x}_n \right] \notag
        \\
        &= \Delta \theta_{t-1} - \dfrac{2\alpha}{N} \sum \limits_{n=1}^N \Delta\theta_{t-1}(\mathbf{x}_n^\top \mathbf{x}_n ) \notag
        \\
        &= \Delta \theta_{t-1} \left[ 1- \dfrac{2\alpha}{N} \sum \limits_{n=1}^N \lVert\mathbf{x}_n\rVert_2^2 \right], \notag
    \end{align}
    thus, the sequence of backward differences is \emph{geometric} with common ratio $1- \dfrac{2\alpha}{N} \sum \limits_{n=1}^N \lVert\mathbf{x}_n\rVert_2^2$.

    Thus, we have that \begin{equation*}
        \Delta \theta_t = \Delta \theta_1 \left[1- \dfrac{2\alpha}{N} \sum \limits_{n=1}^N \lVert\mathbf{x}_n\rVert_2^2\right]^{t-1}
        ,
    \end{equation*}
    which implies that \begin{align*}
        \theta_t &= \theta_0 + \sum\limits_{s=1}^t \Delta \theta_s
        \\&= 
            \theta_0 + \sum\limits_{s=1}^t \Delta \theta_1 
                \left[1- \dfrac{2\alpha}{N} \sum \limits_{n=1}^N \lVert\mathbf{x}_n\rVert_2^2\right]
            ^{s-1}
        \\&= 
            \theta_0 + \Delta\theta_1 \frac{1 - \left[1- \dfrac{2\alpha}{N} \sum \limits_{n=1}^N \lVert\mathbf{x}_n\rVert_2^2\right]^t}{1-\left[1- \dfrac{2\alpha}{N} \sum \limits_{n=1}^N \lVert\mathbf{x}_n\rVert_2^2\right]}
        \\&= 
            \theta_0 + \left(-  \frac{2\alpha}{N}\sum \limits_{n=1}^N \mathbf{x}_n \Big(
            \theta_0^\top \mathbf{x}_n - y_n
            \Big)\right) \frac{1 - \left[1- \dfrac{2\alpha}{N} \sum \limits_{n=1}^N \lVert\mathbf{x}_n\rVert_2^2\right]^t}{\dfrac{2\alpha}{N} \sum \limits_{n=1}^N \lVert\mathbf{x}_n\rVert_2^2}
        \\&= 
            \theta_0 - \left(\sum \limits_{n=1}^N \mathbf{x}_n \Big(
            \theta_0^\top \mathbf{x}_n - y_n
            \Big)\right) \frac{1 - \left[1- \dfrac{2\alpha}{N} \sum \limits_{n=1}^N \lVert\mathbf{x}_n\rVert_2^2\right]^t}{ \sum \limits_{n=1}^N \lVert\mathbf{x}_n\rVert_2^2}
            .
    \end{align*}
    Since the Lipschitz constant of an affine map is simply the $L_2$ norm of the weight vector, we have
    \[
        \Lip(f_{\theta_t}) = \left\|
            \left(
                \theta_0 - \left(\sum \limits_{n=1}^N \mathbf{x}_n \Big(
                \theta_0^\top \mathbf{x}_n - y_n
                \Big)\right) \frac{1 - \left[1- \dfrac{2\alpha}{N} \sum \limits_{n=1}^N \lVert\mathbf{x}_n\rVert_2^2\right]^t}{ \sum \limits_{n=1}^N \lVert\mathbf{x}_n\rVert_2^2}
            \right)^{(1:d)}
        \right\|_2,
    \]
    where $(\,\cdot\,)^{(1:d)}$ denotes the first $d$ components of the vector. 
\end{proof}

\section{Additional Experiments}

\begin{figure}[h!]
    \centering
    
    \begin{subfigure}{0.49\textwidth}
        \centering
        \includegraphics[width=\linewidth]{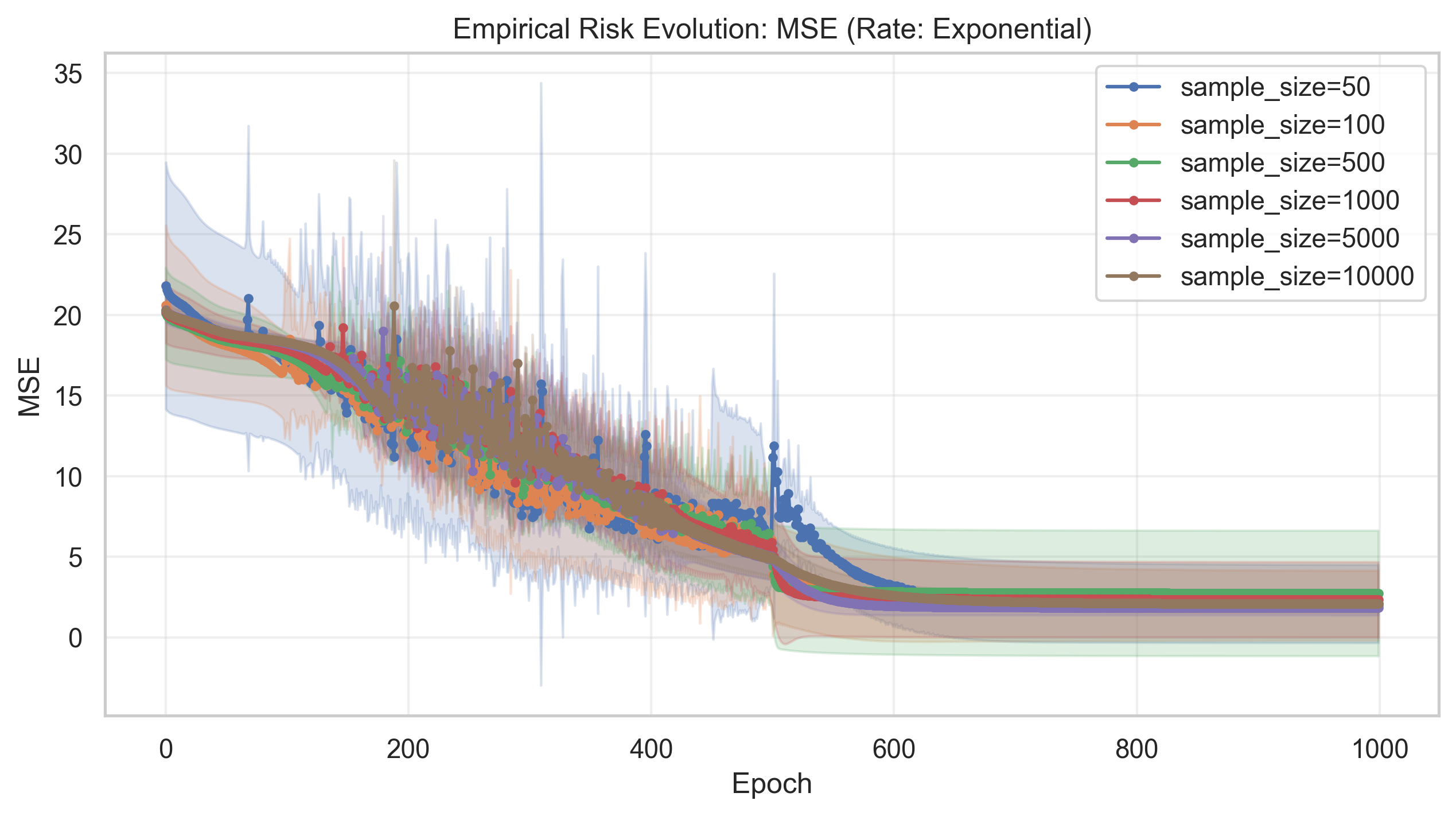}
    \end{subfigure}
    \hfill
    \begin{subfigure}{0.49\textwidth}
        \centering
        \includegraphics[width=\linewidth]{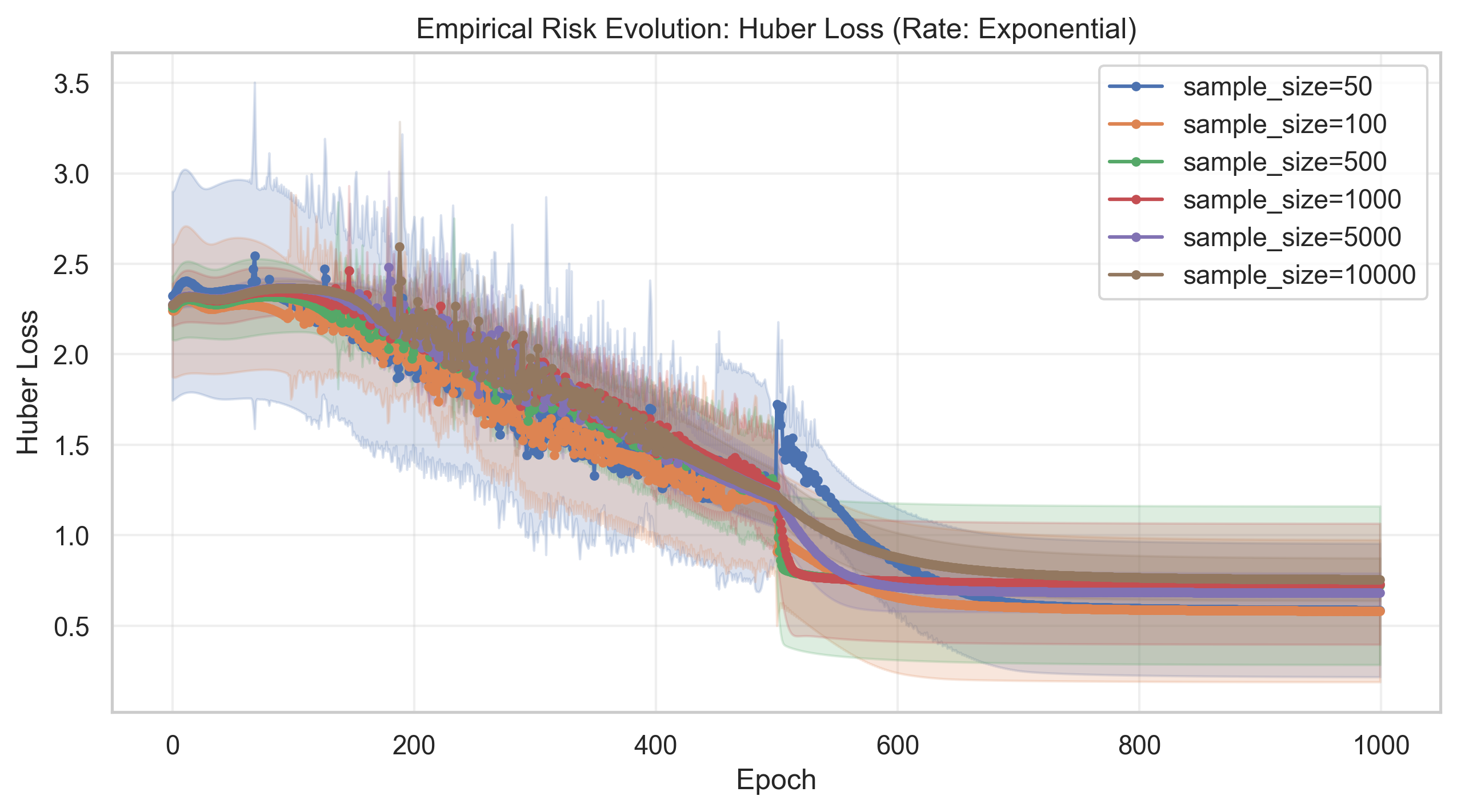}
    \end{subfigure}
    \begin{subfigure}{0.49\textwidth}
        \centering
        \includegraphics[width=\linewidth]{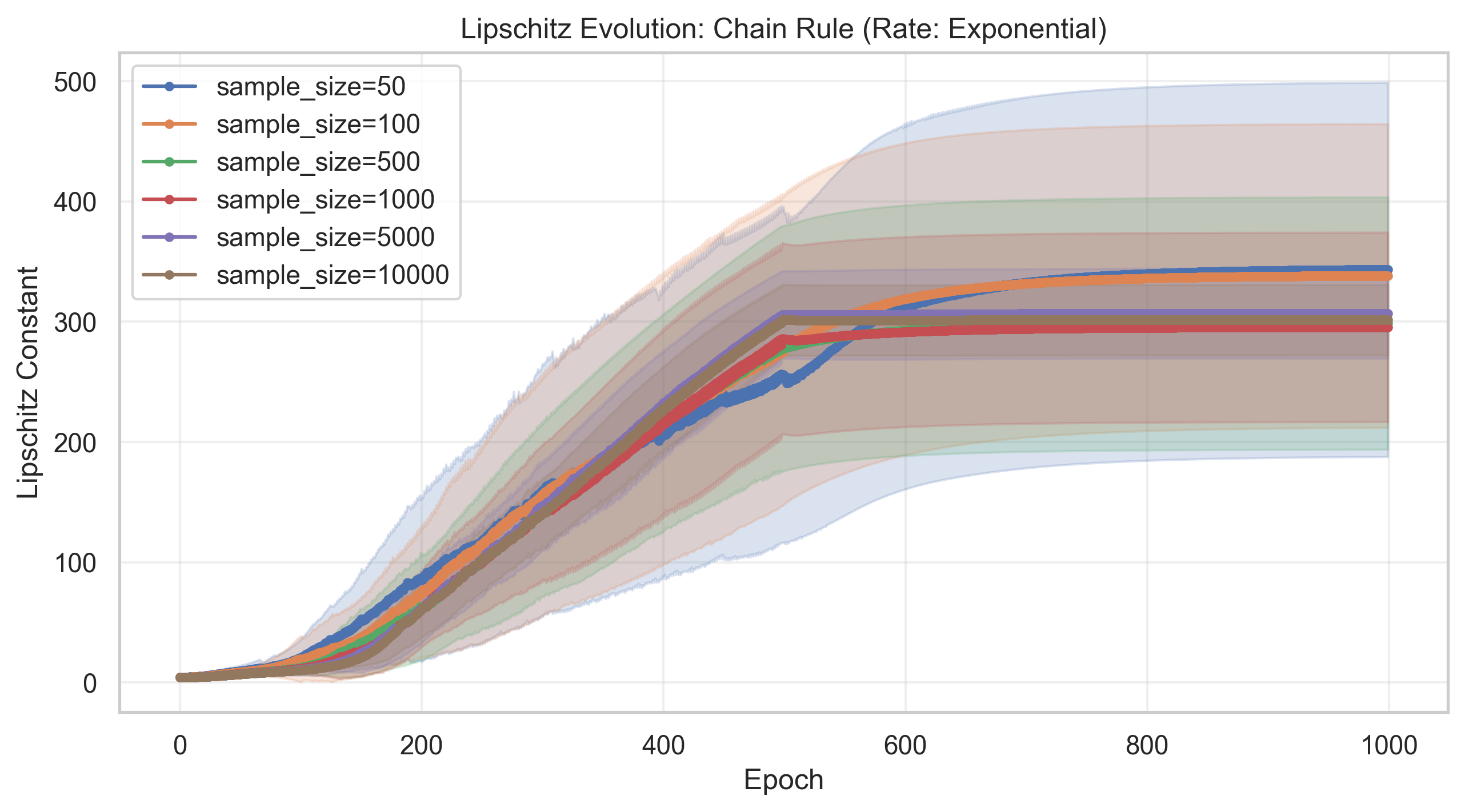}
    \end{subfigure}
        \begin{subfigure}{0.49\textwidth}
        \centering
        \includegraphics[width=\linewidth]{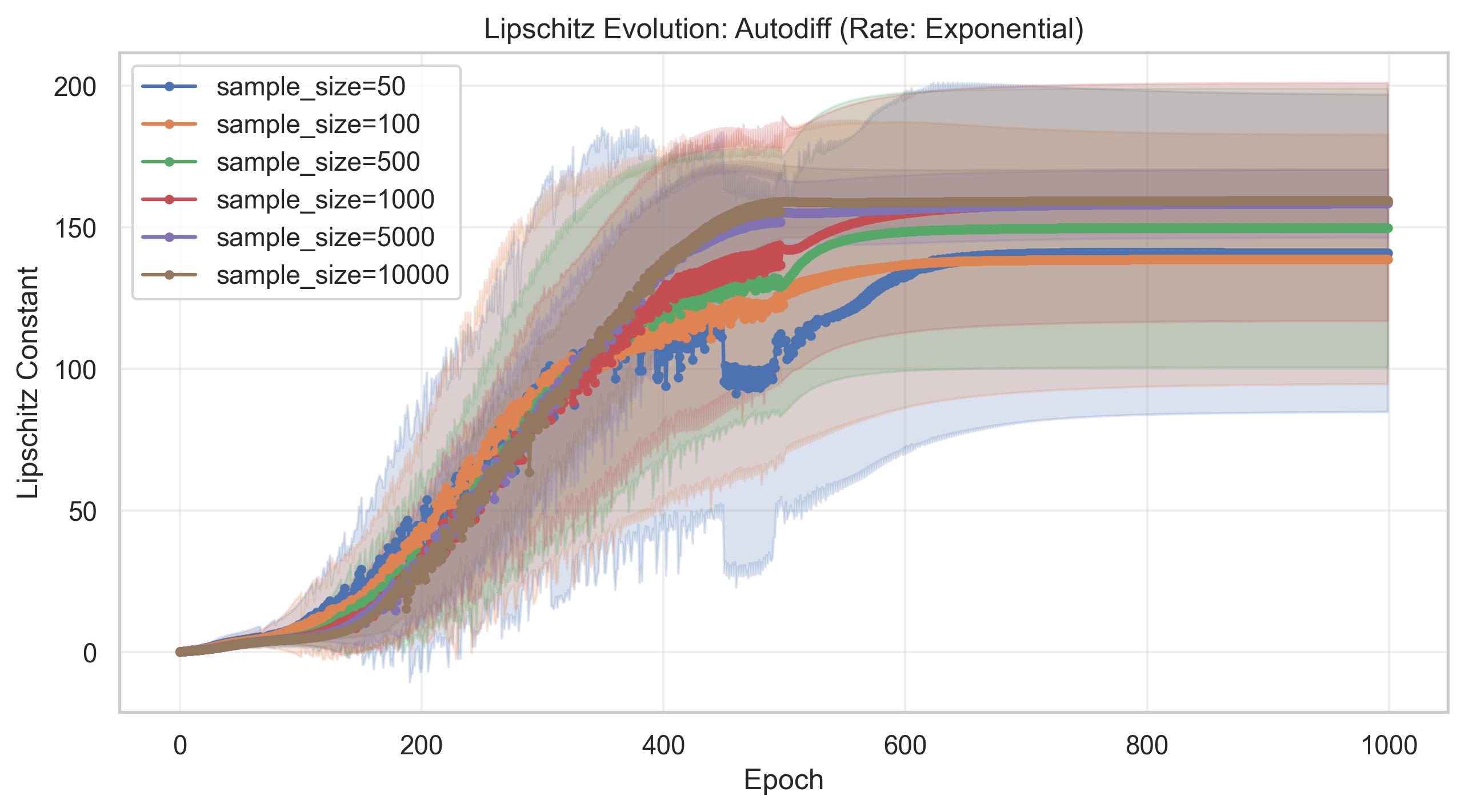}
    \end{subfigure}

    \caption{\textbf{Sample Size Ablation}: We train MLPs with width 100, depth 3, and $\tanh$ activation for 1000 epochs on i.i.d. uniform samples of the Forrester function with standard Gaussian noise with noise level $\beta = 0.5$ and initial LR $\alpha = 0.01$, starting our GD LR Decay Conditions at epoch $T=500$ using the exponential rate function  (Example~\ref{ex:polynomial_convergence}) with $r=0.03$. We lower bound the Lipschitz constant with 6000 gradient samples. We perform this process 10 times, plotting the mean and one standard deviation.}
    \label{fig:experiment:sample_size_ablation}
\end{figure}

\begin{figure}[h!]
    \centering
    
    \begin{subfigure}{0.49\textwidth}
        \centering
        \includegraphics[width=\linewidth]{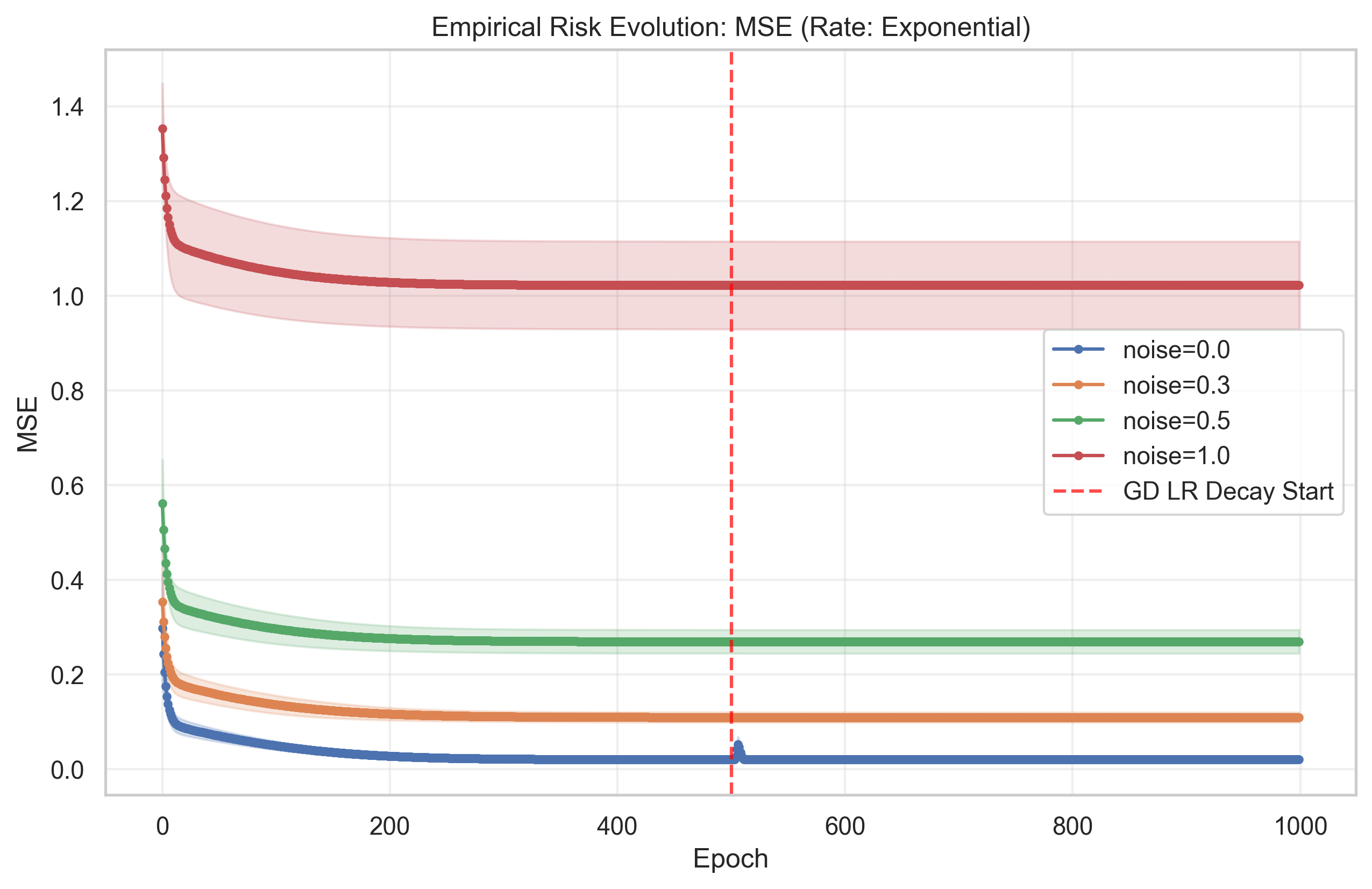}
    \end{subfigure}
    \hfill
    \begin{subfigure}{0.49\textwidth}
        \centering
        \includegraphics[width=\linewidth]{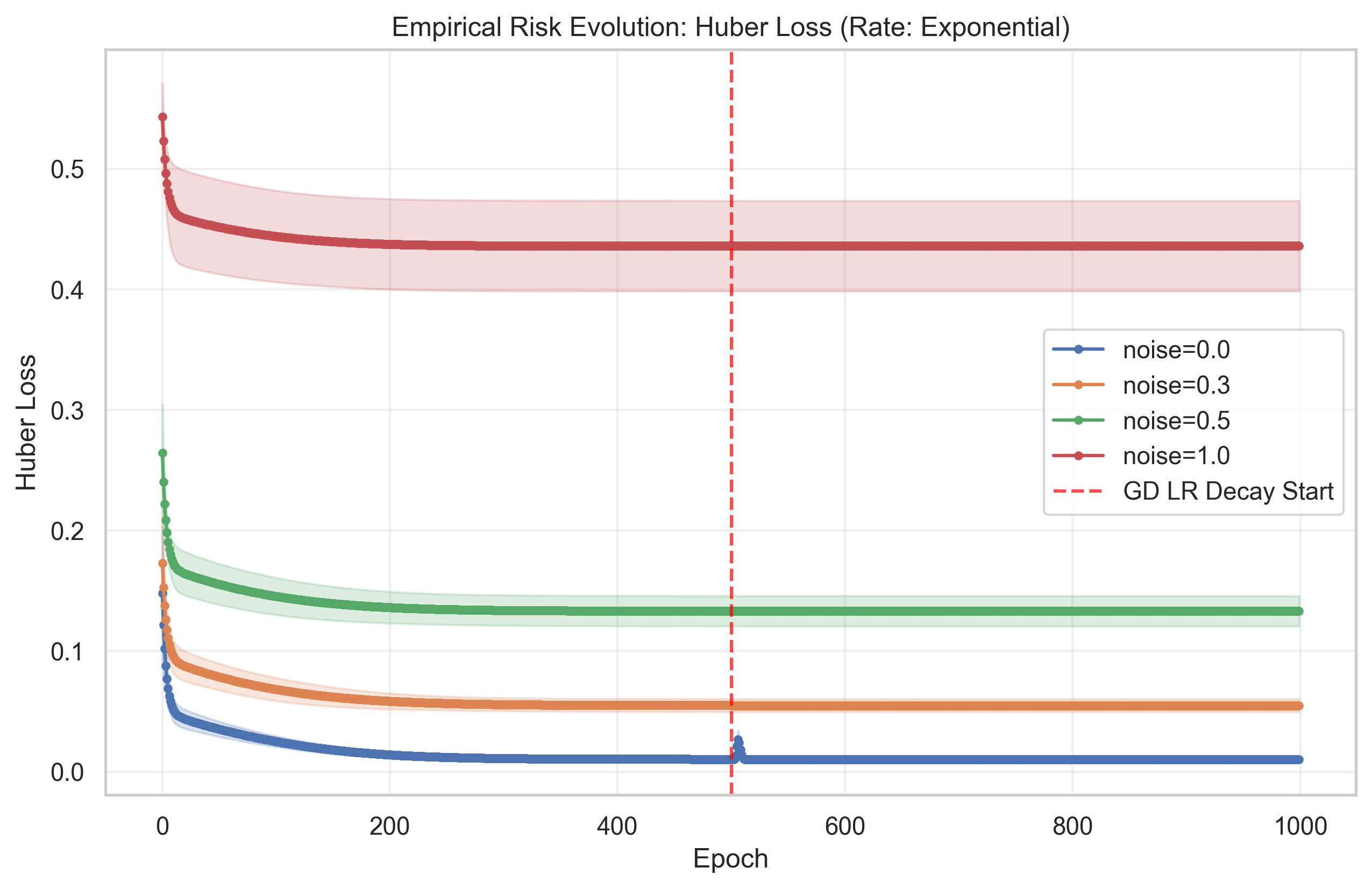}
    \end{subfigure}
    \begin{subfigure}{0.49\textwidth}
        \centering
        \includegraphics[width=\linewidth]{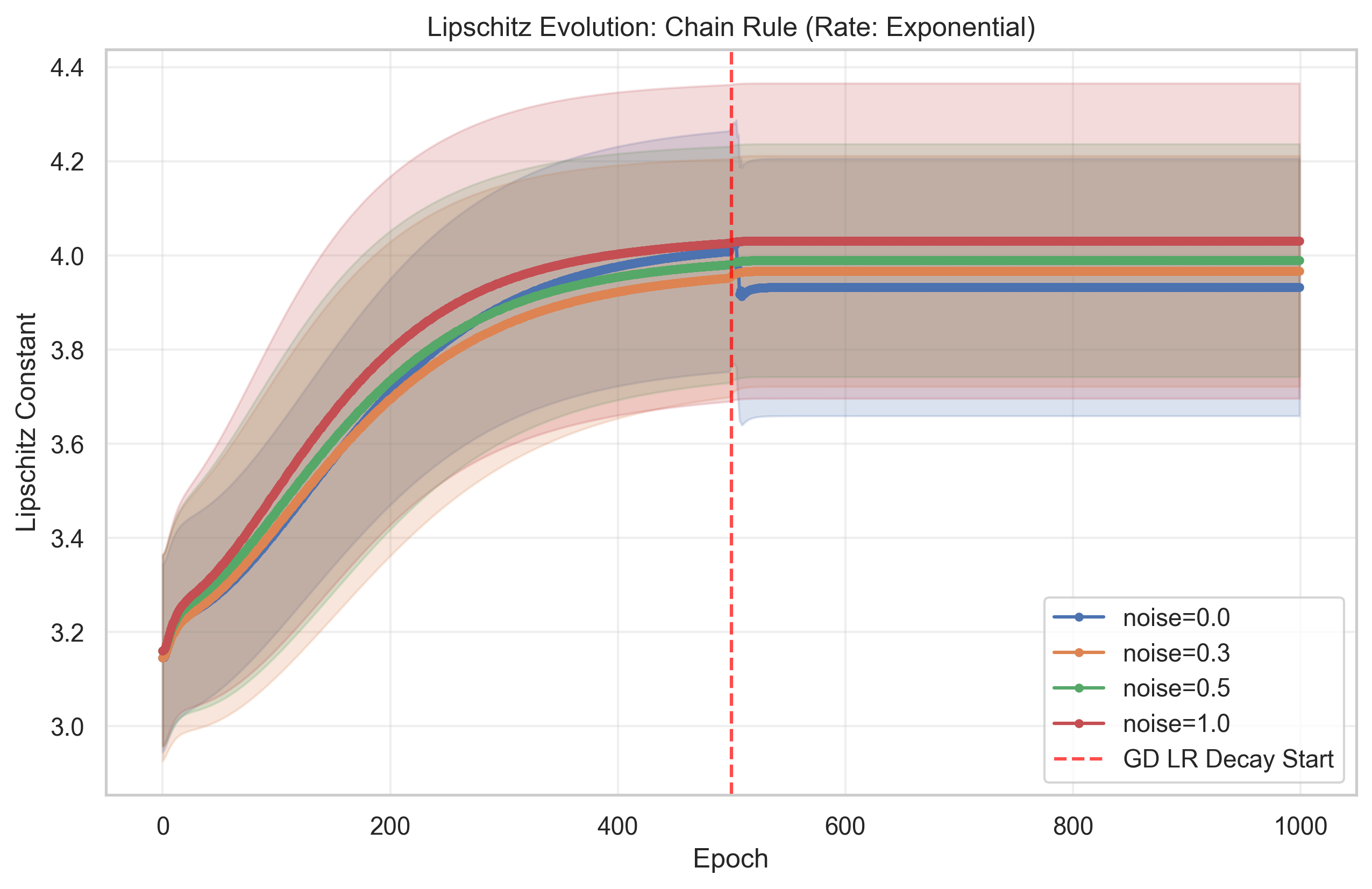}
    \end{subfigure}
        \begin{subfigure}{0.49\textwidth}
        \centering
        \includegraphics[width=\linewidth]{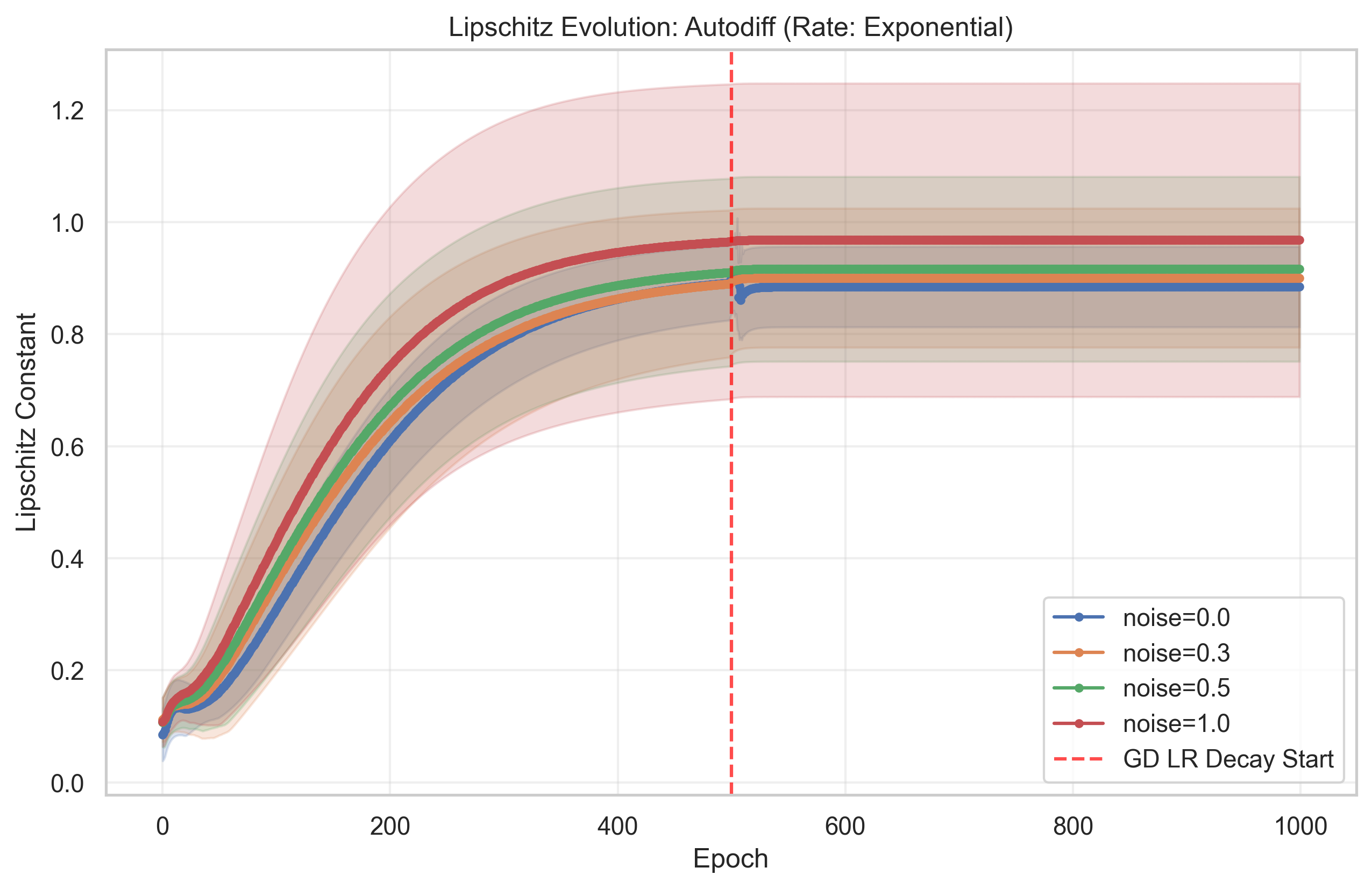}
    \end{subfigure}

    \caption{\textbf{Noise Ablation}: We train MLPs with width 100, depth 3, and $\tanh$ activation for 1000 epochs on $N = 100$ i.i.d. uniform samples of the Forrester function with standard Gaussian noise with varying noise level and initial LR $\alpha = 0.01$, starting our GD LR Decay Conditions at epoch $T=500$ using the exponential rate function  (Example~\ref{ex:polynomial_convergence}) with $r=0.03$. We lower bound the Lipschitz constant with 6000 gradient samples. We perform this process 10 times, plotting the mean and one standard deviation.}
    \label{fig:experiment:noise_ablation}
\end{figure}

We showcase the results of the sample size ablation experiment described in Section~\ref{s:Experiments} in Figure~\ref{fig:experiment:sample_size_ablation}.

Figure~\ref{fig:experiment:noise_ablation} showcases a noise ablation experiment.

\begin{figure}[h!]
    \centering
    
    \begin{subfigure}{0.49\textwidth}
        \centering
        \includegraphics[width=\linewidth]{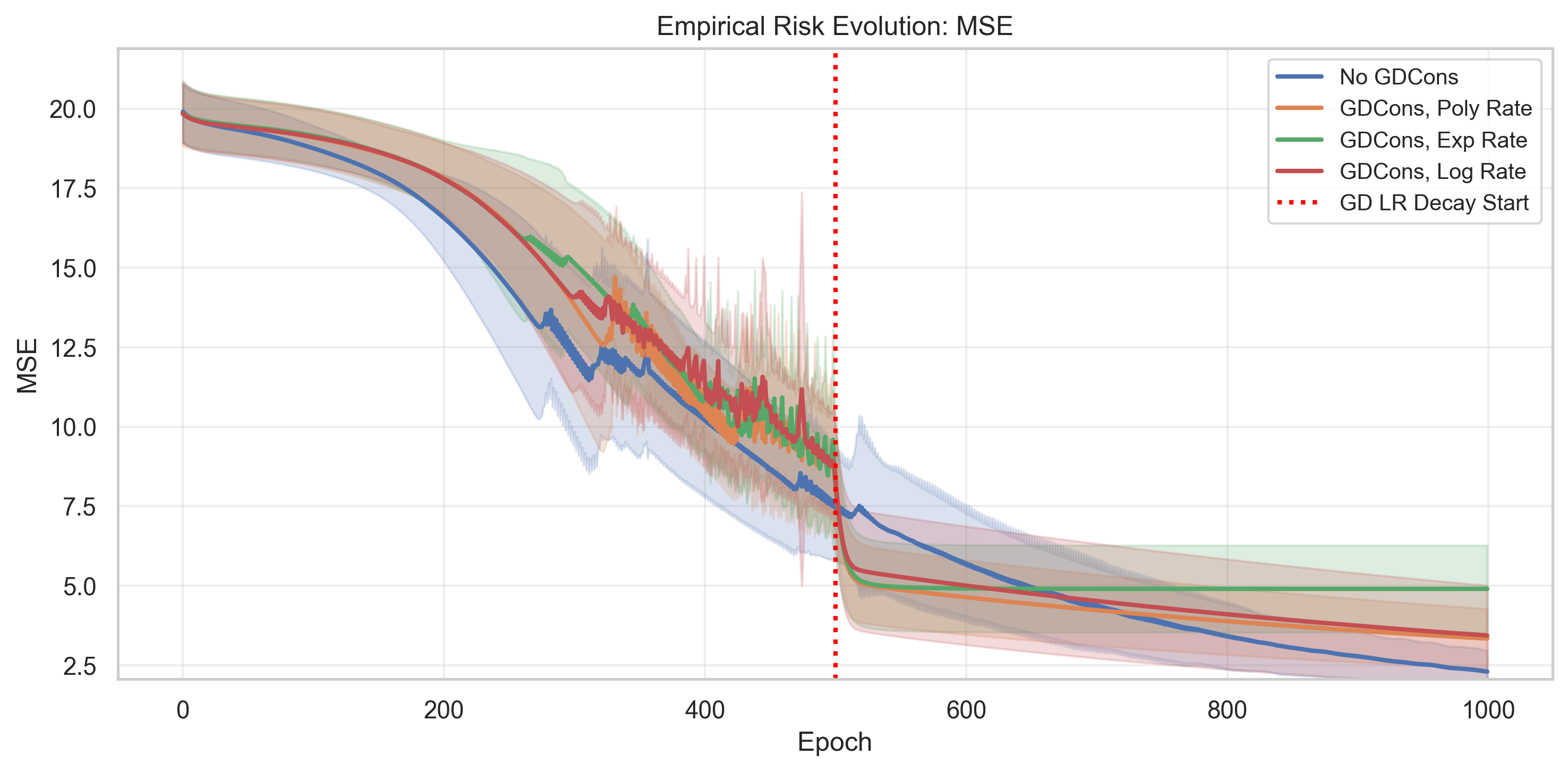}
    \end{subfigure}
    \hfill
    \begin{subfigure}{0.49\textwidth}
        \centering
        \includegraphics[width=\linewidth]{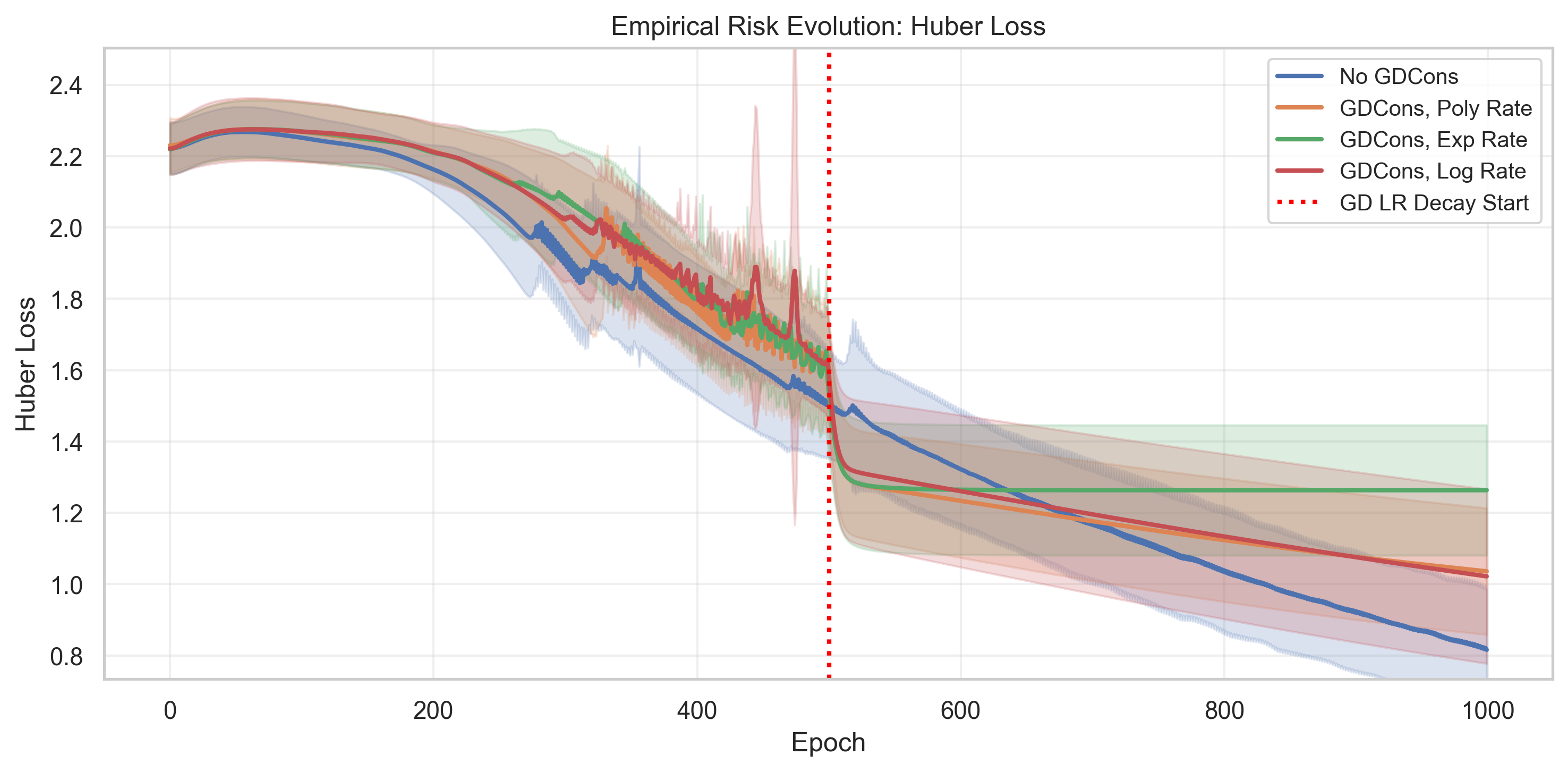}
    \end{subfigure}
    \begin{subfigure}{0.49\textwidth}
        \centering
        \includegraphics[width=\linewidth]{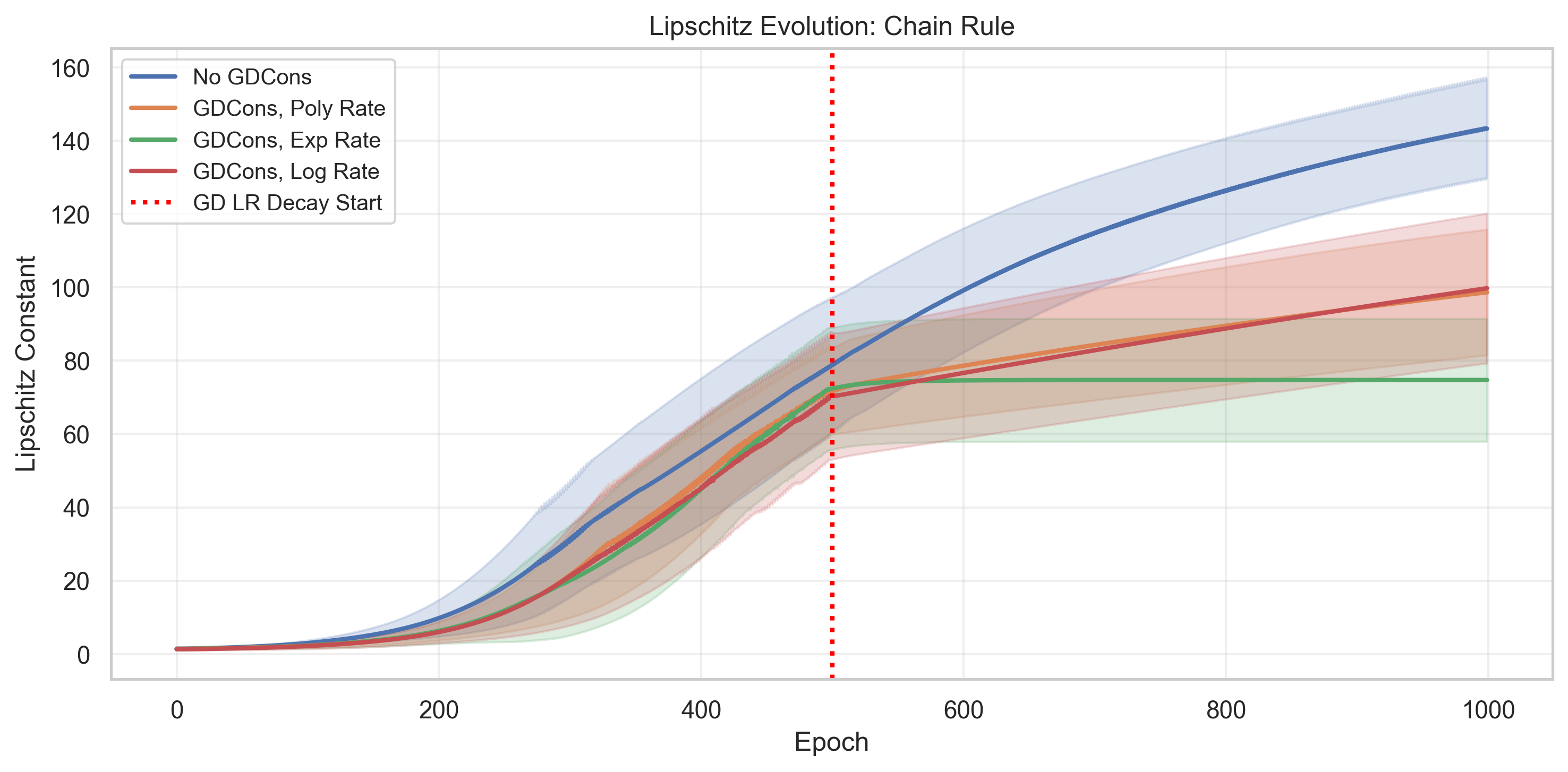}
    \end{subfigure}
    \hfill
    \begin{subfigure}{0.49\textwidth}
        \centering
        \includegraphics[width=\linewidth]{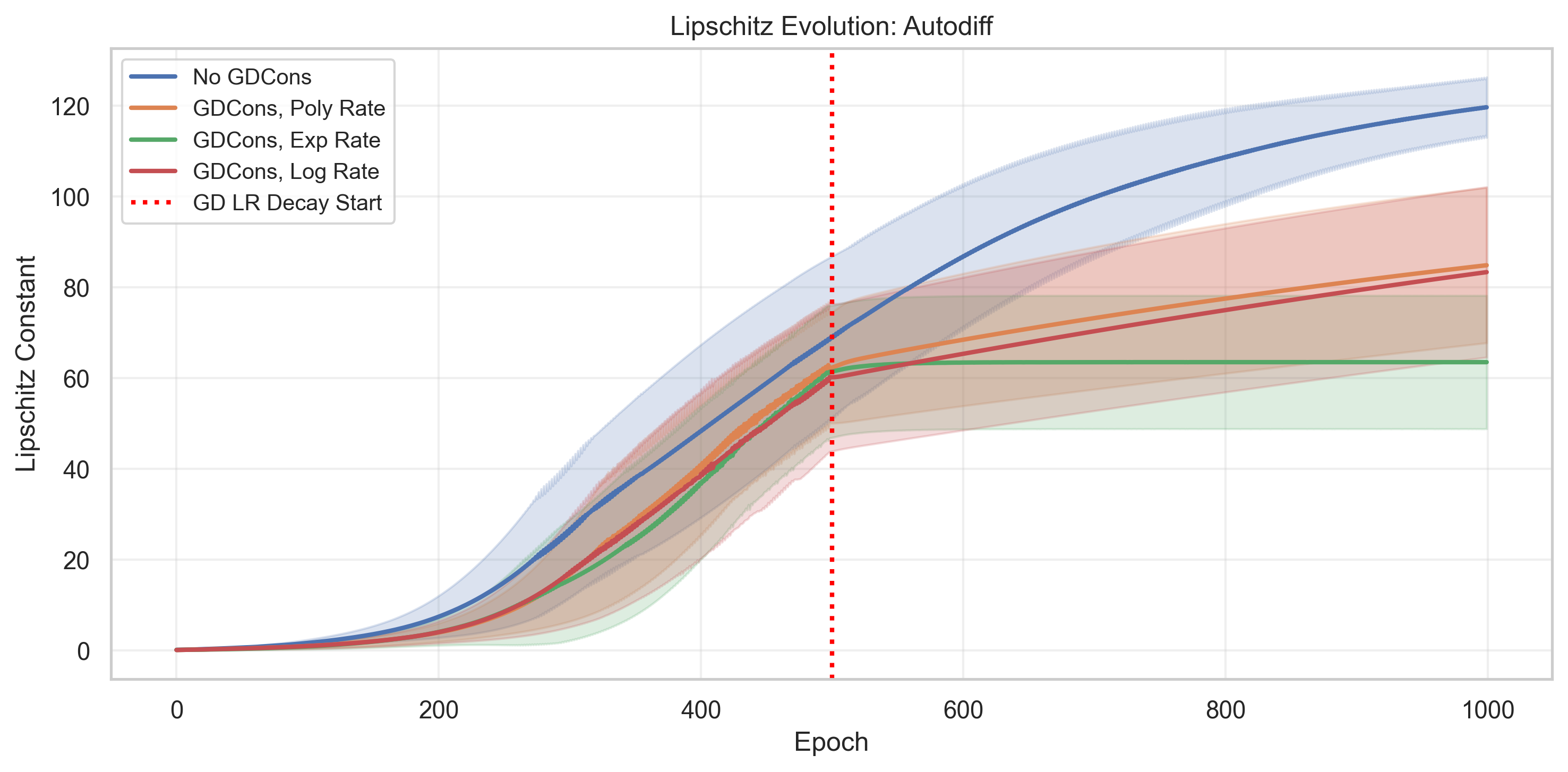}
    \end{subfigure}
    \caption{\textbf{Rate Function Training Dynamics}: We train MLPs with width $16$, depth $3$ and $\operatorname{ReLU}$ activation for $1000$ epochs on $2000$ i.i.d. uniform samples of the Forrester function with standard Gaussian noise with noise level $\beta = 0.3$, with initial LR $\alpha = 0.01$, starting our GD LR Decay Conditions at epoch $T=500$ using exponential, polynomial, and logarithmic rate functions with $r = 0.03$. We lower bound the Lipschitz constant with $6000$ gradient samples. We perform this process $10$ times, plotting the mean and standard deviation.}
    \label{fig:experiment:rate_function_training_dynamics:ReLU}
\end{figure}

We consider the training dynamics for ReLU networks (Figure~\ref{fig:experiment:rate_function_training_dynamics:ReLU}).

\end{document}